\documentclass[10pt,twocolumn,letterpaper]{article}

\usepackage{iccv}              

\definecolor{iccvblue}{rgb}{0.21,0.49,0.74}
\usepackage[pagebackref,breaklinks,colorlinks,allcolors=iccvblue]{hyperref}

\def\paperID{*****} 
\def\confName{ICCV}
\def\confYear{2025}

\title{PlantDreamer: Achieving Realistic 3D Plant Models with Diffusion-Guided Gaussian Splatting}

\author{Zane K J Hartley $\dagger$\\
\and
Lewis A G Stuart $\dagger$ \\
\and 
Andrew P French
\and
Michael P Pound \\
\and
\small School of Computer Science, University of Nottingham, Wollaton Road, NG8 1BB \vspace{0.2em}\\
}

\begin{document}
\maketitle

\begin{abstract}
Recent years have seen substantial improvements in the ability to generate synthetic 3D objects using AI. However, generating complex 3D objects, such as plants, remains a considerable challenge. Current generative 3D models struggle with plant generation compared to general objects, limiting their usability in plant analysis tools, which require fine detail and accurate geometry. We introduce PlantDreamer, a novel approach to 3D synthetic plant generation, which can achieve greater levels of realism for complex plant geometry and textures than available text-to-3D models. To achieve this, our new generation pipeline leverages a depth ControlNet, fine-tuned Low-Rank Adaptation and an adaptable Gaussian culling algorithm, which directly improve textural realism and geometric integrity of generated 3D plant models. Additionally, PlantDreamer enables both purely synthetic plant generation, by leveraging L-System-generated meshes, and the enhancement of real-world plant point clouds by converting them into 3D Gaussian Splats. We evaluate our approach by comparing its outputs with state-of-the-art text-to-3D models, demonstrating that PlantDreamer outperforms existing methods in producing high-fidelity synthetic plants. Our results indicate that our approach not only advances synthetic plant generation, but also facilitates the upgrading of legacy point cloud datasets, making it a valuable tool for 3D phenotyping applications.
\end{abstract}    
\section{Introduction}

\label{sec:intro}
\begin{figure}
\centering
    \includegraphics[width=0.5\textwidth]{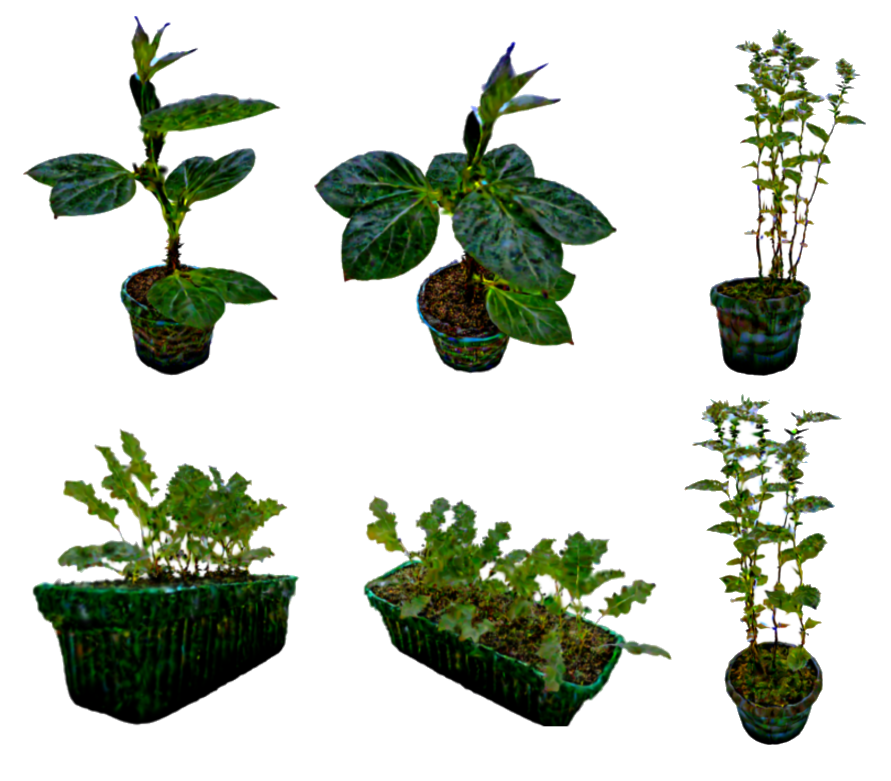}
     \caption{Rendered images of different synthetic plants generated using PlantDreamer.  The images are categorised by plant species: bean, mint, and kale, respectively.}
    \label{fig:mainFig}
\end{figure}

\def\thefootnote{†}\footnotetext{These authors contributed equally to this work}\def\thefootnote{\arabic{footnote}}

AI-driven plant phenotyping plays a crucial role in modern agriculture by supporting food security and crop development \cite{murphy2024deep}. The integration of 3D data enables efficient extraction of key phenotypic traits, and a more comprehensive analysis of complex plant structure \cite{elnashef2019tensor, 7335525, ziamtsov2019machine, feldman2021easydcp}. However, extensive 3D datasets remain scarce compared to their 2D counterparts, despite advancements in robotics, 3D reconstruction methods and plant modeling. This lack of training data poses a challenge for developing robust deep learning models for plant phenotyping \cite{yang2020crop}. To overcome this, synthetic datasets are required, but existing 3D plant generation models often lack the necessary visual realism required for downstream phenotyping tasks.\newline 
Text-to-3D models \cite{poole2022dreamfusiontextto3dusing2d, yi2023gaussiandreamer, Liang_2024_CVPR, Chen_2024_CVPR, wang2023prolificdreamer, chen2023fantasia3d, metzer2023latent, lin2023magic3d} have the potential to automate the generation of high-fidelity 3D datasets. However, existing models are designed to generate a broad range of 3D models based on an input textual description, a format unsuited to specifying precise 3D structure. This often leads to poor results when applied to complex plant morphology, producing low-quality plant representations that fail to capture the detailed geometry and texture required for effective training in downstream phenotyping tasks.\newline 
To address these limitations, we introduce \textbf{PlantDreamer}, a framework specifically designed for generating realistic 3D plant models, represented as a 3D Gaussian Splatting (3DGS) scene. Our approach incorporates several key modules. First, we integrate a depth ControlNet \cite{zhang2023adding} into the image diffusion \cite{ho2020denoising} process to ensure geometric consistency during training. Second, we enhance visual realism by training a Low-Rank Adaptation (LoRA) \cite{hu2021lora} model to better replicate real-world plant textures. Third, we introduce a new Gaussian culling algorithm into the training process to improve object convergence. \newline
PlantDreamer requires an input point cloud for the initialisation of its 3DGS model. We propose two approaches for generating this point cloud: (1) constructing purely \textit{synthetic} plants using procedural modelling methods such as L-Systems~\cite{prusinkiewicz1986graphical}, allowing the creation of realistic 3D plant models without real-world data, and (2) refining \textit{existing} plant point clouds to enhance quality and transform them into dense 3DGS representations, thereby improving existing 3D plant datasets. These approaches aim to improve the geometry of the 3D models to more closely adhere to real world plant structures, as well as enhance the realism of surface textures to more accurately represent real biological specimens. \newline
We evaluate against point cloud data captured from plant species with varied architectures (bean, kale, mint), and generated synthetic counterparts with a custom L-System designed to approximate real plant geometry and colour. To assess realism, we compare rendered images of each 3D generated plant against the ground-truth images using a Peak Signal-to-Noise Ratio (PSNR) masked metric. We evaluate model quality and multi-view alignment using T3 Bench \cite{he2023t3bench}, a standard benchmarking tool for evaluating text-to-3D methods. Our results demonstrate that PlantDreamer  outperforms GaussianDreamer and other state-of-the-art models for 3D plant generation. Specifically, PlantDreamer achieved a PSNR masked score of \textbf{16.12dB} compared to \textbf{11.01dB} for GaussianDreamer. 
Furthermore, we conduct a series of ablation studies to evaluate how point cloud characteristics, such as colour, sparsity and accuracy, impact downstream training. These studies provided insights into the type of 3D data that are effective with our approach. 
Finally, to facilitate further work in this area, we release our codebase and datasets used in this paper. 

\section{Related Work}
\label{sec:background}

\begin{figure*}
\centering
    \includegraphics[width=0.999\textwidth]{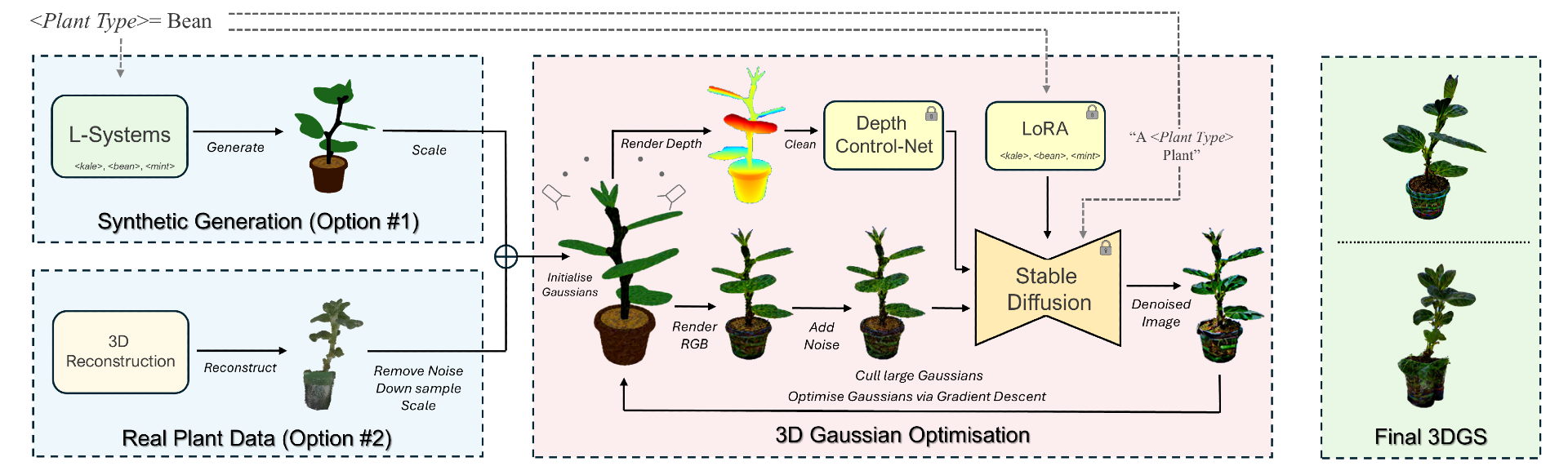}
     \caption{Overview of the process for generating a 3D bean model using PlantDreamer. Our approach either accepts a L-System mesh or a point cloud from a real plant, and outputs a 3DGS model, which is shown for the synthetic and real plant on the right side of the diagram. 
     }
    \label{fig:gaussianDreamer}
\end{figure*}

\textbf{Synthetic Plants}. Many plant phenotyping problems suffer from data scarcity as a bottleneck~\cite{yang2020crop}, and recent literature has increasingly shown that synthetic data can be used to supplement or replace real data for downstream models~\cite{hartley2024domain, Cieslak_2024_CVPR, anagnostopoulou2023realistic, klein2024synthetic, yang2022synthetic}. The generation of synthetic plants has been achieved in a number of ways. Generative techniques including GANs~\cite{goodfellow2014generative} and diffusion models~\cite{ho2020denoising} have been used to generate images for training phenotyping AI, leveraging their photorealism for 2D images~\cite{heschl2024synthset}. Structural plant modeling techniques, notably L-systems~\cite{rozenberg1980mathematical}, provide a framework for the generation of complex biological structures ~\cite{prusinkiewicz1986graphical}, but must be used together with other modeling techniques to create realistic 3D models. Recent works~\cite{Cieslak_2024_CVPR, anagnostopoulou2023realistic, hartley2024domain} use a mixture of 3D morphological modeling tools and image-to-image generative AI to achieve accurate geometry and fine control of image content. However, performing phenotyping in 2D has major limitations due to the complex structures of plants, and significant occlusion from any single viewpoint. 3D phenotyping offers greater versatility, with many recent studies on 3D phenotyping seeking to leverage this additional data~\cite{wang20223dphenomvs,boogaard2023added}, though the increased cost of 3D data acquisition remains a significant challenge~\cite{forero2022lidar,lewis2024fideltiy}. Our aim is to reduce this bottleneck by generating 3D plant data using diffusion and Gaussian Splatting.\newline

\textbf{3D Representations}. Traditionally, 3D plant data is stored as point clouds \cite{10.1371/journal.pone.0256340, zhu2024crops3d, dutagaci2020rose, agriculture13071321}.
However, point clouds are often sparse and exhibit noise from data capture. Recent view synthesis models aim to generate novel views of a scene, but have also been used for 3D reconstruction of various plants with high-fidelity and accuracy \cite{jignasu2023plant, arshad2024evaluating, Hu_2024, saeed2023peanutnerf}. 3D Gaussian Splatting \cite{kerbl20233d} has emerged as a technique in which a scene is populated with 3D Gaussians that encode colour and density at positions within an environment. These scenes use a point cloud to initialise a set of 3D Gaussians, and then optimise their position, colour and opacity. 3DGS offers strong performance in view rendering speed and accuracy, with these representations being shown to work effectively for reconstructing a variety of different plants \cite{splants2024Tommy,lewis2024fideltiy}. However, these approaches require a set of real images in order to reconstruct the environment, and capturing vast amounts of image data can be time consuming and expensive. Our aim is to generate synthetic plants without the need for a set of ground truth images, while also supporting the conversion of real plant point clouds into 3DGS formats with its associated benefits. \newline

\textbf{Text-to-3D Models}. Text-to-3D models \cite{li2024generativeaimeets3d} aim to generate a diverse range of 3D models based on an input text prompt. A variety of such models have been proposed \cite{wang2023prolificdreamer, chen2023fantasia3d, metzer2023latent, lin2023magic3d}, demonstrating impressive results for a wide range of 3D objects, with several of these models being optimised for 3DGS generation \cite{yi2023gaussiandreamer, Liang_2024_CVPR, Chen_2024_CVPR}. One such approach, GaussianDreamer by Yi et al~\cite{yi2023gaussiandreamer} uses a two-stage approach to generation. First, generating a simple model using 3D diffusion~\cite{jun2023shap} and converting it to a 3D Gaussian splat. Second, utilising score distillation sampling (SDS) \cite{poole2022dreamfusiontextto3dusing2d} to improve the initial 3DGS object. This is an iterative process that renders an image of the object and uses 2D Diffusion~\cite{rombach2022high} to improve its quality, which then enhances the Gaussian splat using gradient descent optimisation. While this approach produces visually appealing 3D assets, we find it shares limitations with other works for plant generation, specifically when controlling geometry and with realistic texturing. \newline
Significant progress has been made in improving object structural controllability during the generation process \cite{li2023mvcontroladdingconditionalcontrol, chen2023control3dcontrollabletextto3dgeneration, seo2024let2ddiffusionmodel}, but we found that these methods remain insufficient for accurately replicating the complex, organic geometry of plant structures using simple priors. Similarly, advancements in texture synthesis have led to notable improvements in object realism \cite{cao2023texfusionsynthesizing3dtextures, richardson2023texture}, however existing methods are not specifically designed for photorealistic plant modeling. \newline 
Several domain-specific text-to-3D models have been proposed, such as avatars \cite{jiang2023avatarcraft, Cao_2024_CVPR}, faces \cite{zhou2024headstudiotextanimatablehead, han2023headsculptcrafting3dhead} and room scenes \cite{hollein2023text2room}. However, these proposed frameworks are optimised for their respective domains and are not directly transferable for realistic synthetic 3D plant generation.

\section{Methodologies}
\label{sec:methodologies}

\subsection{Achieving Object Generation with 3DGS}

Our framework builds upon existing 3DGS text-to-3D approaches \cite{yi2023gaussiandreamer}, which  optimise a 3D scene into a complete object through the following iterative process: (1) selecting a new camera viewpoint and rendering an image, (2) introducing noise and applying diffusion-based denoising to refine the image, and (3) updating the 3DGS representation accordingly. \newline 
3DGS consists of a set of 3D Gaussians in Euclidean space, which can be rendered efficiently to produce images. A 3DGS scene is parameterised as $\theta = \{\mu_k, \Sigma_k, \alpha_k, c_k\}$, where $\mu_k$ represents the Gaussian center positions, $\Sigma_k$ the covariance, $\alpha_k$ the opacity, and $c_k$ the color, for each Gaussian in scene $k$. Rendering an image involves casting rays into the scene, with each intercepted Gaussian contributing to the final pixel based on its current opacity, colour and the ray transmittance. This is expressed as:

\begin{equation}
x = \sum_{p \in \text{image}} \sum_{i \in \mathcal{G}_p} T_i \cdot \alpha_{i} \cdot c_i  
\end{equation}

Where $p$ represents the pixels in the image, $\mathcal{G}_p$ the intercepted Gaussians, $T$ the current ray transmittance, $\alpha$ the opacity of the Gaussian in the context of $p$ and $c$ is the Gaussian colour. \newline
To refine the 3D Gaussians into a realistic object based on a given text prompt, we use Stable Diffusion 2.1\cite{rombach2022high} to generate high-quality scene images, denoted as $\phi$. The optimisation process is controlled via the SDS loss, which determines the difference between the predicted denoised image and the rendered image: 

\begin{equation}
L_{SDS}(\theta) =  \mathbb{E}_{t,\epsilon}[w(t)|\nabla_x( \hat{\epsilon}_\phi(\tilde{x}_t, t, y) - x_t)|]
\end{equation} 

where $t$ is the current diffusion timestep, $w(t)$ the weighting function for the current timestep, $\hat{\epsilon}$ the predicted noise from the diffusion model and $y$ the text prompt. Hence, this loss function predicts the update direction of the scene solely based on the rendered image, effectively decoupling the complex 3D representation from the diffusion process. The 3DGS scene is then optimised using  stochastic gradient descent:

\begin{equation}
\theta_{t+1} =  \theta_t - \gamma \cdot\nabla L_{SDS}(\theta_t)
\end{equation}

where $\gamma$ represents the learning rate. \newline
This process of rendering novel views, refining the images using diffusion and updating the 3DGS scene continues for a fixed number of iterations, gradually improving the texture fidelity and geometric coherency.

\subsection{Enhancing Texture Realism}

General diffusion models are ill-suited for generating realistic plant surfaces, often producing saturated colours and cartoon-like textures, which fail to capture the distinctive characteristics of specific plant species or varieties that are underrepresented in the original training set. \newline
To address this, we incorporate a Low-Rank Adapter (LoRA) \cite{hu2021lora} to provide style transfer to the target species. For each species, we train the LoRA module on a dataset of 30 images that were captured of the different available plants against a plain background. \newline
In addition, we observed that large, erroneous Gaussians were extending across the plant structure during training, causing overly smooth or distorted surfaces, which restricted the diffusion models from projecting detailed texturing onto the object. To resolve this issue, we developed a novel culling algorithm, that removes Gaussians which exceed the standard volume for all other Gaussians. This process is as follows:

\begin{equation}\label{eq:large_gaussian removal}
\text{cull} = \begin{cases} \text{True}, & \text{if } V > \mu_v + C\sigma_v \\ \text{False}, & \text{otherwise} \end{cases}
\end{equation}

\begin{equation}\label{eq:gaussian_volume}
V =\sqrt{\sum_{i} \left( e^\text{s}_i \right)^2},  
\end{equation}

where $V$ is the volume, $C$ the culling threshold, $v$ the volume across all Gaussians, $s$ the Gaussian scale and $i$ is each dimension.

\subsection{Controlling Plant Geometry}

We found that 2D diffusion struggles to generate images that are 3D aware, leading to feature hallucinations where occluded or distant parts of the object are incorrectly interpreted. Over many iterations, plant structures become fused together and fine details, such as individual leaves, are gradually lost. Therefore, we incorporate a depth map ControlNet~\cite{zhang2023adding}, which provides visual conditioning to the diffusion model based on the depth of the noised image. To prevent unintended drift from the initial model, we preserve the original point cloud geometry throughout the training loop and render each depth map from this static reference. This ensures that the model remains anchored to its initial geometry, preserving morphological detail, while still allowing for flexible texture and detail enhancements. Finally, we employ mask thresholding, erosion, and dilation techniques to refine the generated depth maps, which reduced issues with the background being included in texturing, which improves the overall fidelity. The diffusion model is updated to accommodate these changes as:

\begin{equation}
\hat{\epsilon}_{\phi'}(\tilde{x}_t, t, y, d), \phi' = \phi(x|C,L)
\end{equation}

where $d$ denotes the cleaned depth map of the fixed initial Gaussian model, $C$ the controlnet weights and $L$ the LoRA weights. 

\begin{figure}[b]
    \centering
\includegraphics[width=0.355\textwidth]{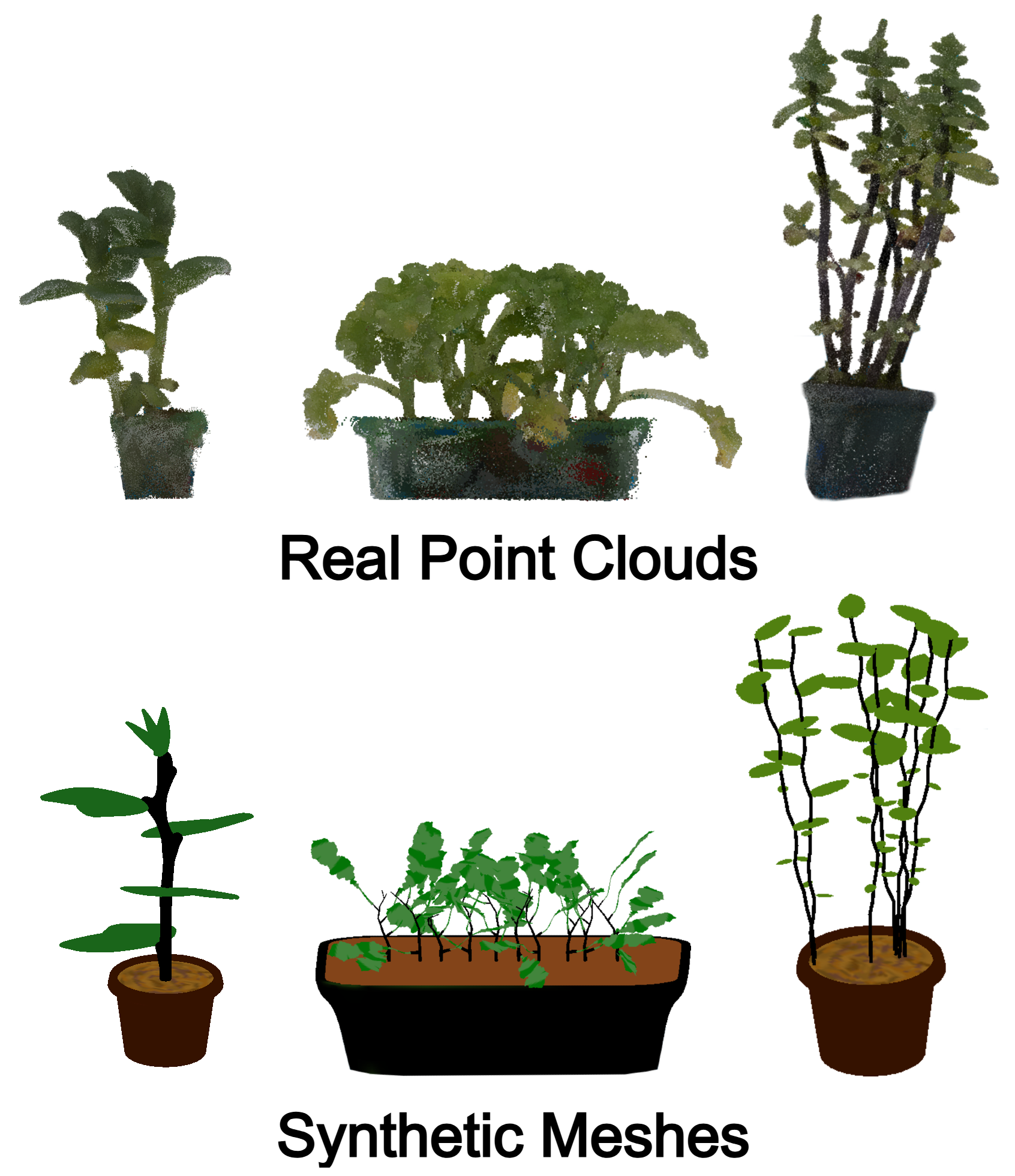}
     \caption{Our real bean, kale and mint captured point clouds with the corresponding synthetic meshes.}
     \label{fig:inputData}
\end{figure}

\subsection{Initisalisaion with L-Systems and Real Point Clouds}

3D Gaussians are initialised using a point cloud to ensure effective starting 3D object structure and assisting in effective refinement. Existing approaches, such as GaussianDreamer, often rely on 3D diffusion models like Shape-E \cite{jun2023shap} for effective scene initialisation. These models are useful for general image categories, but produce sparse, low-fidelity point clouds which lack the geometric precision and colour for realistic 3D plant generation. \newline
 PlantDreamer address this by using meshes generated using L-Systems for initialisation, which use rule-based procedural generation. By incorporating morphological rules, L-Systems facilitate generation of individually unique meshes that resemble real-world plant geometry, resulting in realistic synthetic outputs for a variety of different species. \newline
 PlantDreamer also supports initialisation using real plant point cloud reconstructions. This allows point cloud data to be transformed into high-fidelity 3DGS scenes, enhancing the quality and usability of existing datasets. To refine potentially noisy point clouds, we apply several pre-processing steps. Each point cloud is filtered with statistical outlier removal~\cite{Zhou2018} to remove noise. Next, point clouds are downsampled using voxel grid sampling, which downsample based on the plant size to reduce the point cloud to approximately 100,000 points, which ensures uniform distribution across surfaces. The point cloud is then scaled appropriately and translated to the centre of the scene.

\section{Experiments}
\label{sec:experiments}

To effectively evaluate PlantDreamer's ability to generate high-fidelity 3DGS models, we selected bean, mint and kale as our subject plant species. These species exhibit distinct structure and texturing, posing a challenge for detailed 3D plant generation. Kale plants exhibit dense leaf structure with heavy occlusion, mint plants have long, thin stems which are difficult to identify and replicate in 3D, and bean plants have large, broad leaves that create occlusions and require accurate texturing. By evaluating over a diverse range of morphological characteristics, we aim to assess PlantDreamer's ability to generalise across different plant species, given a valid trained LoRA and L-System prior. \newline
We captured point clouds from a set of real plants for each species using 3DGS, and generated a set of synthetic mesh counterparts using L-Systems, which share similar geometry with the point clouds while still being unique. In total, we captured twelve bean, three kale, six mint plants, each with corresponding real and synthetic representations. 


\begin{figure*}[b]

\centering
\setlength{\tabcolsep}{1.0pt}
\renewcommand{\arraystretch}{0.5}
\begin{tabular}{ccc} 


\includegraphics[width=0.333\textwidth]{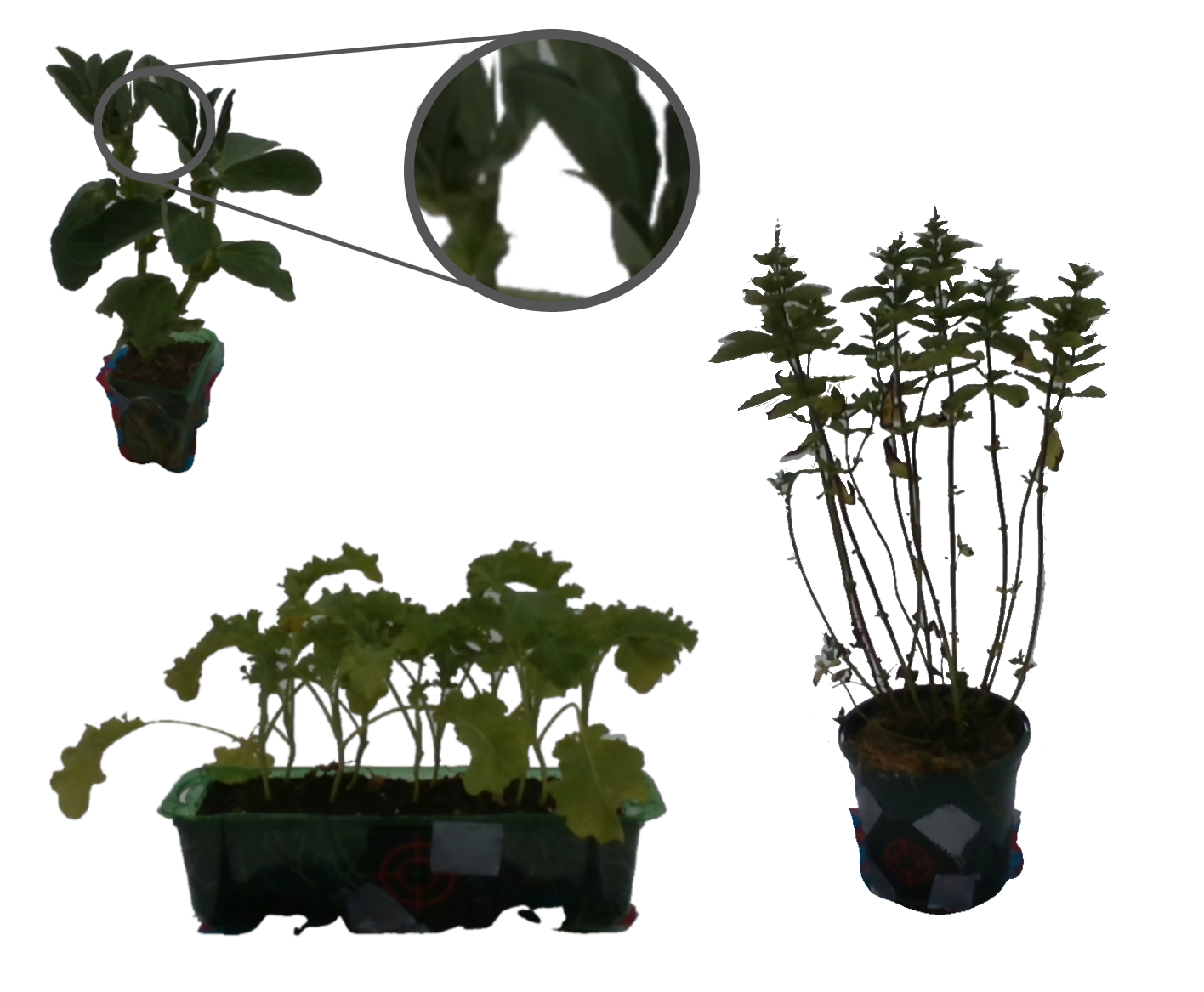} & \includegraphics[width=0.3333\textwidth]{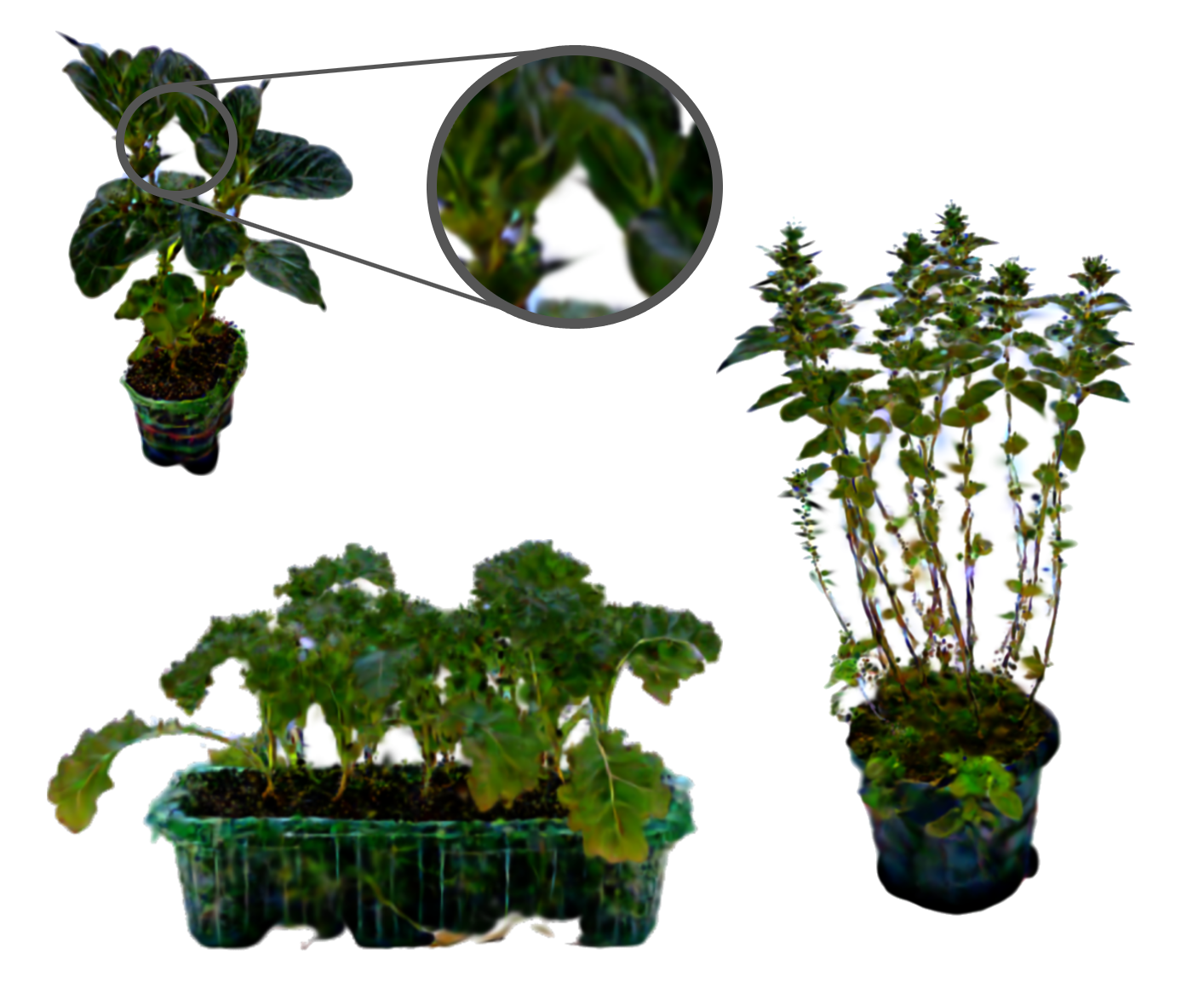} & \includegraphics[width=0.3333\textwidth]{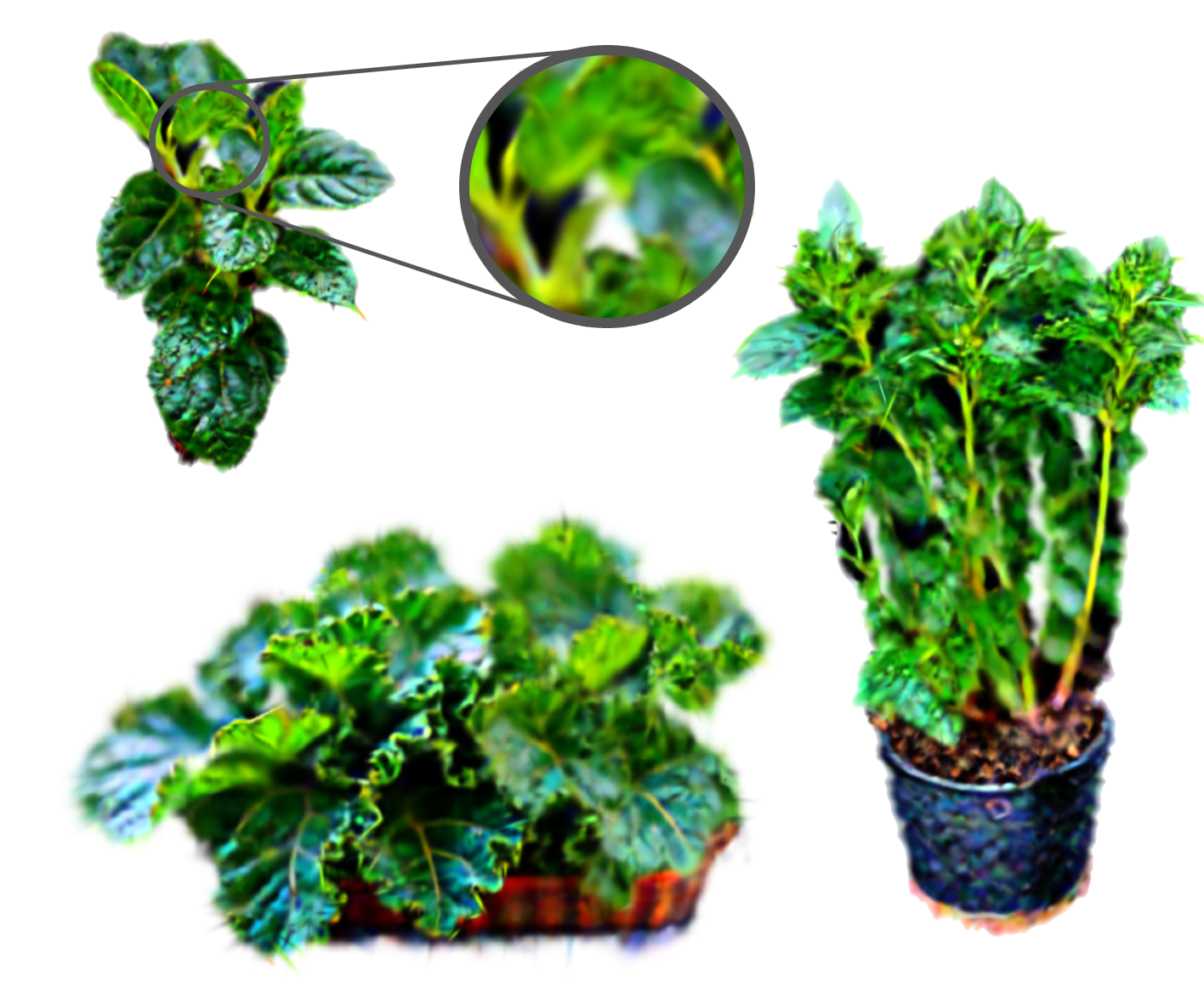}  \\ 

\vspace{0.1cm}

Ground Truth & PlantDreamer &  GaussianDreamer\\

\end{tabular}

\caption{A comparison between the ground truth and 3D plant models generated by PlantDreamer and GaussianDreamer. Each model was initialised with the same real point cloud for the bean, kale and mint respectively.}
\label{fig:RealPlantComparison}

\end{figure*}

\begin{figure*} 

\centering
\setlength{\tabcolsep}{1.0pt}
\renewcommand{\arraystretch}{0.5}
\begin{tabular}{ccccccc}


& \multicolumn{2}{c}{\textit{Bean Plant}} & \multicolumn{2}{c}{\textit{Kale Plant}}  & \multicolumn{2}{c}{\textit{Mint Plant}} \vspace{0.2cm}\\ 

\raisebox{3em}{\parbox{2.3cm}{\centering Latent-NeRF }} & \includegraphics[width=0.13\textwidth]{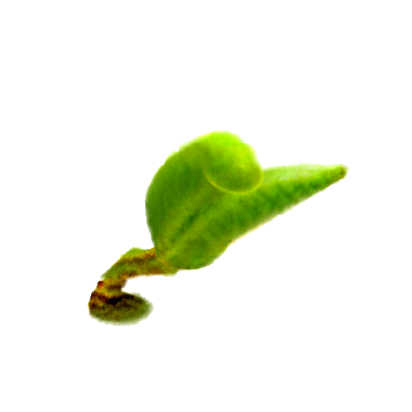} & \includegraphics[width=0.13\textwidth]{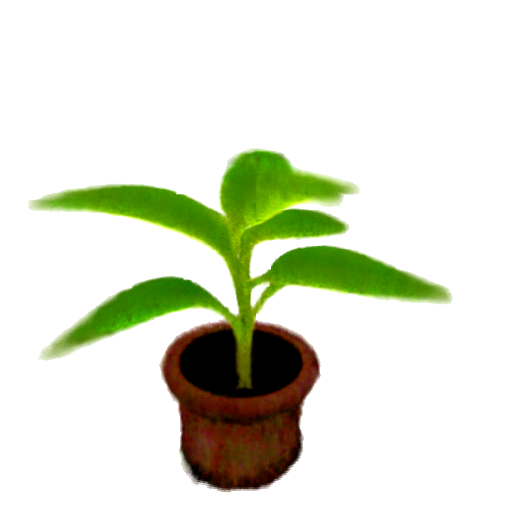} & \includegraphics[width=0.13\textwidth]{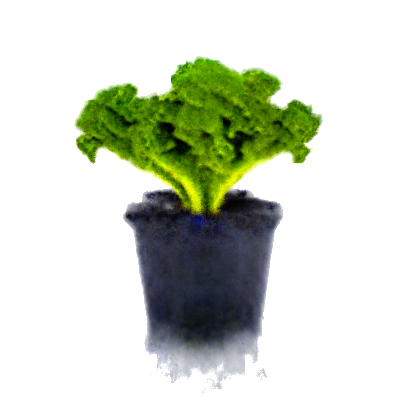} & \includegraphics[width=0.13\textwidth]{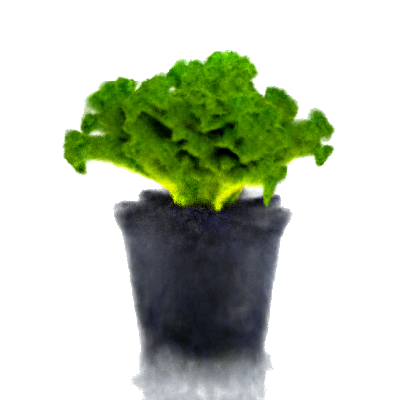} & \includegraphics[width=0.13\textwidth]{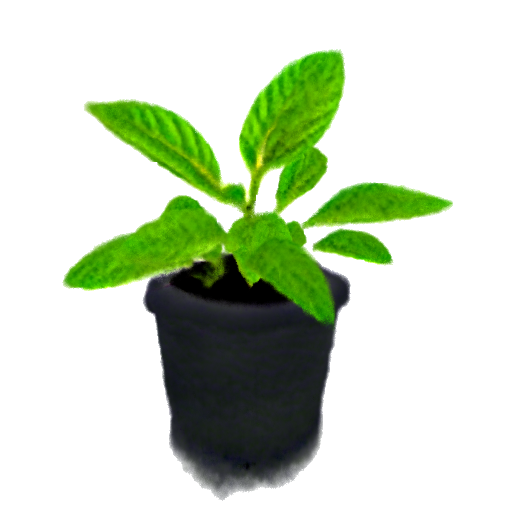} & \includegraphics[width=0.13\textwidth]{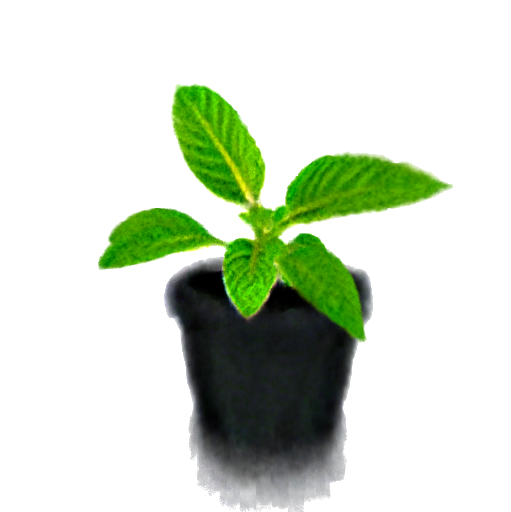} \\ 

\raisebox{3em}{\parbox{2.3cm}{\centering Magic3D }} & \includegraphics[width=0.13\textwidth]{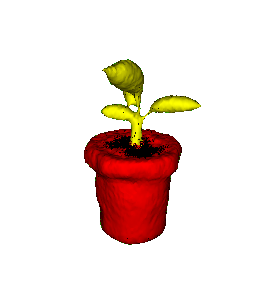} & \includegraphics[width=0.13\textwidth]{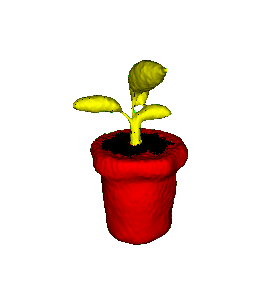} & \includegraphics[width=0.13\textwidth]{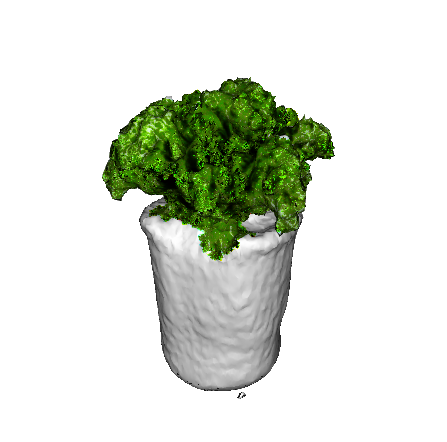} & \includegraphics[width=0.13\textwidth]{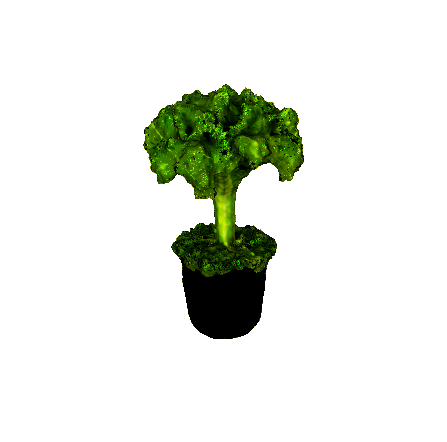} & \includegraphics[width=0.13\textwidth]{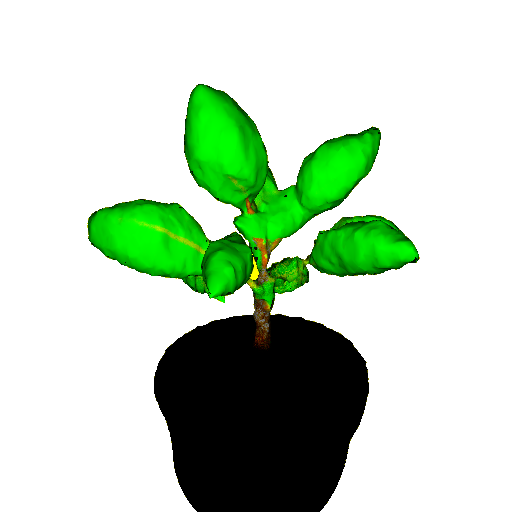} & \includegraphics[width=0.13\textwidth]{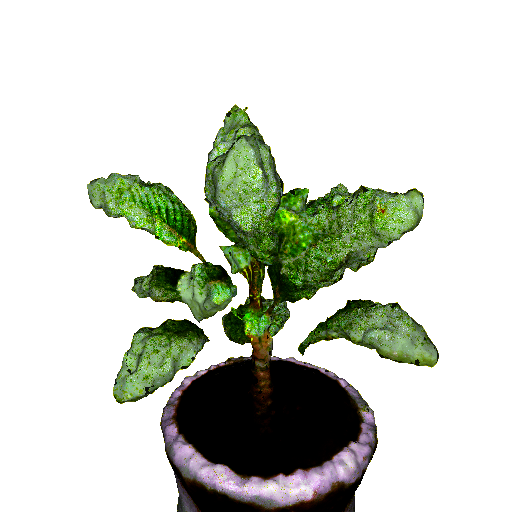} \\ 

\raisebox{3em}{\parbox{2.3cm}{\centering Fantasia3D }} & \includegraphics[width=0.13\textwidth]{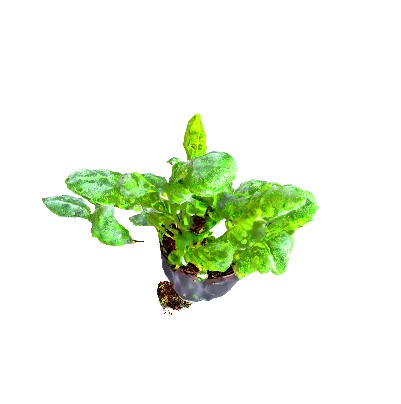} & \includegraphics[width=0.13\textwidth]{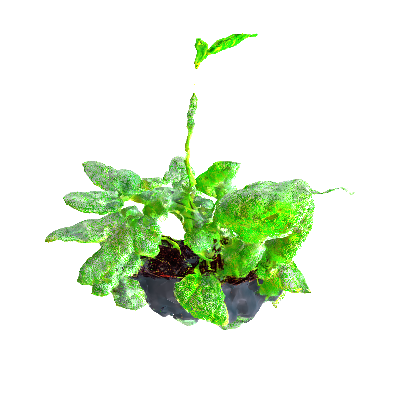} & \includegraphics[width=0.13\textwidth]{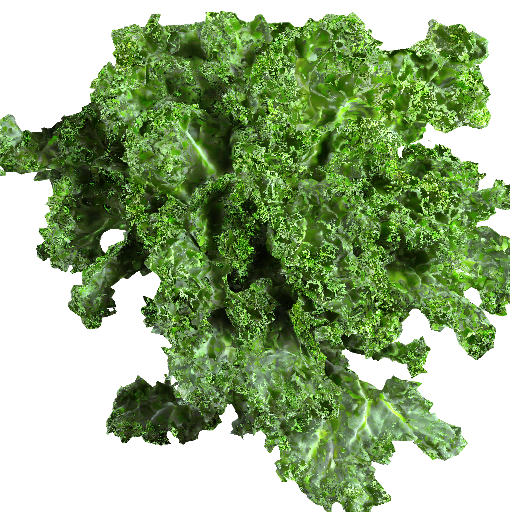} & \includegraphics[width=0.13\textwidth]{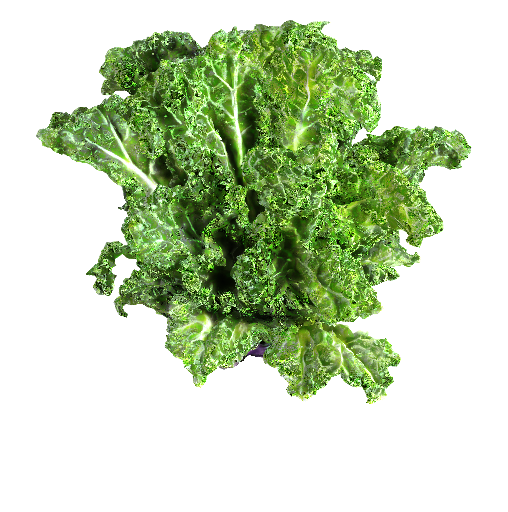} & \includegraphics[width=0.13\textwidth]{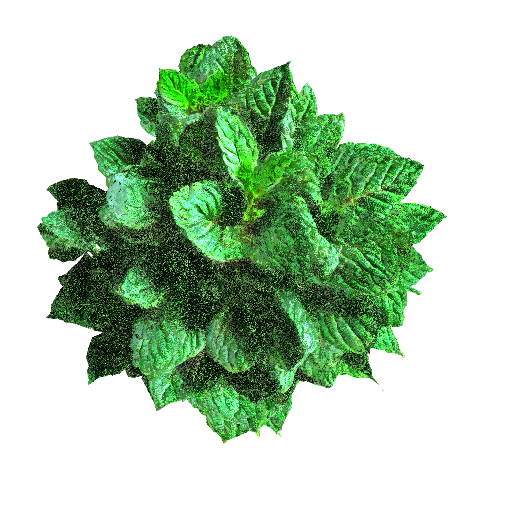} & \includegraphics[width=0.13\textwidth]{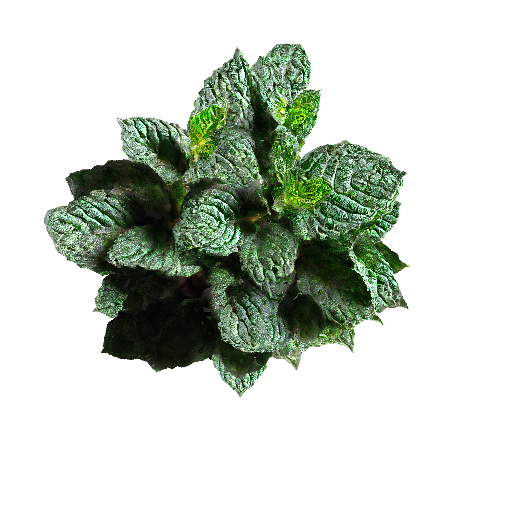} \\ 

\raisebox{3.5em}{\parbox{2.3cm}{\centering GaussianDreamer }} & \includegraphics[width=0.15\textwidth]{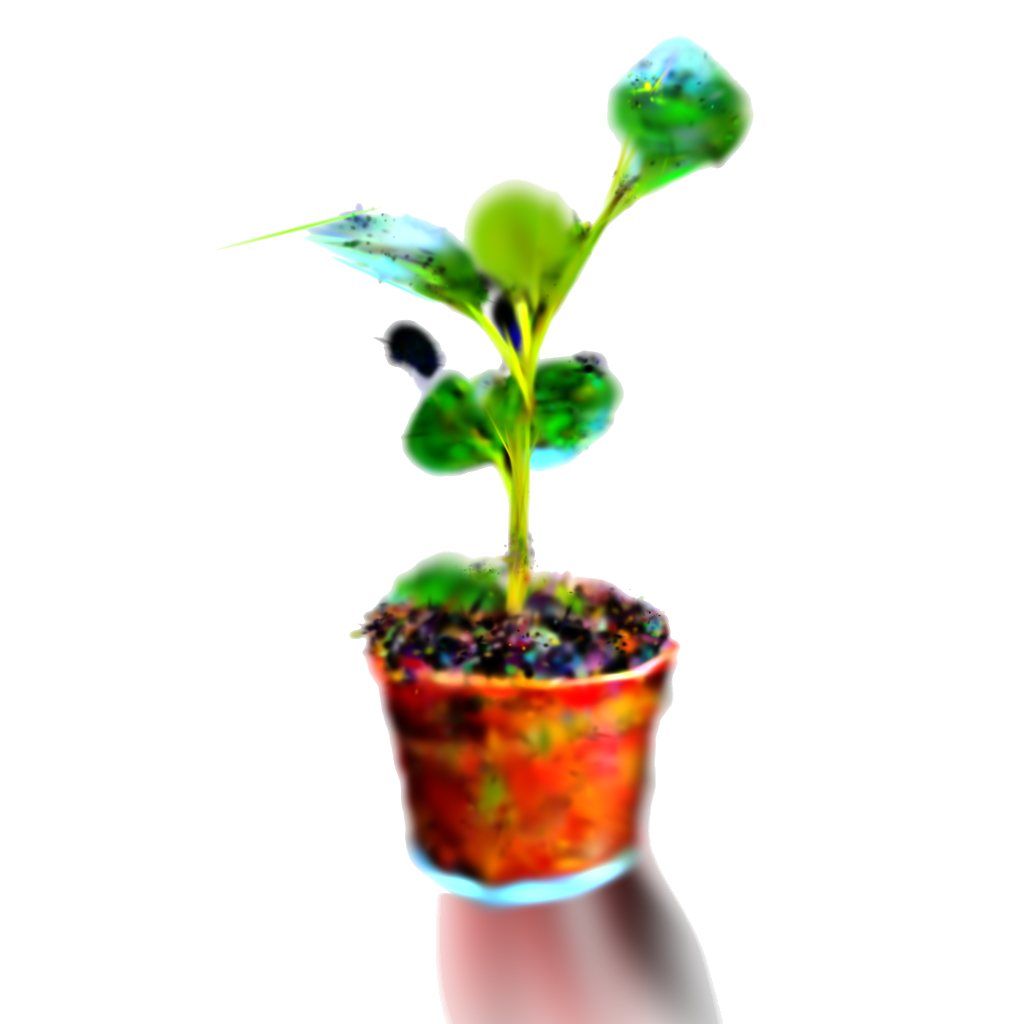} & \includegraphics[width=0.15\textwidth]{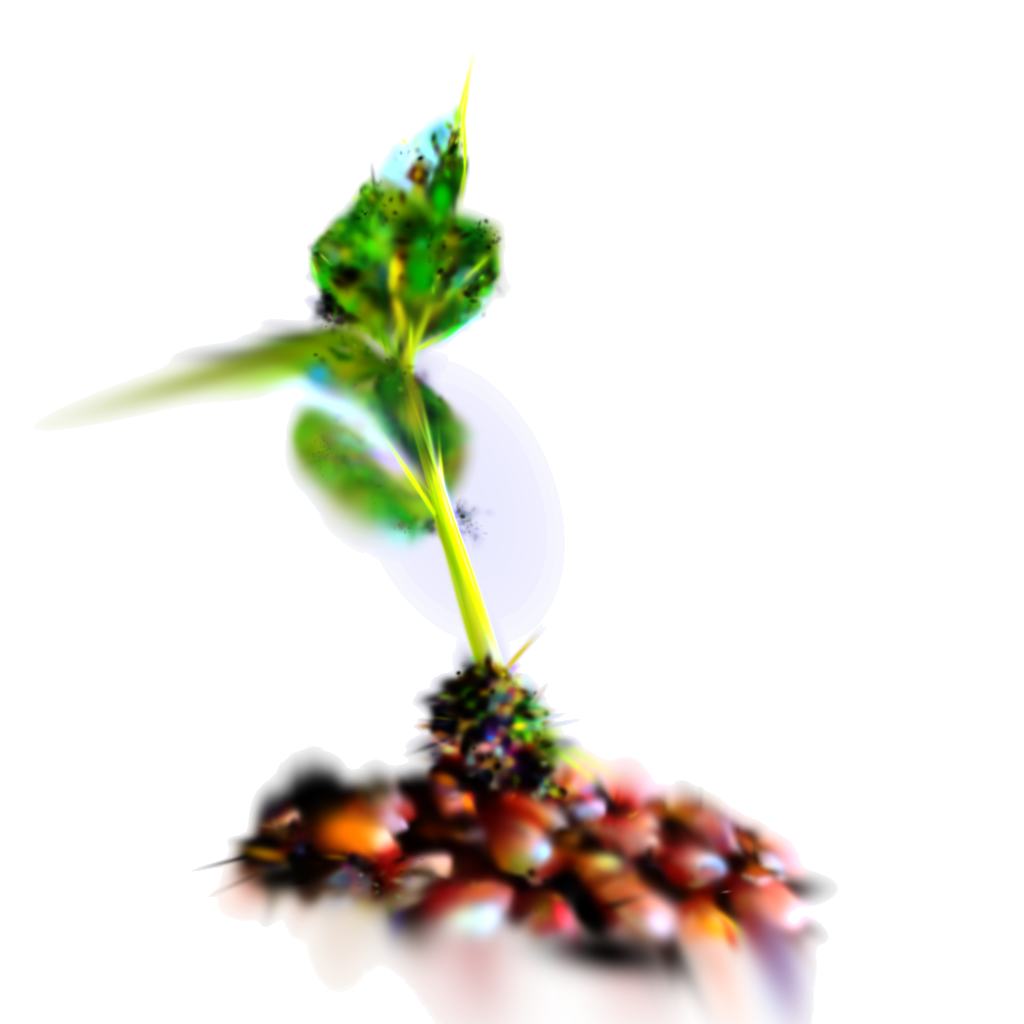} & \includegraphics[width=0.15\textwidth]{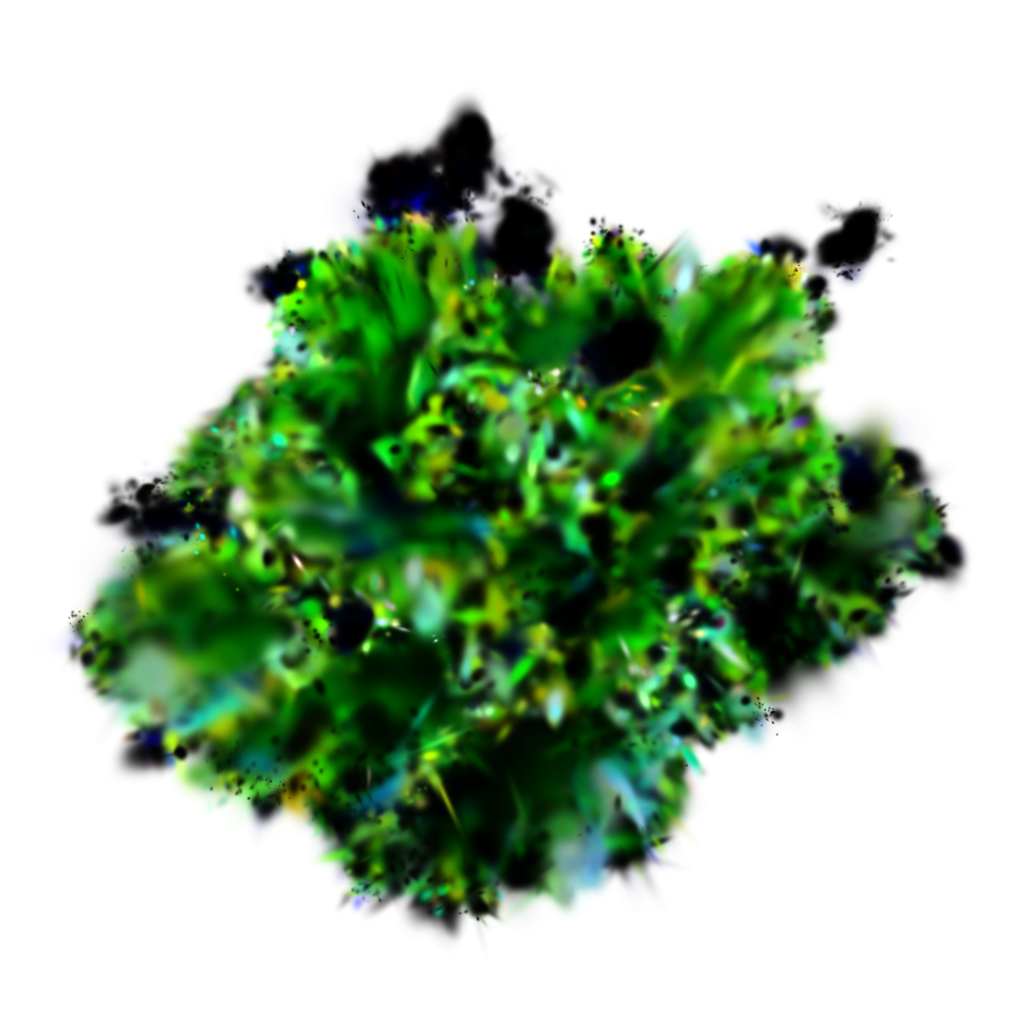} & \includegraphics[width=0.15\textwidth]{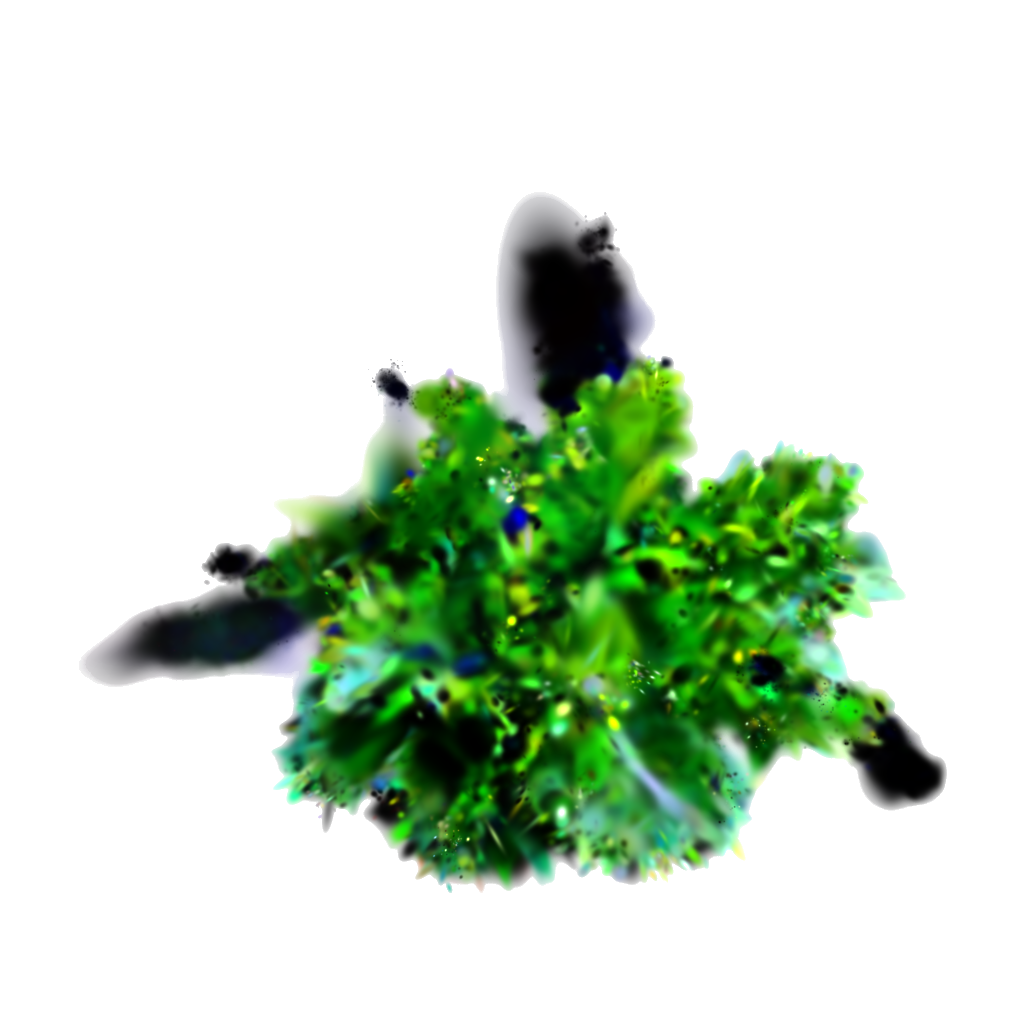} & \includegraphics[width=0.15\textwidth]{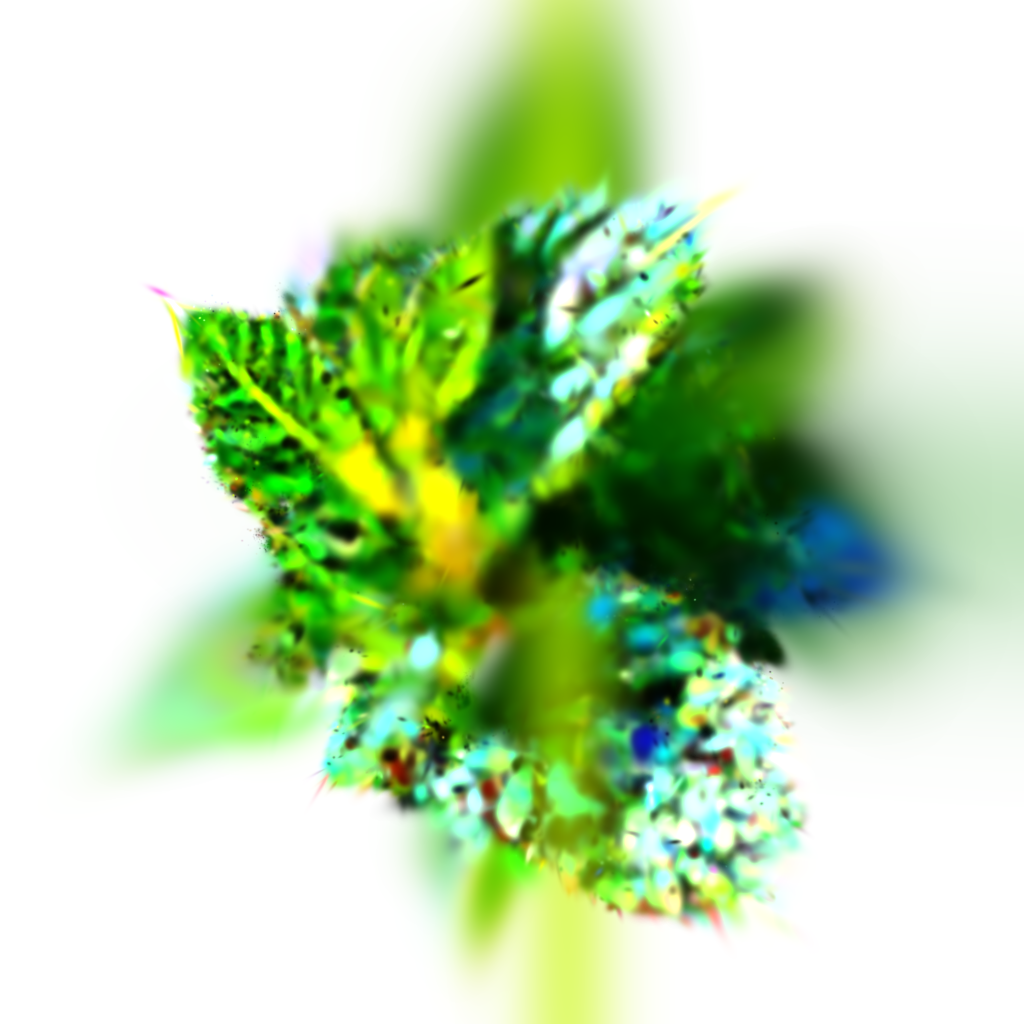} & \includegraphics[width=0.15\textwidth]{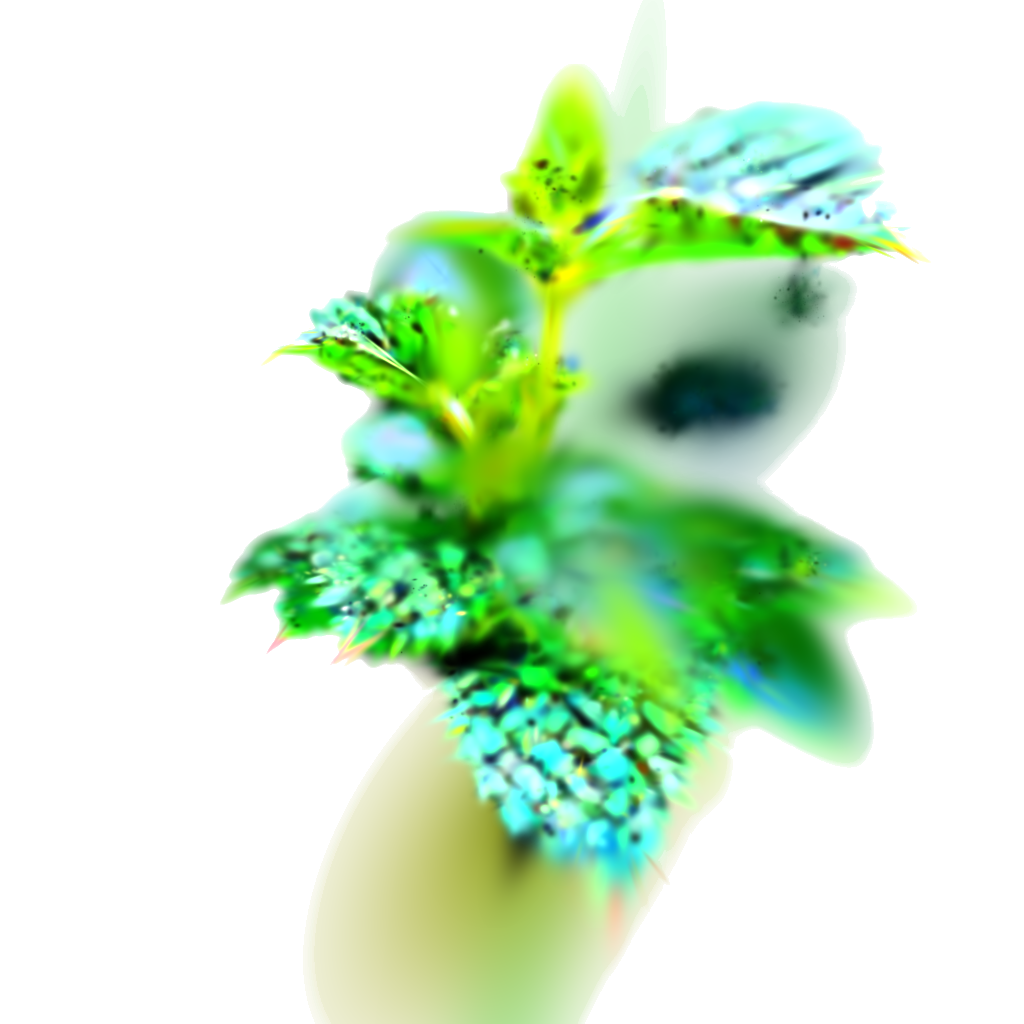} \\ 

\raisebox{3.5em}{\parbox{2.3cm}{\centering PlantDreamer }} & \includegraphics[width=0.15\textwidth]{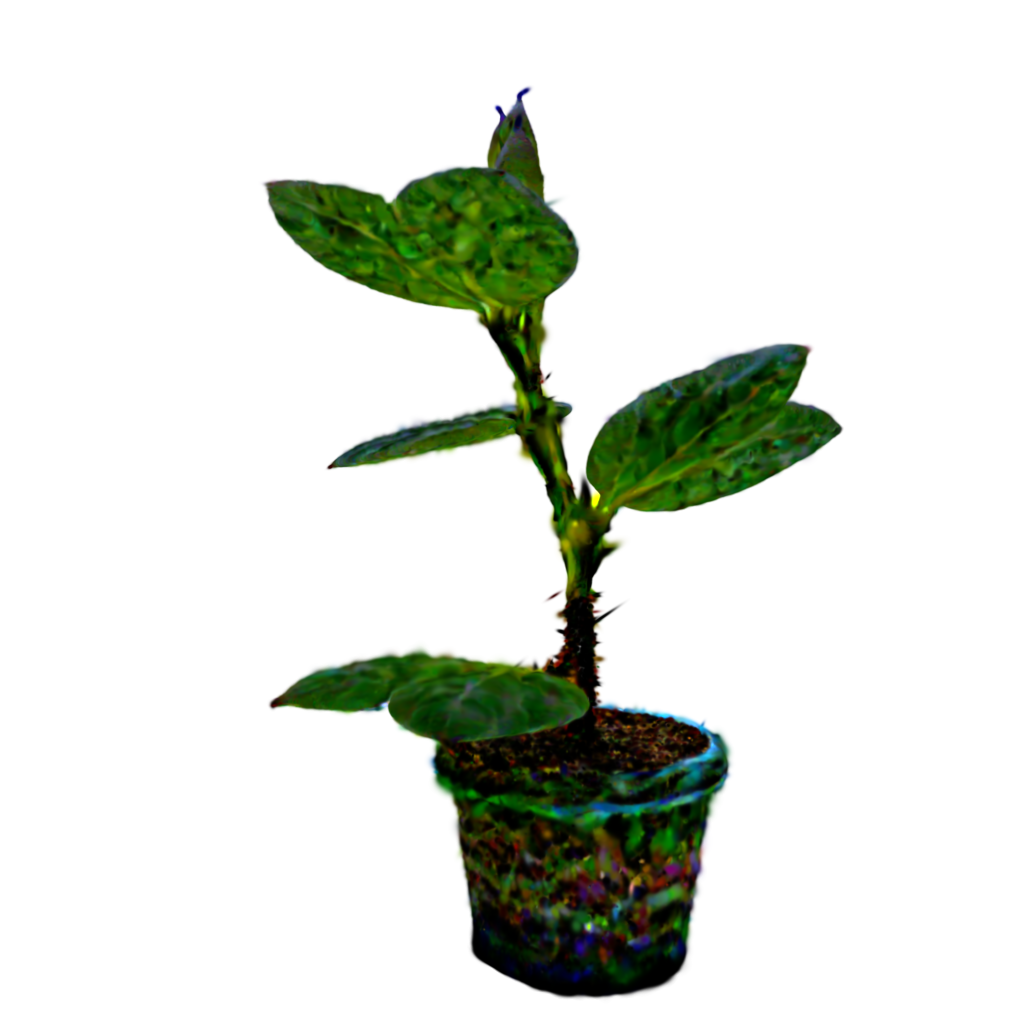} & \includegraphics[width=0.15\textwidth]{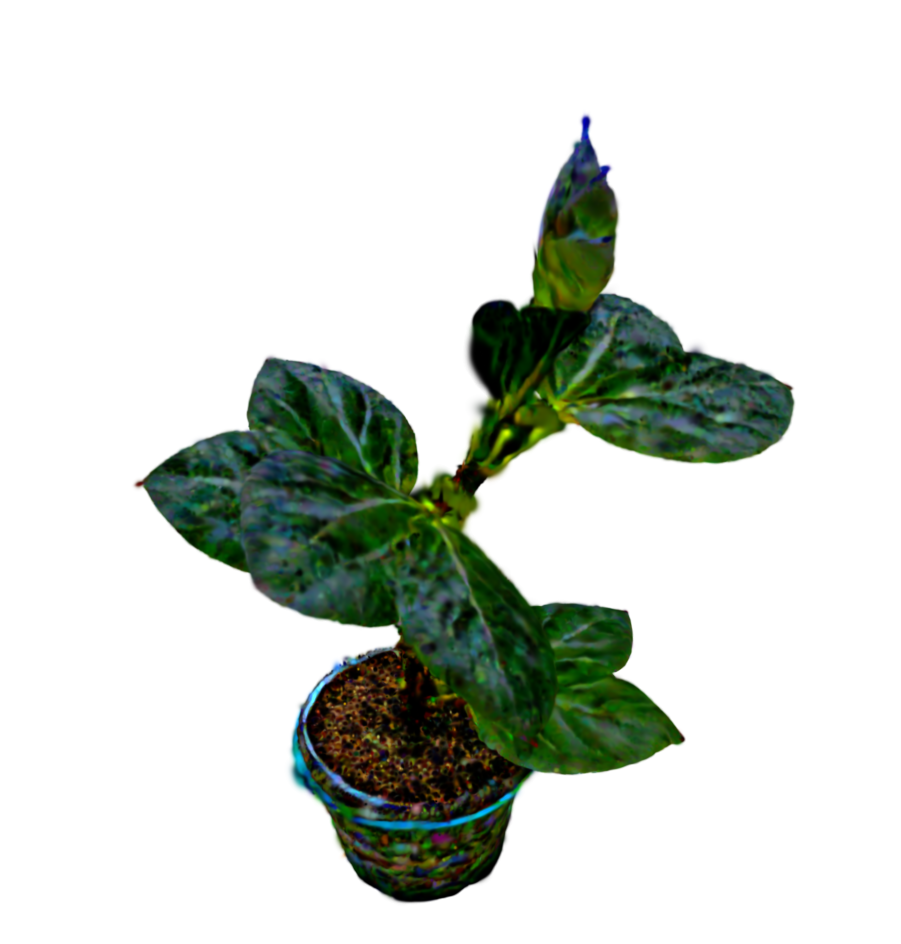} & \includegraphics[width=0.15\textwidth]{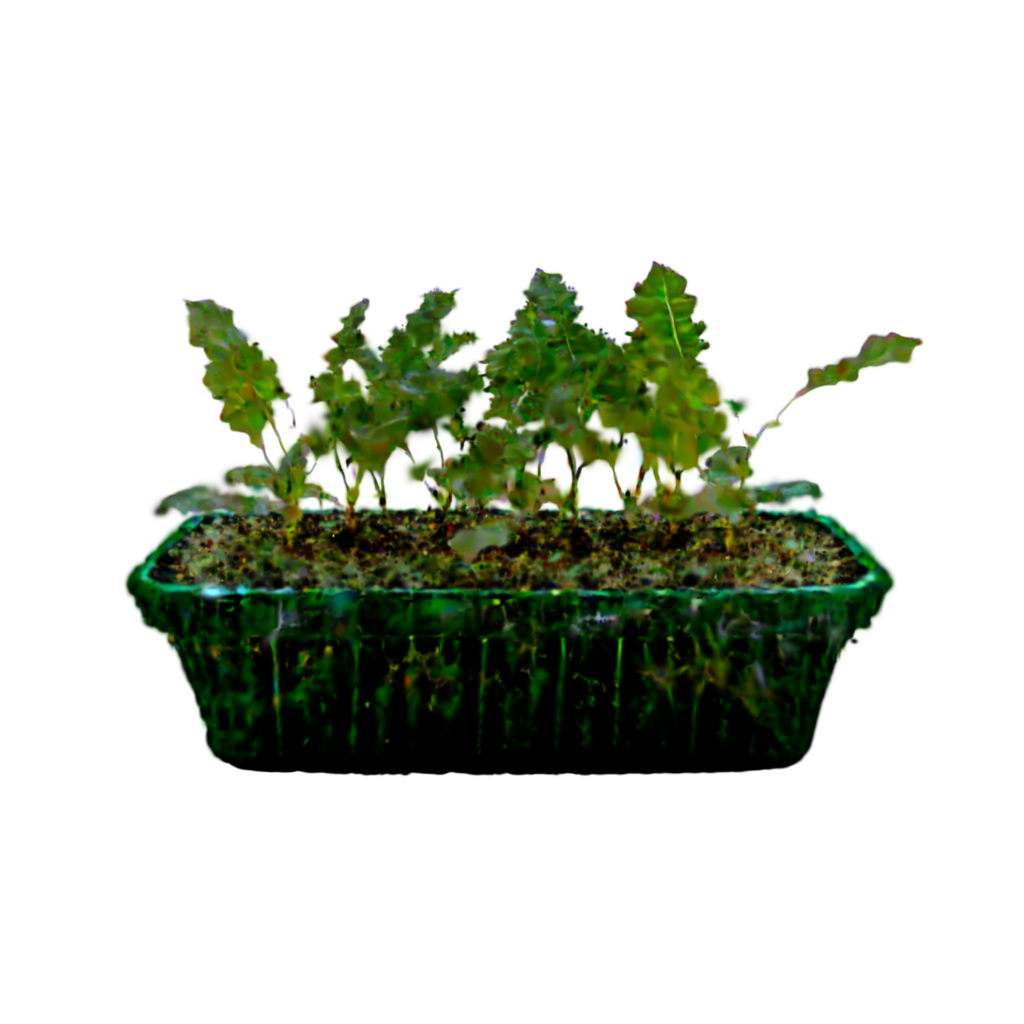} & \includegraphics[width=0.15\textwidth]{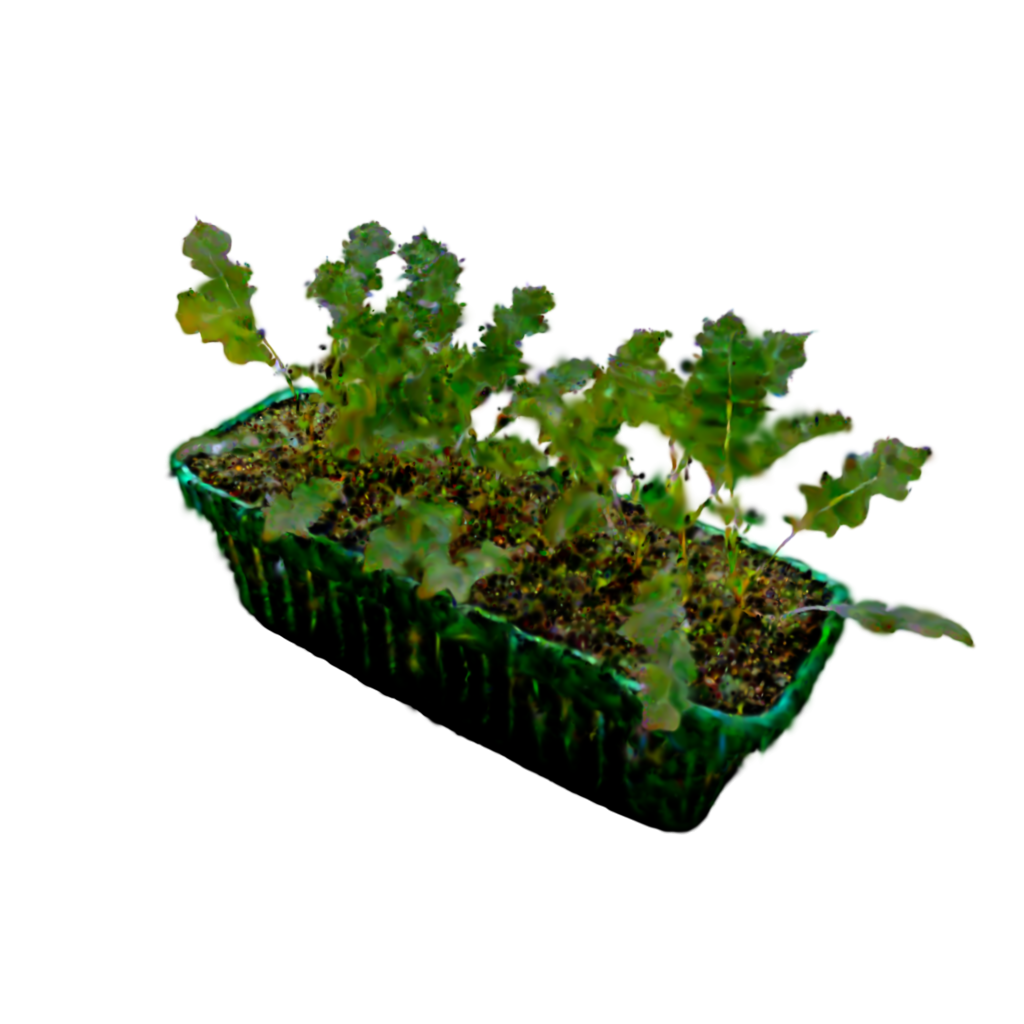} & \includegraphics[width=0.15\textwidth]{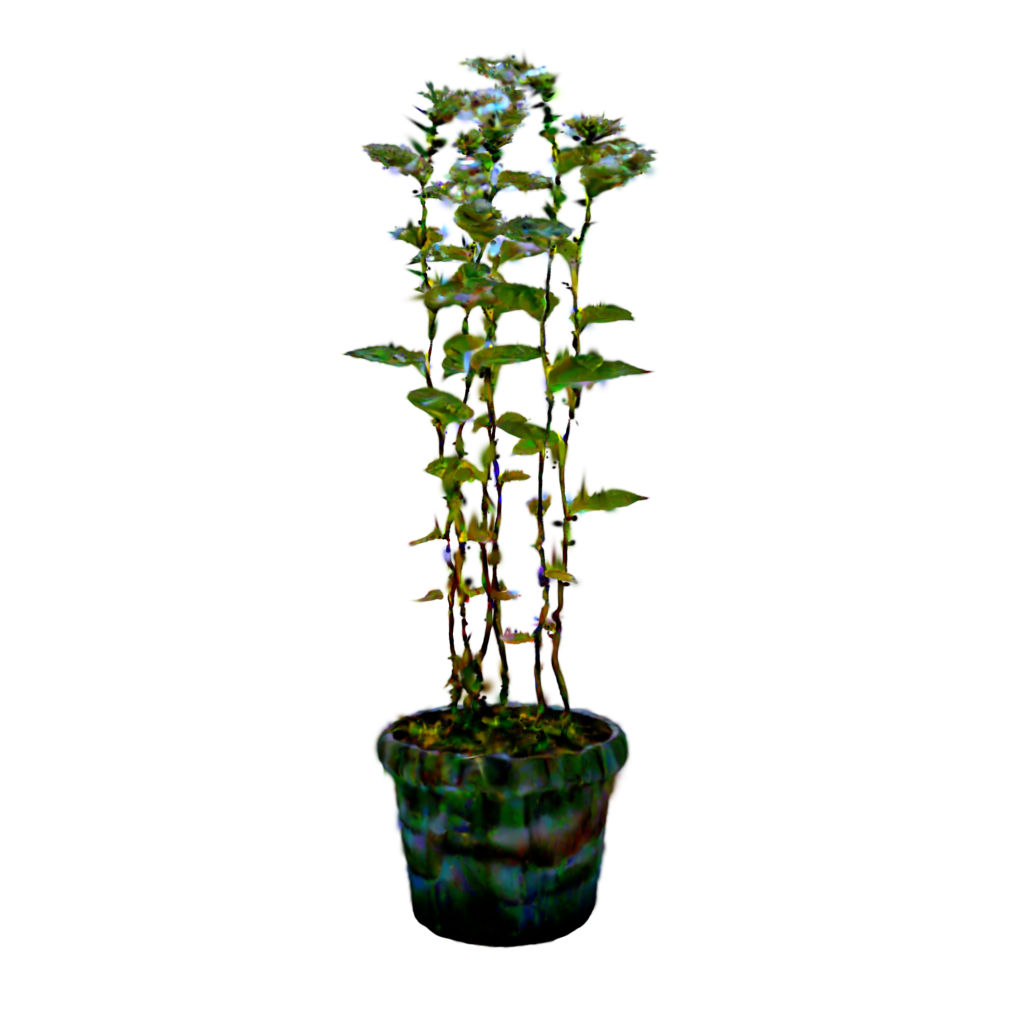} & \includegraphics[width=0.15\textwidth]{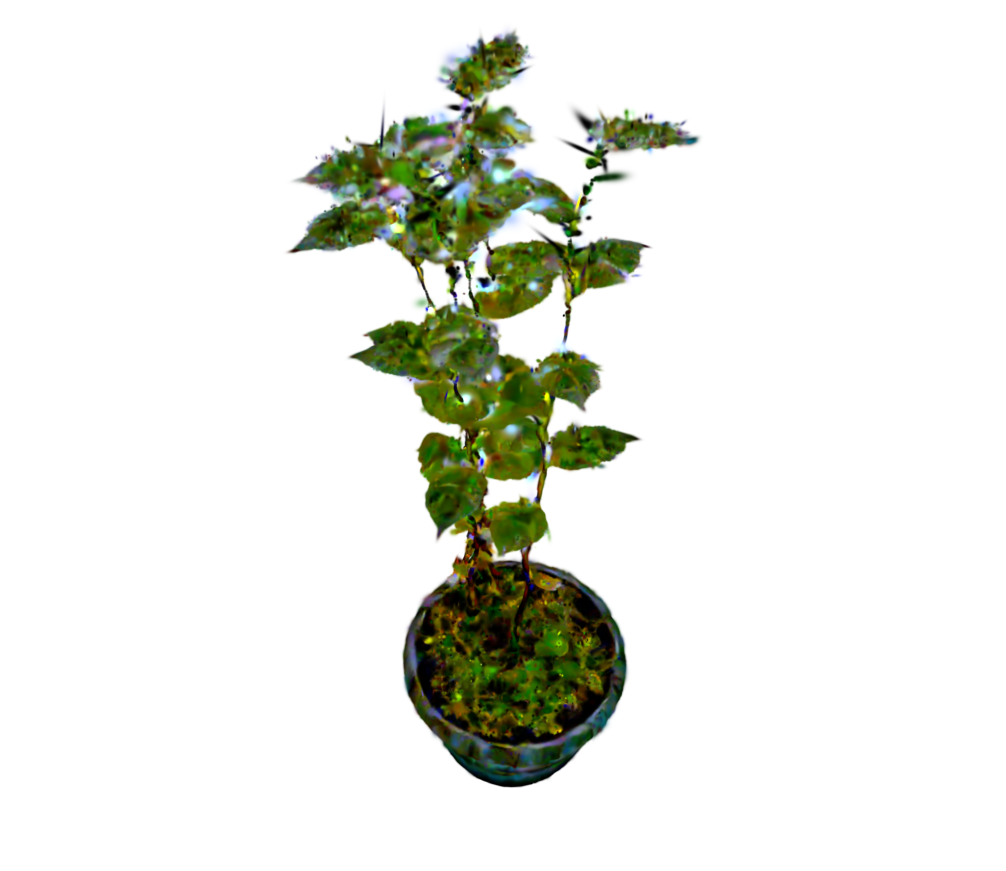}  \\ 

\end{tabular}

\caption{Visual comparison between rendered outputs for two selected synthetic bean, kale and mint plants for Latent-NeRF, Magic3D, Fantasia3D models, GaussianDreamer and PlantDreamer. Each model was initialised from their native methods.}
\label{fig:Synthetic_Comparison} 
\end{figure*}

\subsection{Capturing Real Plant Point Clouds}

For each plant instance, a set of 300 images were captured around the specimen using the approach outlined in Stuart et al. (2025) \cite{lewis2024fideltiy}.
Initial point clouds were produced using the original 3DGS model \cite{kerbl20233d}, which were trained on the captured images and then converted into dense point clouds using the 3DGS-to-PC framework \cite{stuart20253dgstopcconvert3dgaussian}. Each 3DGS representation was converted into approximately one million points to ensure dense coverage, and then manually cleaned to ensure an accurate ground truth. \newline
We also generated alternative point cloud representations via Multi-View Stereo (MVS) \cite{seitz2006comparison}  and Structure-from-Motion (SfM) \cite{iglhaut2019structure} using the COLMAP framework \cite{schoenberger2016sfm}. These point clouds exhibit higher levels of noise, and are utilised in Section \ref{sec:ablation} to evaluate the impact of noisy input point clouds on performance. 

\subsection{Generating Synthetic Plants using L-Systems}
For our synthetic experiments, we utilised Blender modeling software~\cite{Blender} to generate 3D L-system meshes of bean, mint and kale plants. Each synthetic plant species was designed based on observation of its real-world counterpart, with stochastic variations introduced within the L-System grammar, enabling diversity across individual plant instances. The plant structures were modeled such that stems were generated procedurally using L-System rules, while the leaf models were manually crafted in Blender and attached to the stems automatically based on the generated L-System, which created an authentic plant structure. Although these synthetic plants provide plausible recreations of their respective species, fine details were intentionally omitted to allow PlantDreamer to refine them during the generation process. \newline
Each element of the plants were densely subdivided to generate a uniform distribution of vertices throughout the 3D structure. These vertices were then assigned basic colour values - green for leaves, brown for soil - which served as the starting point for the 3DGS initialisation. We evaluate how altering the colours of these vertices impacts the final generated model in Section \ref{sec:ablation}.

\subsection{Implementation Details}
PlantDreamer was executed for 3600 epochs with a batch size of 1. The learning rate was set to 0.00005 and 0.0125 for the position and spherical harmonics respectively. The noise weighting function was adjusted dynamically across training ranging between: 0.2-0.98 for epochs below 600, 0.12–0.35 between epochs 600 and 1000, 0.12-0.25 between epochs 1000 and 2000, and 0.075–0.15 for the remaining epochs. The threshold for the large Gaussian culling algorithm was set to 3. \newline 
We evaluated PlantDreamer against the state-of-the-art text-to-3D models Latent-NeRF \cite{metzer2023latent}, Magic3D \cite{lin2023magic3d}, Fantasia3D \cite{chen2023fantasia3d} and GaussianDreamer \cite{yi2023gaussiandreamer}. Each of these models were executed using default parameter settings. We used the following text prompt for each model: "a [species] plant in a pot," where [species] was replaced with the respective plant type. To ensure consistency, we generated five instances per species using these models with random seeds. Experiments were conducted on an NVIDIA RTX 4090 Ti GPU, and we recorded the time required to generate each model. \newline
Each text-to-3D model was initialised using its native generation process for the synthetic plants. While for the real plants, GaussianDreamer and PlantDreamer were initalised with the real point clouds. We noticed that GaussianDreamer would often generate large, erroneous Gaussians that occluded the camera from some angles, leading to artificially low evaluation scores. Hence, we included our large Gaussian culling algorithm with a threshold value of 3 to ensure a fair comparison.
\section{Evaluation}
\label{sec:evaluation}

\begin{table*}
  \centering
  \small
    \renewcommand{\arraystretch}{1.1}
  \begin{tabular}{c|c|cc|cc|cc|c}
     & & \multicolumn{2}{c|}{Bean} & \multicolumn{2}{c|}{Kale} & \multicolumn{2}{c|}{Mint} & Time\\
    Initialisation \ & Model &  Quality  &  Alignment  &  Quality  &  Alignment  &  Quality  \vspace{0.05cm} & 
  Alignment & (Approx) \\
    \midrule
    \multirow{6}{*}{Synthetic} & \ 
     Latent-NeRF & 24.87 & 70.0 & 17.03 & 35.0 & 28.17 & \textbf{75.0} & 3 hours\\
    & Magic3D & 27.91 & 45.0 & 36.04 & \textbf{50.0} & 47.87  & 10.0 & 1.5 hours\\
    & Fantasia3D & 15.84 & 38.89 & 22.95 & 41.67 & 33.59 & 31.25 & 1 hour\\
    & GaussianDreamer & 41.27 & 5.00 &  \textbf{50.92} & \textbf{50.00} & 50.73 & 45.83 & \textbf{30 minutes} \\
    & PlantDreamer & \textbf{48.88} & \textbf{75.0} & 38.97 & 33.33 & \textbf{52.98}  & 45.83 & \textbf{30 minutes}\\
     \midrule
     \multirow{2}{*}{Real} & GaussianDreamer & 46.50  & 60.71 &  \textbf{58.94} & \textbf{75.0} & 51.23  & \textbf{50.00} & \textbf{30 minutes} \\
     & PlantDreamer &  \textbf{49.30} & \textbf{68.75} & 50.10 & 66.67 & \textbf{57.64} & 46.53 & \textbf{30 minutes} \\

     
  \end{tabular}
  \caption{T3 Bench Quality and Alignment scores for bean, kale and mint plants generated using  Latent-NeRF, Magic3D, Fantasia3D, GaussianDreamer and PlantDreamer. For synthetic initialisation each model used its native method. For real initialisation, PlantDreamer and GaussianDreamer used real point clouds. Scores are averaged across all instances per species, with average run time also recorded.}
  \label{tab:t3andtimecomparison}
\end{table*}

\begin{table}
  \centering
\renewcommand{\arraystretch}{1.3}
  \begin{tabular}{c|ccc}
    & \multicolumn{3}{c}{PSNR Masked (db)} \\
    Model & \enspace Bean \enspace & \enspace Kale \enspace & \enspace Mint \enspace  \\
    \midrule
    GaussianDreamer & 10.88 & 11.28 & 10.86  \\ 
    PlantDreamer & \textbf{16.23} & \textbf{17.20} & \textbf{14.93}   
  \end{tabular}
  \caption{The average PSNR Masked result across the all real plants for both PlantDreamer and GaussianDreamer.}
  \label{tab:plantDreamervsGaussianDreamer}
\end{table}

We conducted a comparative evaluation of 3D models generated from the real and synthetic plant datasets. Our evaluation of PlantDreamer focused on two key aspects: the 3D quality of synthetic and real plants compared to state-of-the-art text-to-3D models, and the fidelity of 3DGS models when initialised with real point clouds compared to the captured ground truth.  

\subsection{3D Geometry Quality and Alignment}
We evaluate our method against state-of-the-art text-to-3D models using T3Bench~\cite{he2023t3bench}, a popular benchmark for scoring text-to-3D models. T3Bench evalutes models against the prompt used to generate them using two metrics, \textit{quality} and \textit{alignment to the prompt}. Quality scoring by T3Bench is achieved using text-image scoring models CLIP and ImageReward, while regional convolutions are used to ensure quality is uniform across all angles of the image. Alignment scoring captures a textual caption for the object from a range of views, and Large Language Models (LLM) including GPT-4 are used to evaluate these captions against the input prompt. These results are presented in Table \ref{tab:t3andtimecomparison} and renders are shown in Figure \ref{fig:Synthetic_Comparison}.\newline
Our model outperformed most text-to-3D approaches and performed comparably to GaussianDreamer. This was largely due to its capability to generate plant structures that remained consistent across different views with coherent texturing. Additionally, our model reliably generated all elements of the prompt, particularly the plant pot, which many other text-to-3D models struggled to produce.\newline
PlantDreamer outperformed GaussianDreamer for bean and mint plants, likely because GaussianDreamer struggled reconstructing complex plant structures, often generating a disorganised collection of leaves without a pot. However, GaussianDreamer performed better for Kale, which we attribute to our L-System meshes for Kale containing fewer leaves relative to the sizable pot. This affected the performance when evaluating against CLIP embeddings and LLMs, which are trained to associated large leaves with kale. This highlights the considerable effect that the L-System grammar has on the final plant model.

\begin{table*}
    \small
  \centering
\renewcommand{\arraystretch}{1.1}
  \begin{tabular}{c|ccc|cc}
     & \multicolumn{3}{c|}{Real} & \multicolumn{2}{c}{Synthetic} \\
    \enspace Colour \enspace & \enspace PSNR Masked (db) \enspace & \enspace Quality \enspace & \enspace Alignment \enspace & \enspace Quality \enspace & \enspace Alignment \enspace\\
    \midrule
     Original & \textbf{16.12} & 51.80 & \textbf{61.90} & \textbf{48.64} & \textbf{60.71}  \\
      Noise & 10.63 & \textbf{52.69} & 60.71 & 46.95 & 33.33 \\
      Black & 14.96 & 43.87 & 47.62 & 38.27 & 39.29 \\
      White & 3.41 & 32.19 & 23.81 & 26.55 & 18.75 \\
  \end{tabular}
  \caption{PSNR Masked and T3 scores, averaged over all synthetic and real models, for normal, black, white and noised point clouds.}
  \label{tab:colourInit}
\end{table*}

\subsection{Textural and Structural Realism}

To evaluate the real plant data, we initialised PlantDreamer and GaussianDreamer with identical point clouds, enabling a direct comparison between the two approaches. Since we captured real images of the initialised plant, we could directly compare the rendered 3DGS outputs to the corresponding images.
Rendered images that are similar to the ground truth imply higher levels of realism for the texture and structure of the plant. To quantify fidelity, we computed the Peak Signal-to-Noise Ratio (PSNR) masked score for each plant, which measures the pixel intensity differences using the mean squared error formula, applied exclusively to plant pixels to exclude background interference. Higher PSNR values indicate greater similarity between the synthetic and real images, with results presented in Table \ref{tab:plantDreamervsGaussianDreamer}. \newline
Our results show that PlantDreamer outperformed GaussianDreamer across all plant species, producing higher PSNR masked scores and more realistic 3D representations. As highlighted in Figure \ref{fig:RealPlantComparison}, PlantDreamer can replicate the nuanced textures of plant structures, particularly around leaf surfaces, which GaussianDreamer struggles to achieve. This is clearest in the bean plants, where GaussianDreamer's leaves appear over-saturated and lacking rigid detailing.  Additionally, GaussianDreamer would distort the plant morphology, leading to unrealistic structure expansion and loss of fine details; for example, in the mint plants, individual stems were merged into an unrealistic amalgamation of foliage. \newline
Overall, PlantDreamer generated more realistic and faithful 3DGS representations, primarily due to the geometry constraints imposed by the ControlNet and enhanced detail provided by the LoRA.

\subsection{Computation Performance}

PlantDreamer demonstrated competitive performance compared to other text-to-3D models, as shown in Table\ref{tab:t3andtimecomparison}.
Its runtime was similar to GaussianDreamer, since both share a similar optimisation pipeline, with ControlNet and LoRA introduing minimal computation overhead.

\section{Ablation Studies}
\label{sec:ablation}

\begin{table}
  \centering
  \begin{tabular}{c|ccc}
  & \multicolumn{3}{c}{Real} \\
    Method & PSNR Masked (db) & Quality & Alignment\\
    \midrule
    3DGS & \textbf{16.12} & \textbf{51.80} & \textbf{61.90} \\
    MVS & 8.71 & 47.38 & 59.52 \\
    SfM & 10.72 & 18.75 & 22.62 \\
  \end{tabular}
  \caption{PSNR Masked and T3 scores, averaged over all real PlantDreamer models, for the 3DGS, MVS and SfM initial point clouds.}
  \label{tab:reconstructionAlternatives}
\end{table}

While PlantDreamer is capable of producing high-quality models from an initial point cloud, our experiments used dense, coloured and accurate point clouds to ensure effective 3DGS initialisation. However, \textit{real world} captured data is often inaccurate and can contain erroneously coloured points. Therefore, we conducted ablation studies to analyse the impact of erroneous point clouds on plant structure and texture. A visual showcase of the rendered outputs of this study can be found in section \ref{sec:ablation_data} of the supplementary material. \newline
\textbf{Impact of Noise and Sparsity}. To evaluate PlantDreamer’s robustness to noise and structural inaccuracies, we tested using point clouds generated using MVS and SfM, both of which are typically less accurate and noisier than 3DGS reconstructions.
As shown by the decrease in PSNR Masked and T3 Bench scores in Table \ref{tab:reconstructionAlternatives}, less accurate initial point clouds have a substantial impact on the final 3DGS model. Since the ControlNet penalises Gaussians that deviate from the initial structure, the model was unable to generate new Gaussians to counteract missing or incorrect geometry. SfM produced the lowest fidelity models, since poor initial geometry results in the diffusion process failing to converge effectively, which leads to anomalies in the final geometry. Interestingly, the PSNR masked metric is higher for SfM, likely because its larger, inaccurate Gaussians cover more mask area. \newline
\textbf{Impact of Point Colours}. To assess the impact of point colours, we experimented with three distinct variations: white, black, and noised point colours. For the white and black conditions, all colour channels were set to 255 and 0, respectively. The noised condition assigned each point a random colour value, between 0 and 255 for the three individual channels. As shown in Table \ref{tab:colourInit}, altering the initial colours of the point cloud resulted in a decrease in PSNR masked and T3 bench scores. We found that the initial colours had a large influence over the final point cloud, with black colours producing dark plants, and white colours producing saturated plants. Interestingly, noised point clouds produced  higher fidelity than the black and white plants, as randomised colours have reduced bias allowing the diffusion process to generate a more normalised texture. We foresee alterations in the initial point cloud being leveraged for controllable shading of the final output, such as facilitating simulations of different lighting environments.

\section{Limitations}
While PlantDreamer produces high-fidelity 3D models for specific plant species, it does not generalise across all plant types. Each species must be effectively constructed in the L-Systems grammar for valid initialisation, which inherently lacks the intricacies and natural variability of real-world plants. Furthermore, a dedicated LoRA is required for each species, although these one-time steps are uncomplicated to achieve for each new species. Future work will focus on extending PlantDreamer to support a range of plant types without relying on predefined species priors.

\section{Conclusion}
\label{sec:conclusion}
In this paper, we present PlantDreamer, a novel framework for generating high-fidelity 3D plant models from both synthetic L-System meshes and real-world point cloud data. Our innovative pipeline integrates a depth ControlNet, LoRA and large Gaussian culling, greatly improving the textural realism and morphological consistency of generated plants. To evaluate our approach, we captured an new dataset of real plant point clouds, and generated synthetic counterparts using L-Systems. Our results demonstrate that PlantDreamer performs better in both realism and cross-view consistency compared to state-of-the-art text-to-3D models for 3D plant generation. Finally, we conducted ablation studies, exploring how point cloud initialisation impacts 3D plant generation. Overall, PlantDreamer provides an efficient and powerful solution for generating 3D plant models, addressing a critical data gap and thereby enabling future advancements in 3D phenotyping research.

{
    \small
    \bibliographystyle{ieeenat_fullname}
    \bibliography{main}
}

\clearpage
\setcounter{page}{1}
\maketitlesupplementary

\section{More Ablation Study Data}
\label{sec:ablation_data}

\begin{figure*}[b] 
\centering
\setlength{\tabcolsep}{1.5pt}
\renewcommand{\arraystretch}{0.5}
\begin{tabular}{ccccccc} 

& \multicolumn{2}{c}{\textit{Bean Plant}} & \multicolumn{2}{c}{\textit{Kale Plant}}  & \multicolumn{2}{c}{\textit{Mint Plant}} \vspace{0.3cm}\\ 

\raisebox{3.5em}{\parbox{2.3cm}{\centering 3DGS }} &
\includegraphics[width=0.14\textwidth]{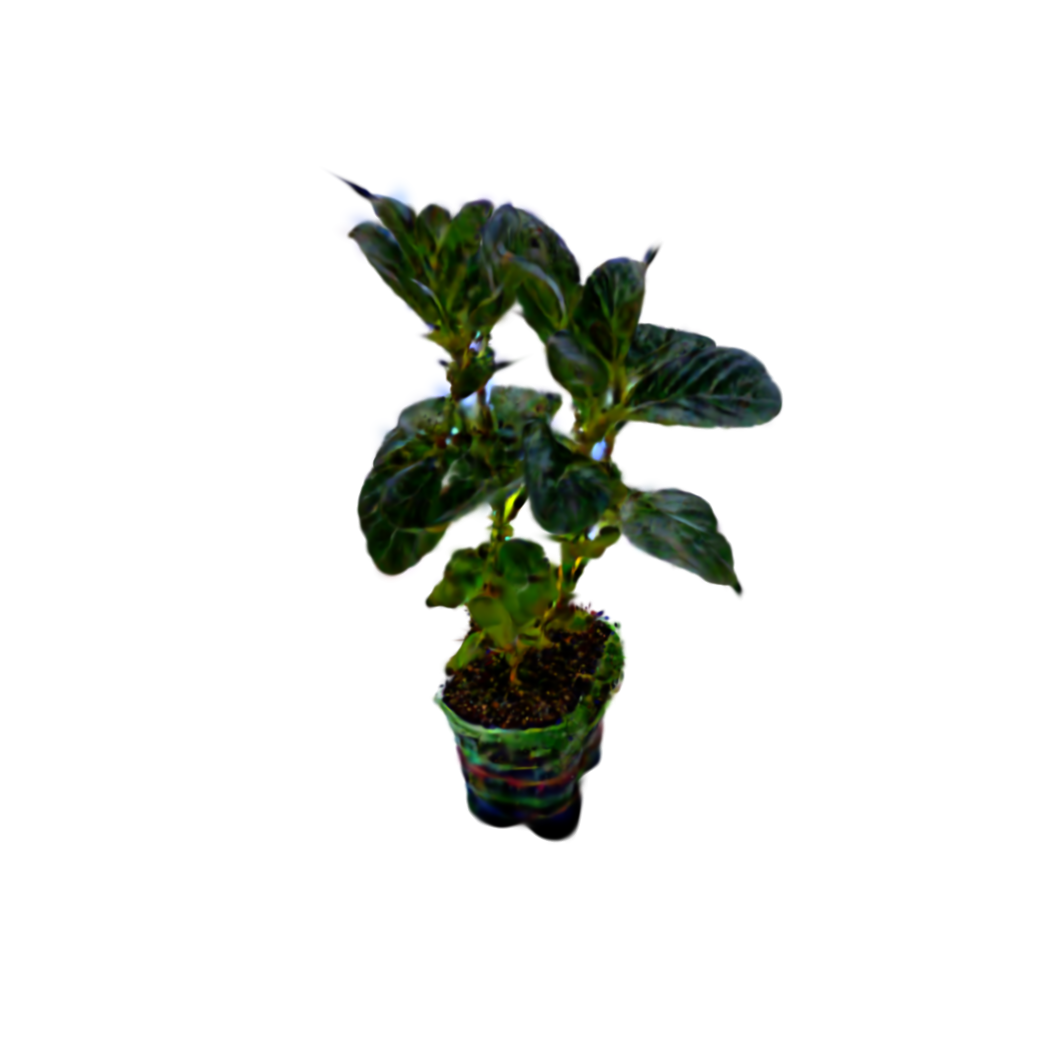} & 
\includegraphics[width=0.14\textwidth]{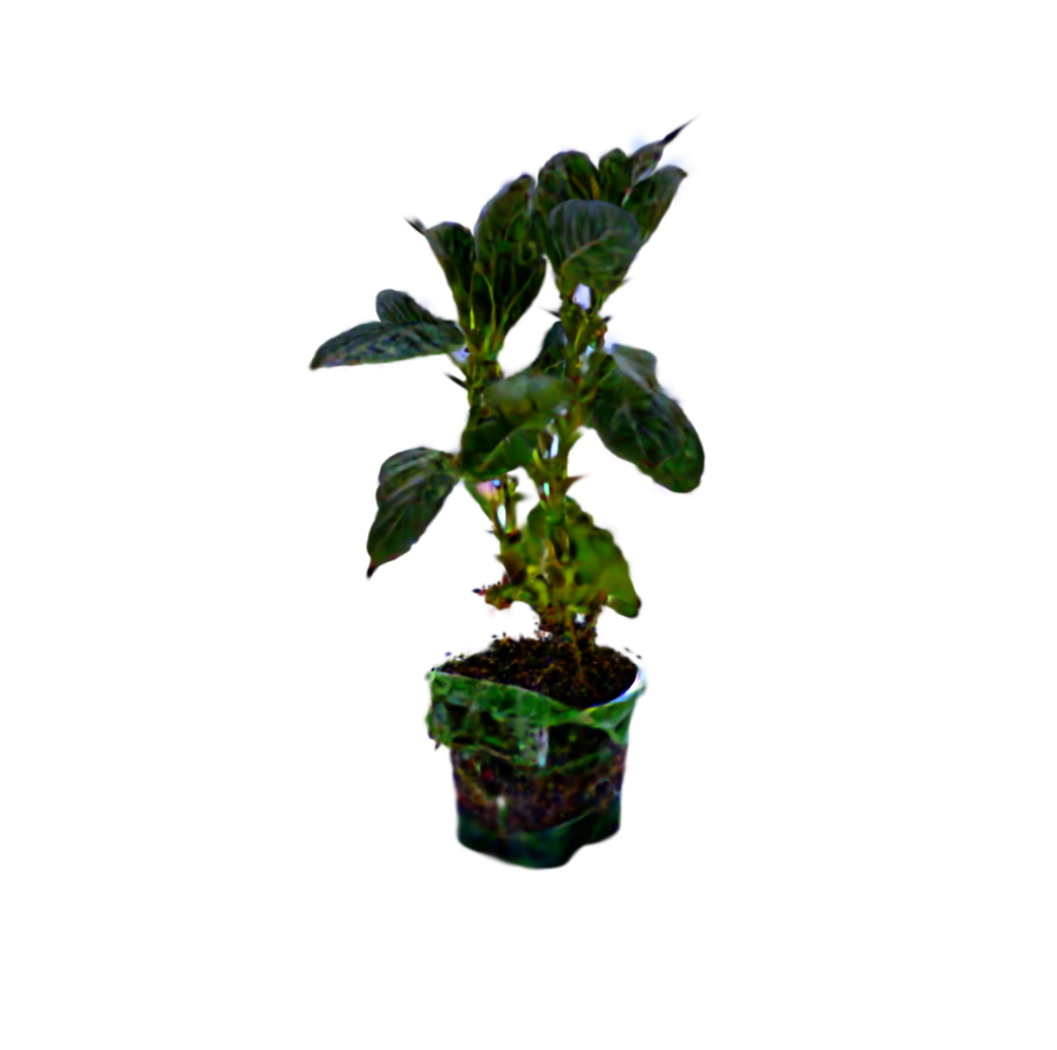} &\includegraphics[width=0.14\textwidth]{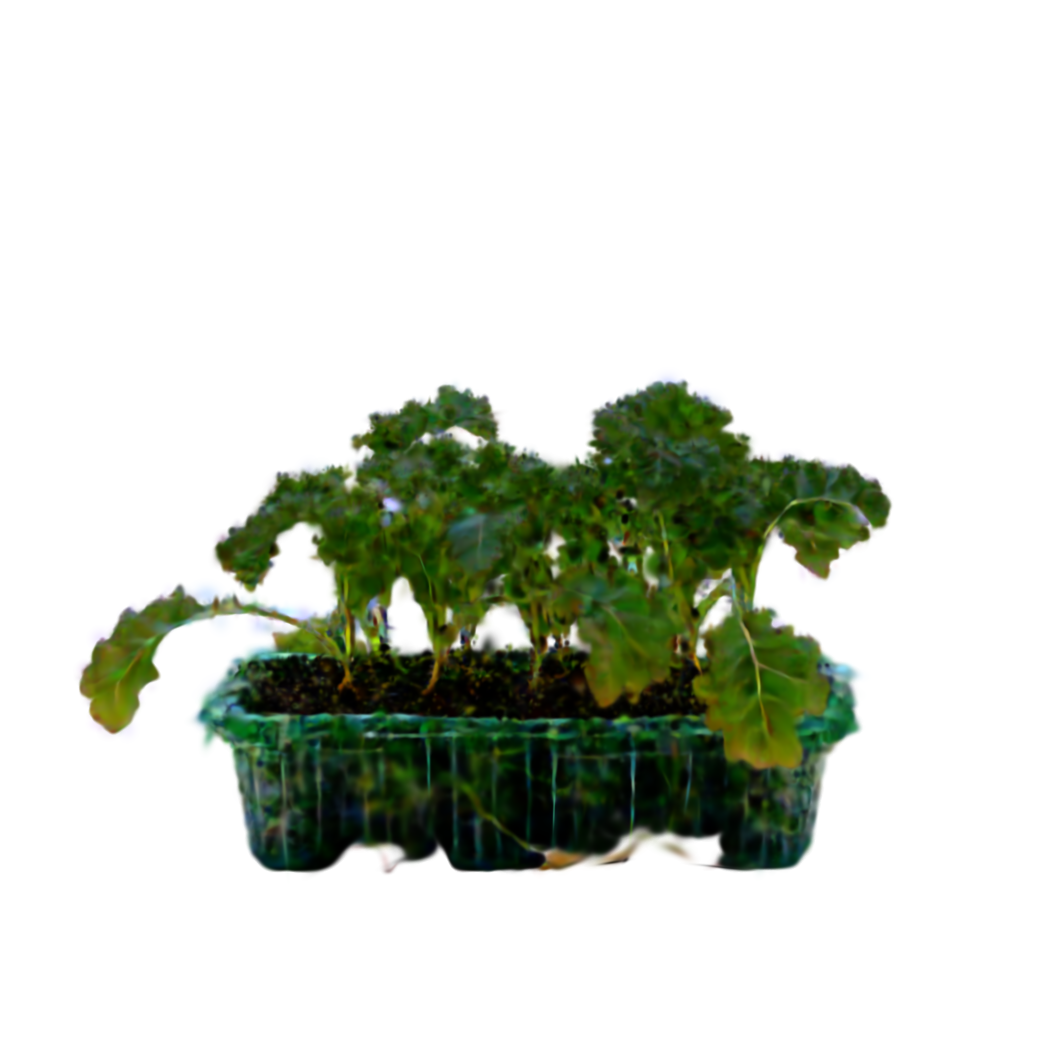} & \includegraphics[width=0.14\textwidth]{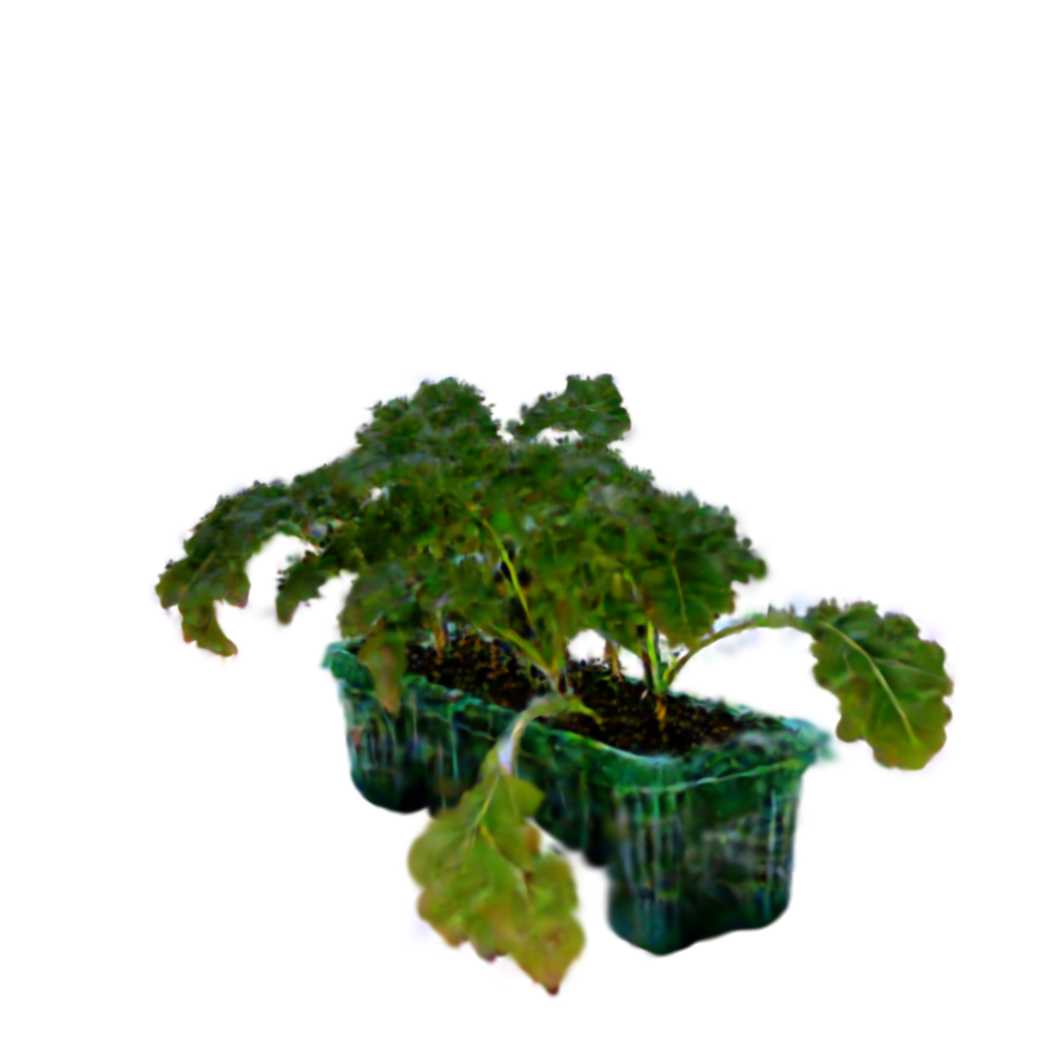} & \includegraphics[width=0.14\textwidth]{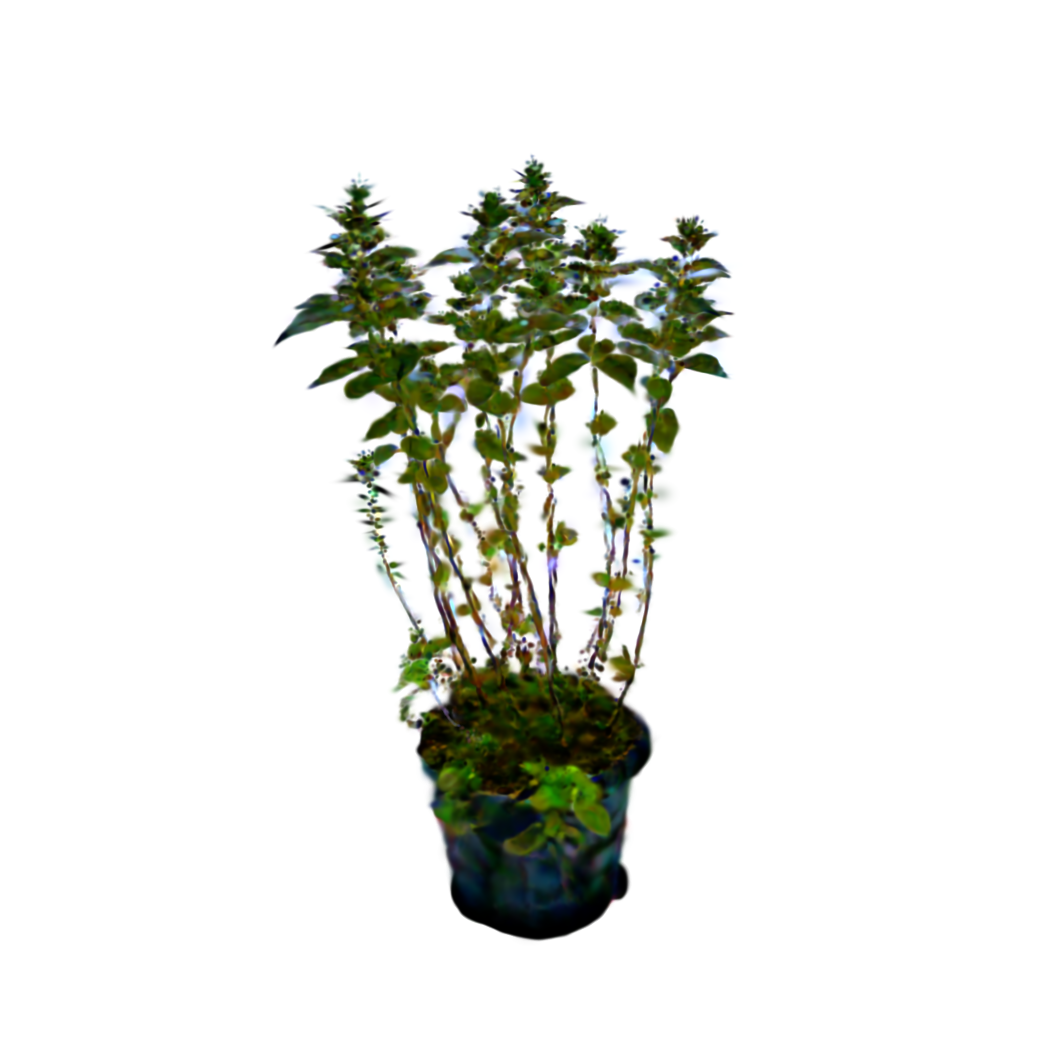} & \includegraphics[width=0.14\textwidth]{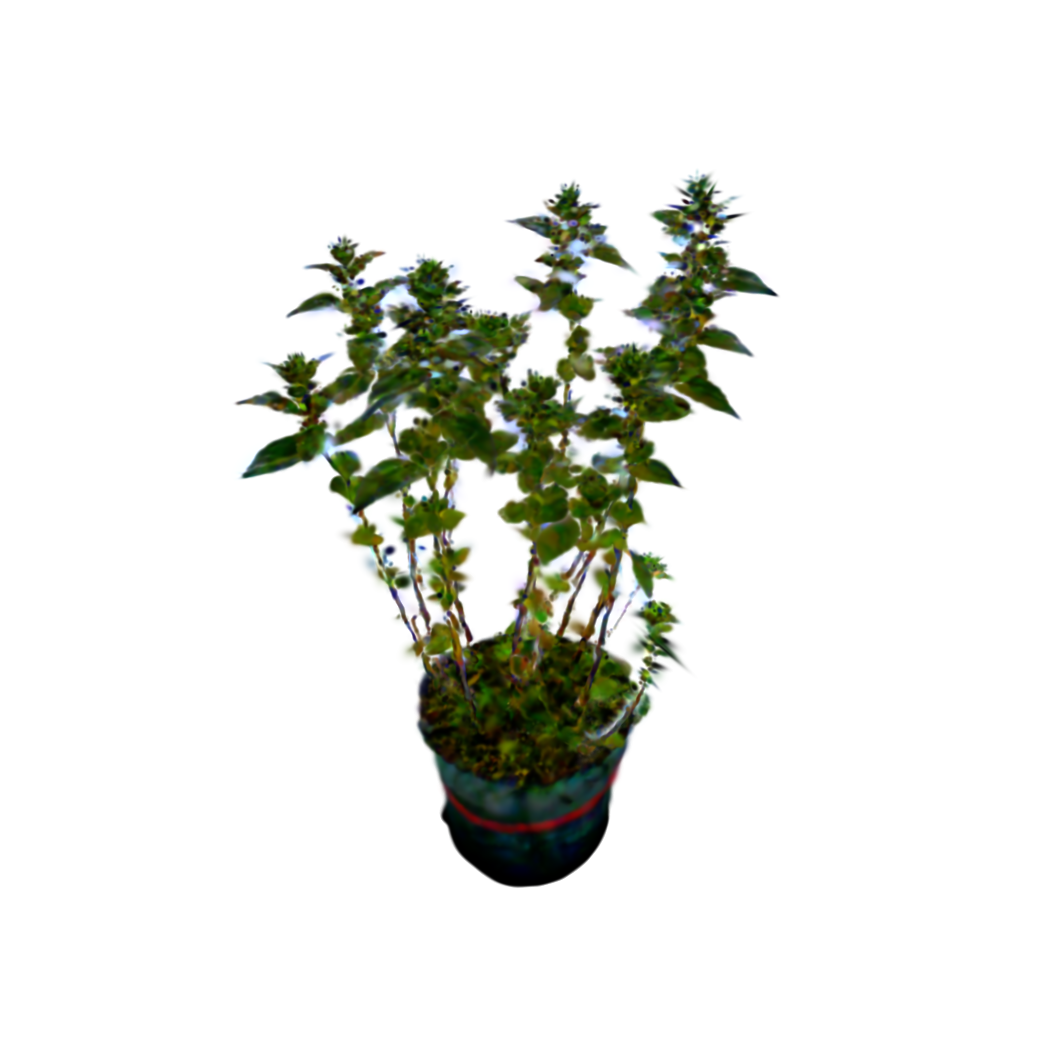} \\ 

\raisebox{3.5em}{\parbox{2.3cm}{\centering MVS }} & 
\includegraphics[width=0.14\textwidth]{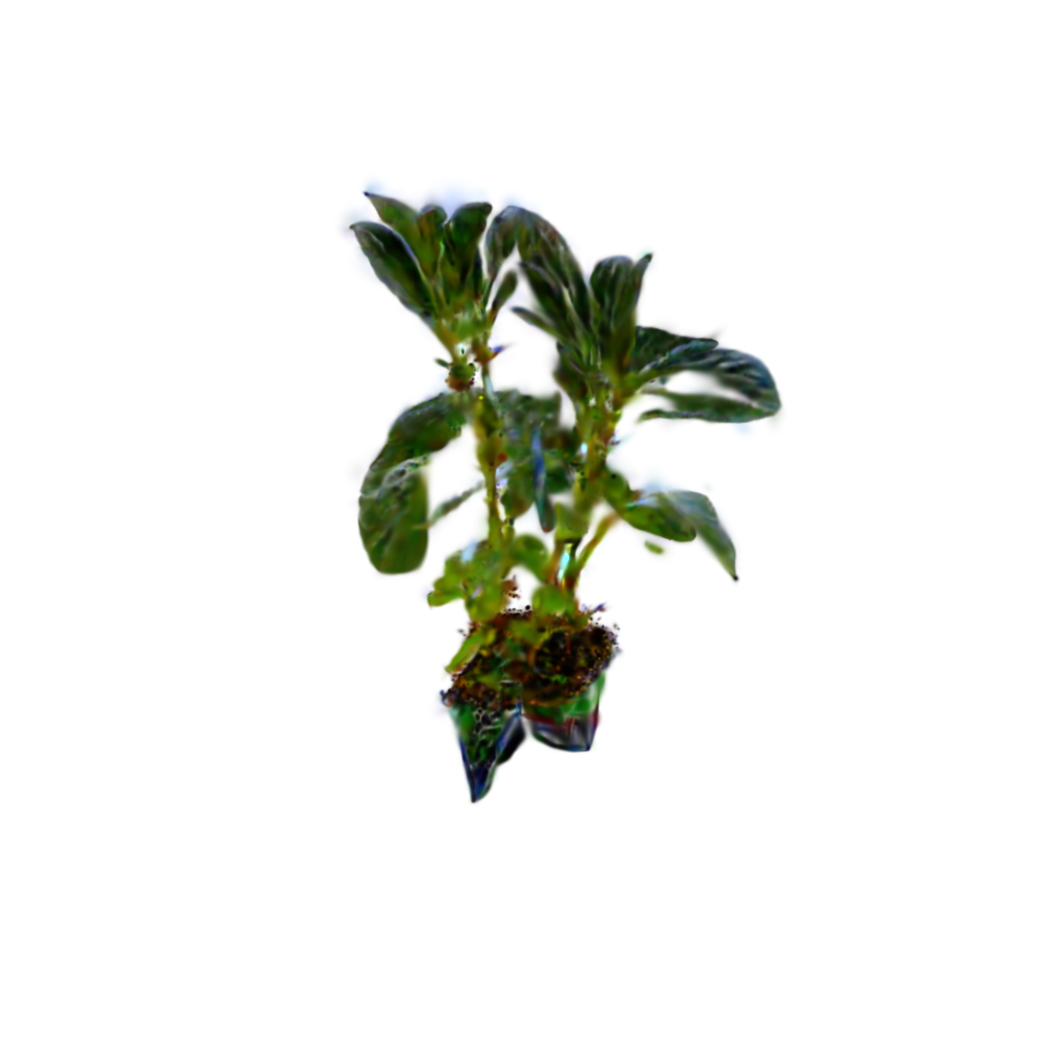} & 
\includegraphics[width=0.14\textwidth]{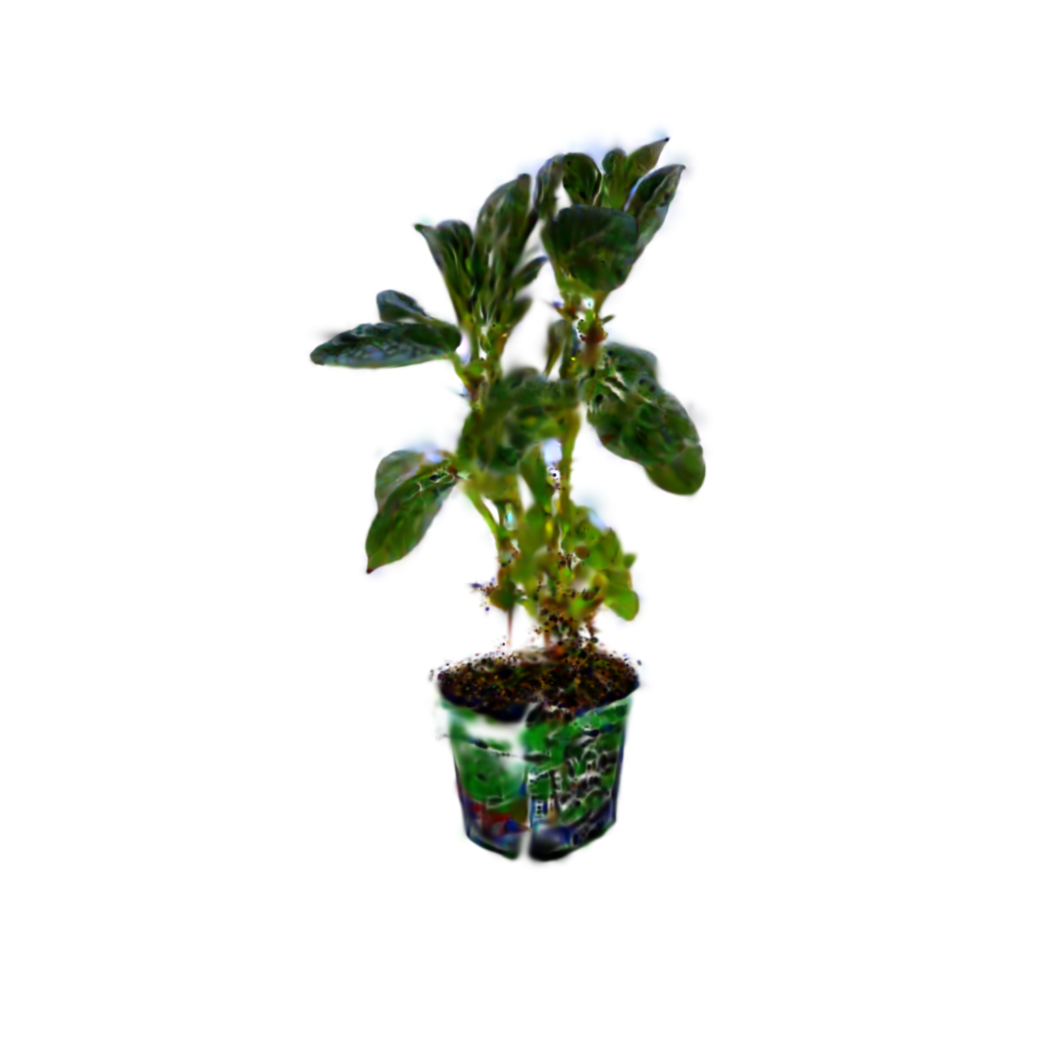} & \includegraphics[width=0.14\textwidth]{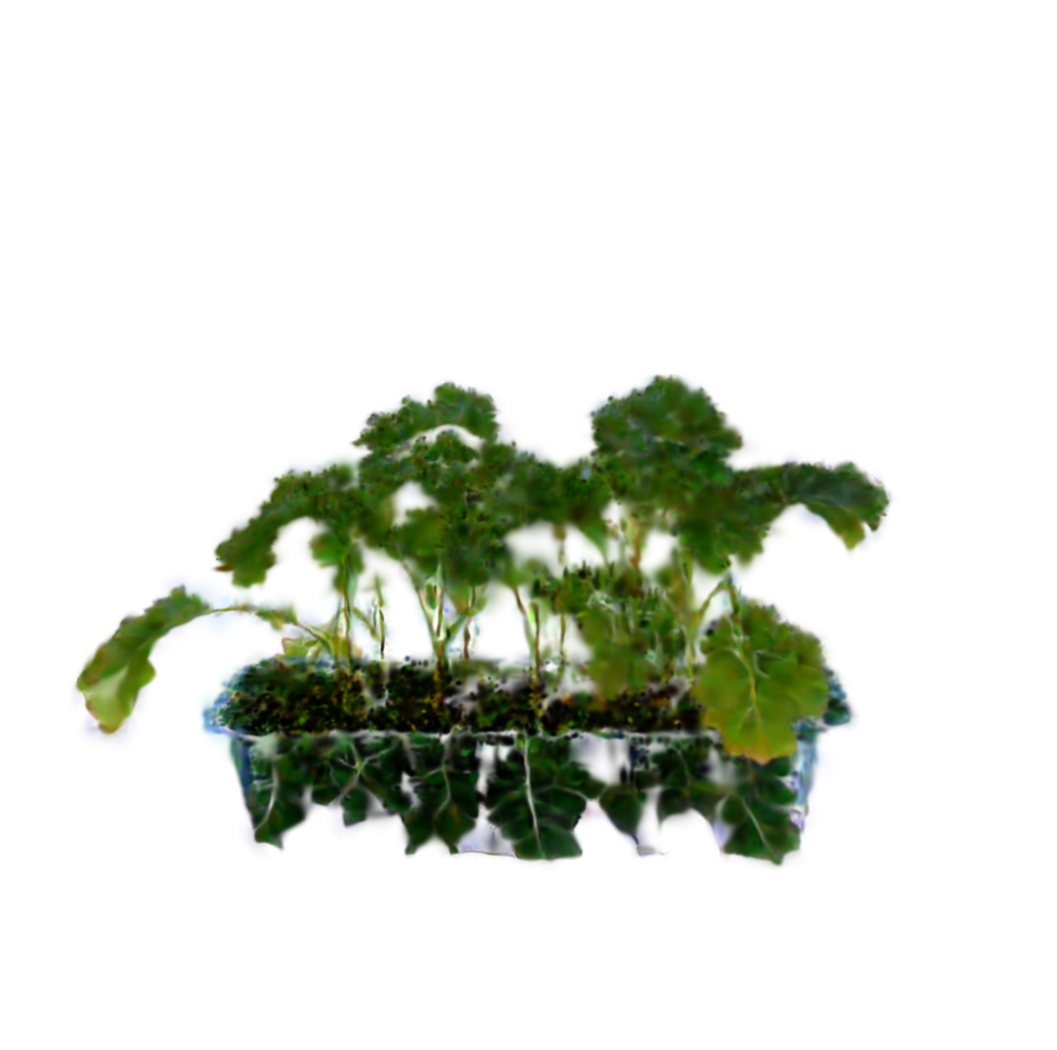} & \includegraphics[width=0.14\textwidth]{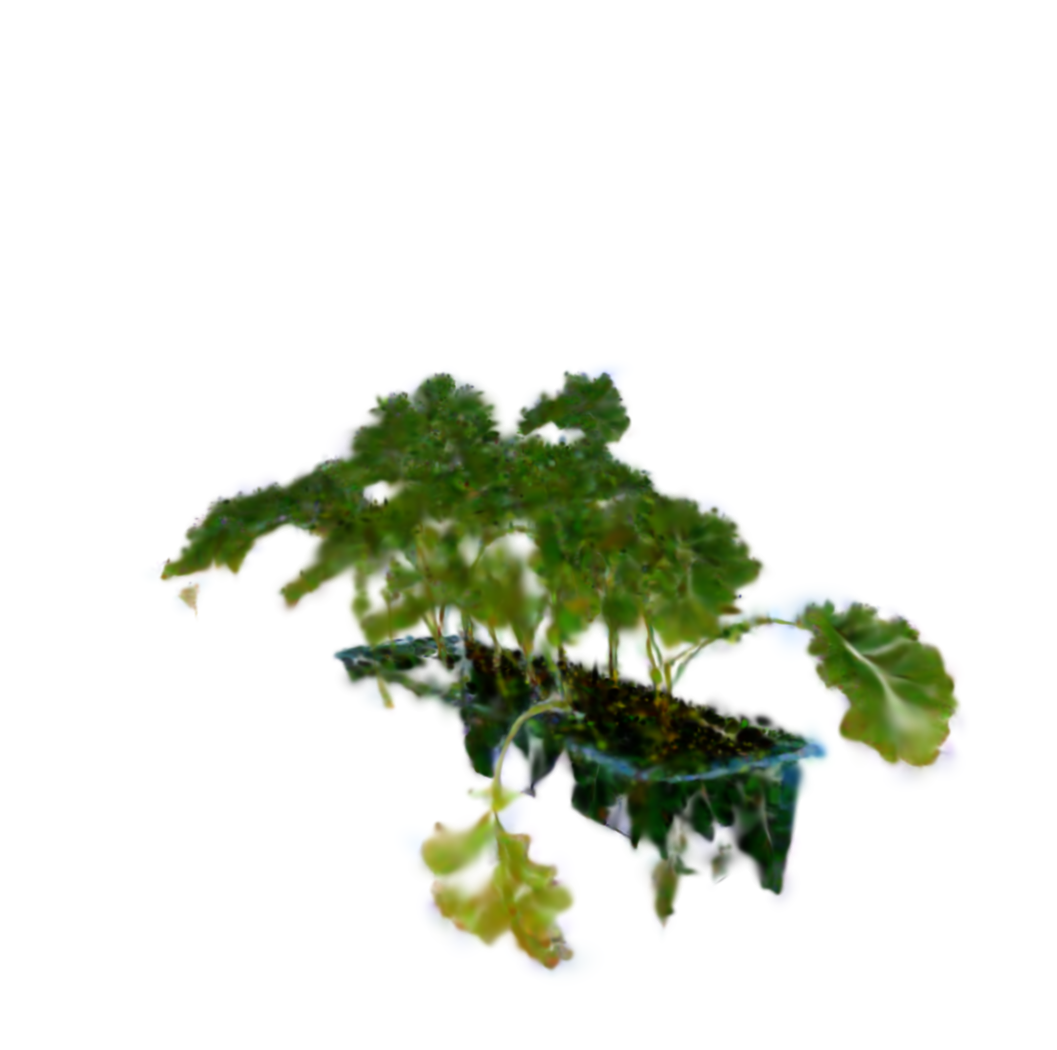} & \includegraphics[width=0.14\textwidth]{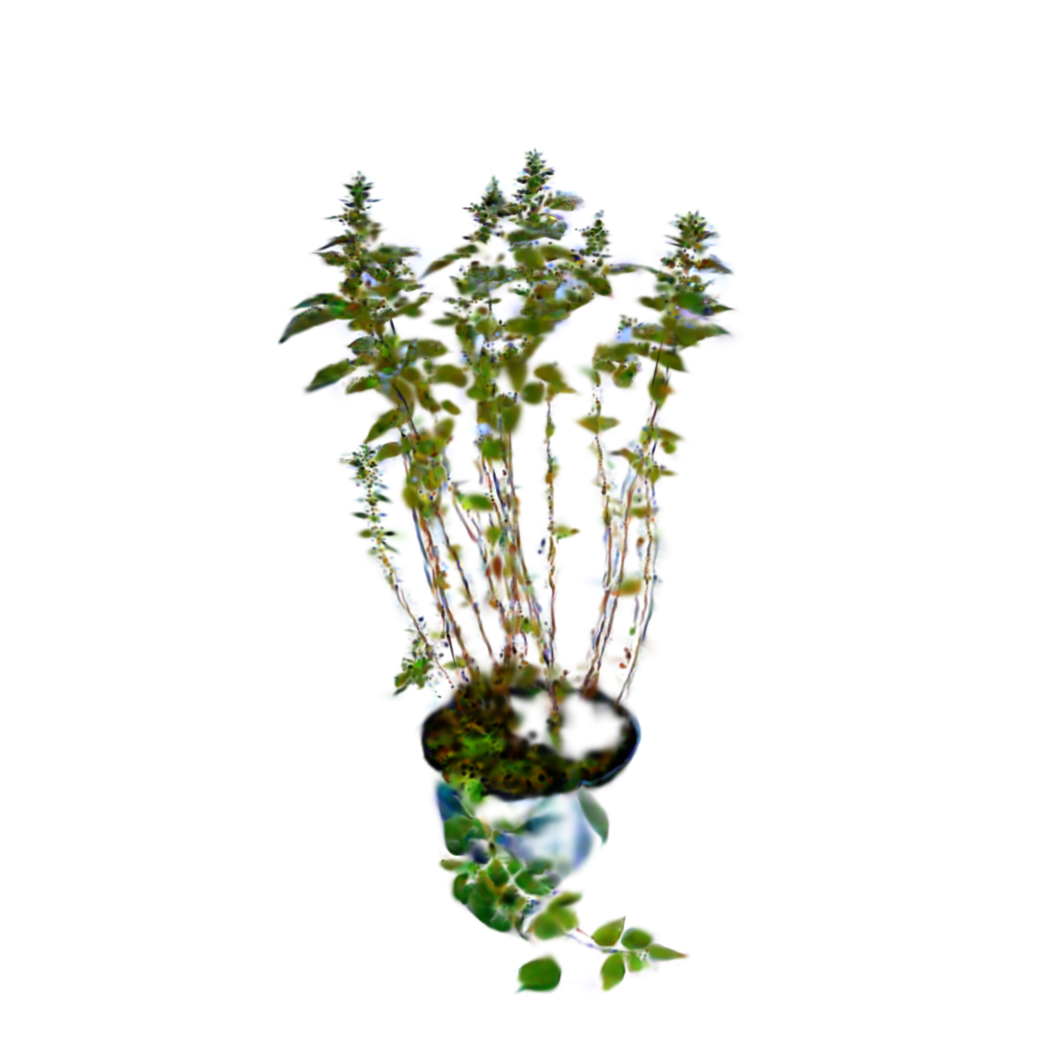} & \includegraphics[width=0.14\textwidth]{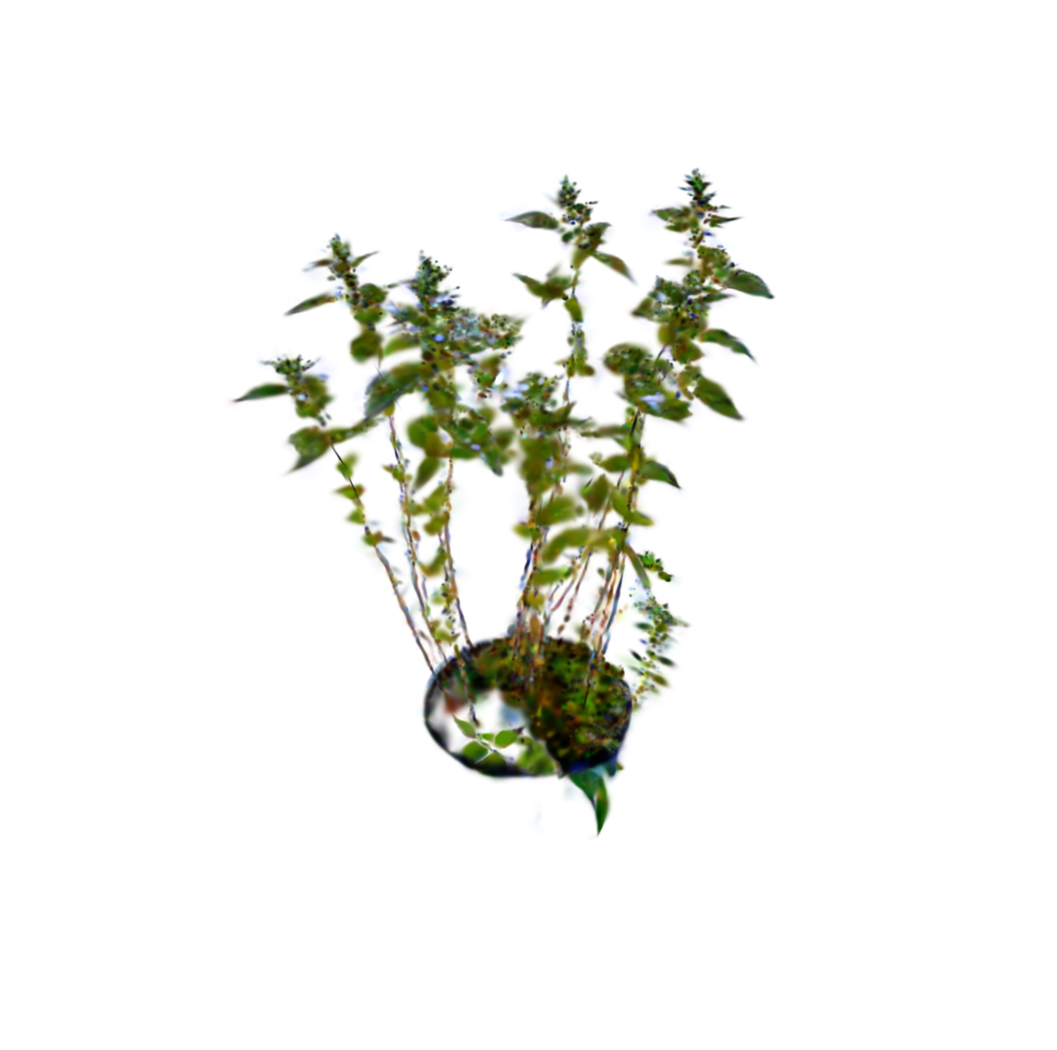} \\ 

\raisebox{3.5em}{\parbox{2.5cm}{\centering SfM }} & 
\includegraphics[width=0.14\textwidth]{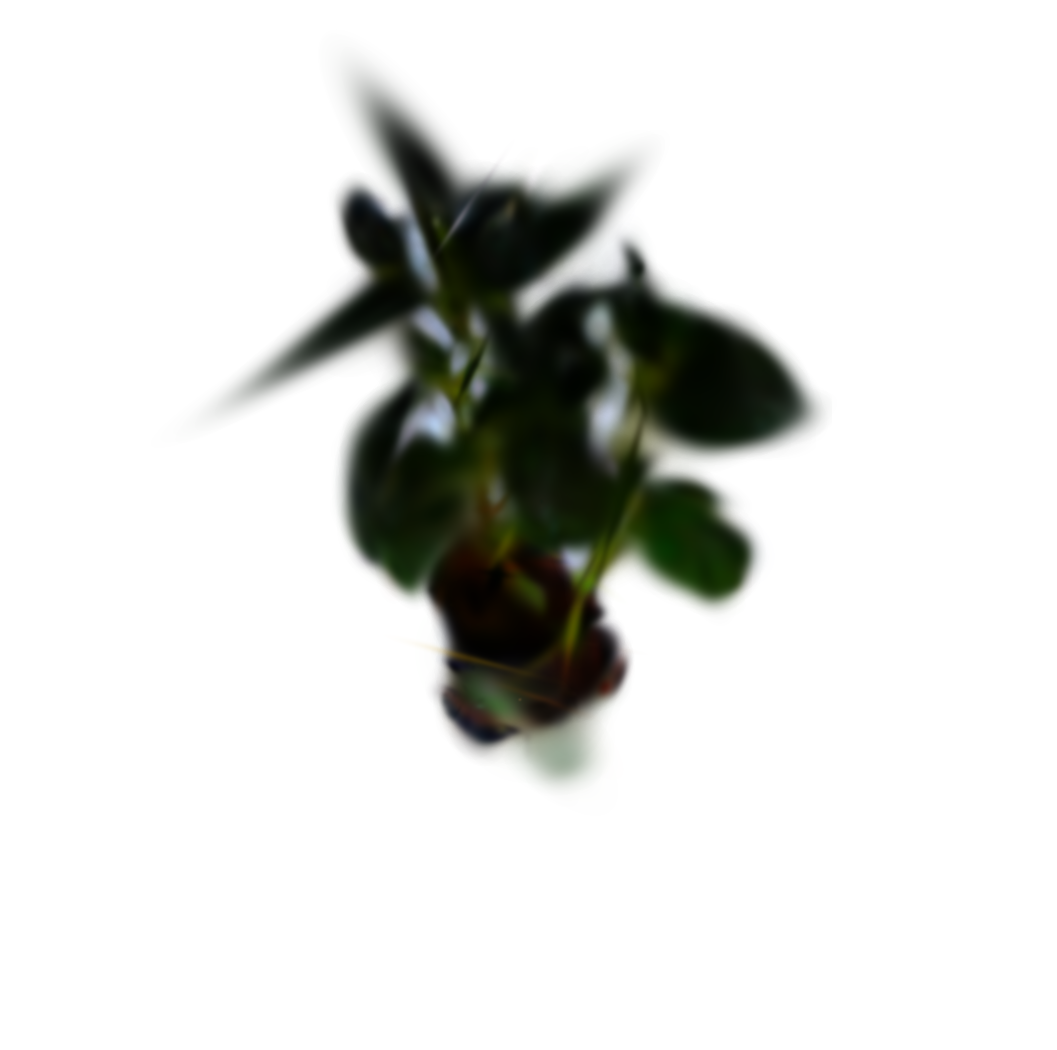} & 
\includegraphics[width=0.14\textwidth]{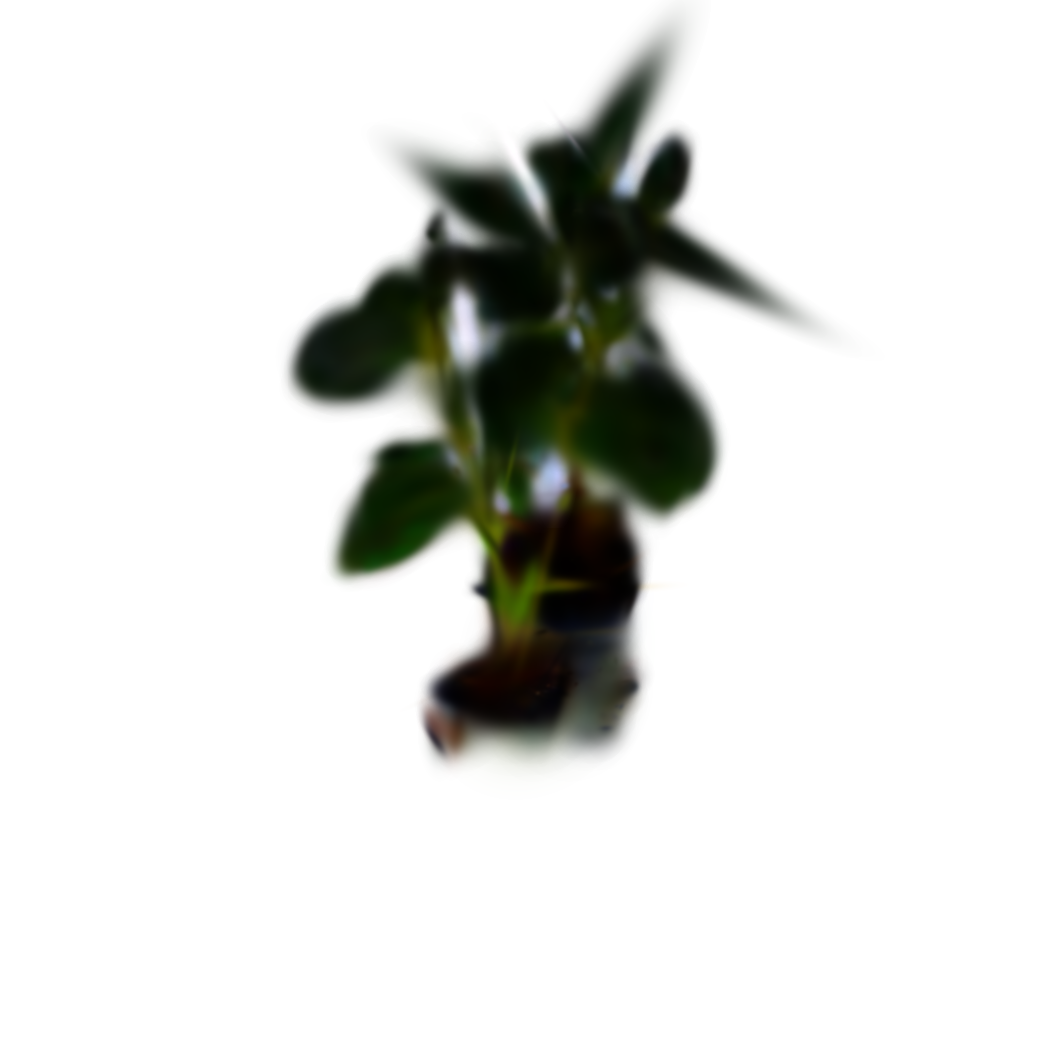} & \includegraphics[width=0.14\textwidth]{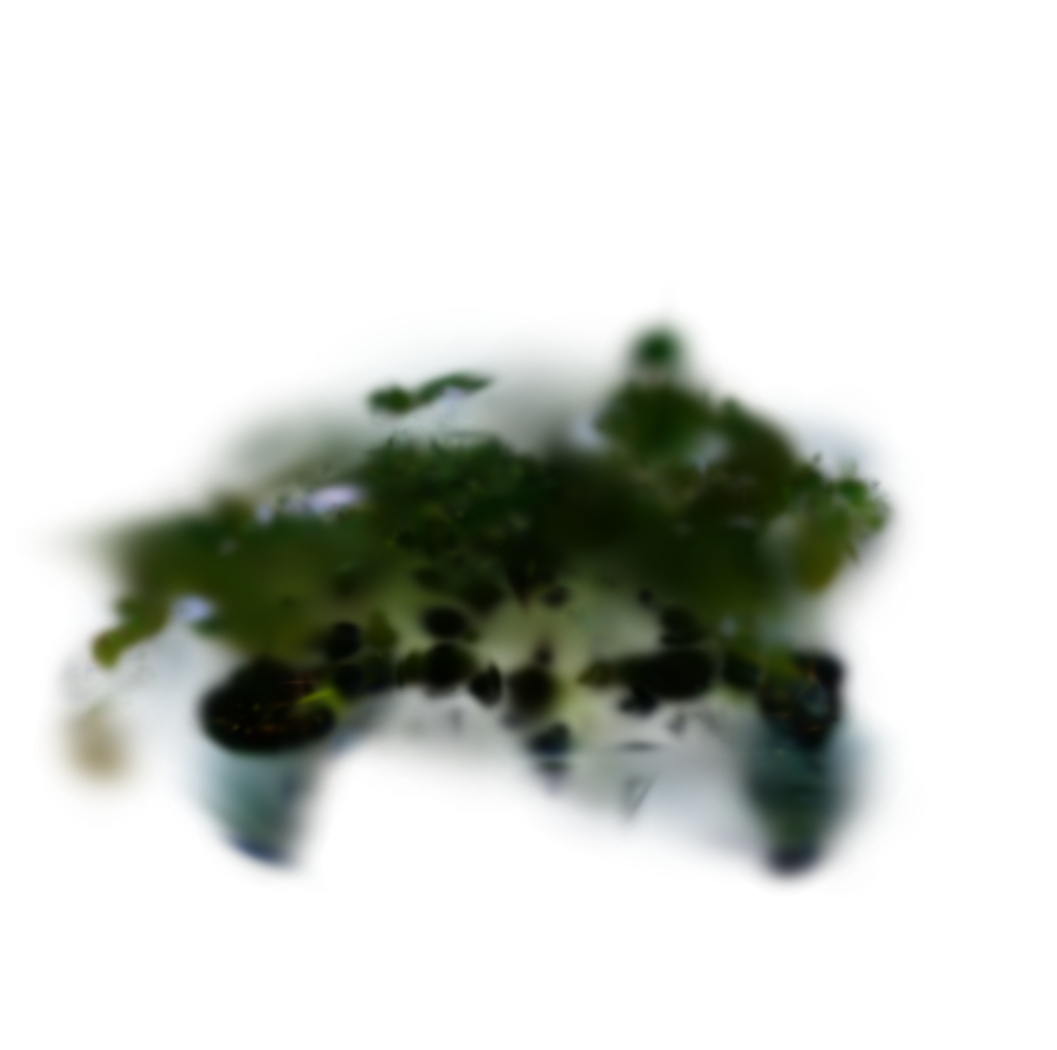} & \includegraphics[width=0.14\textwidth]{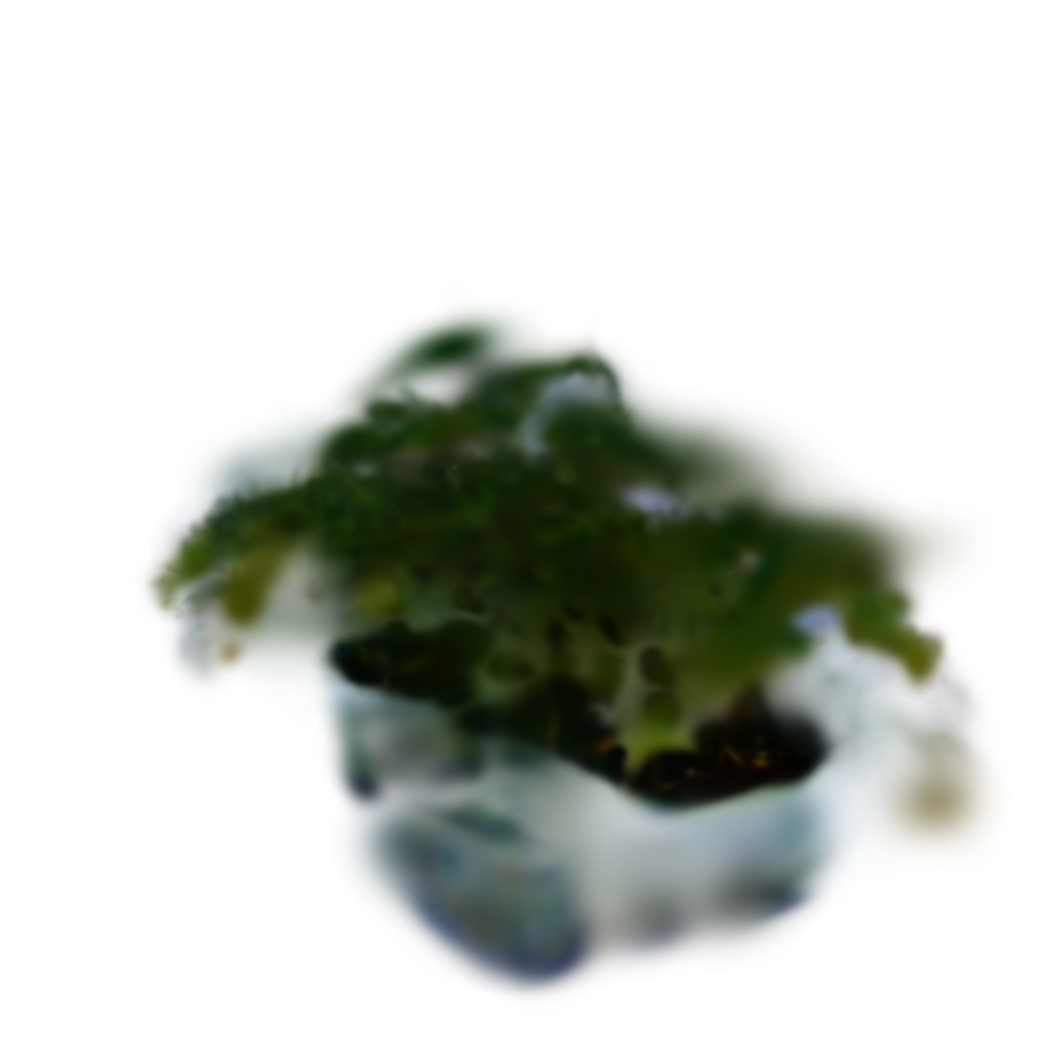} & \includegraphics[width=0.14\textwidth]{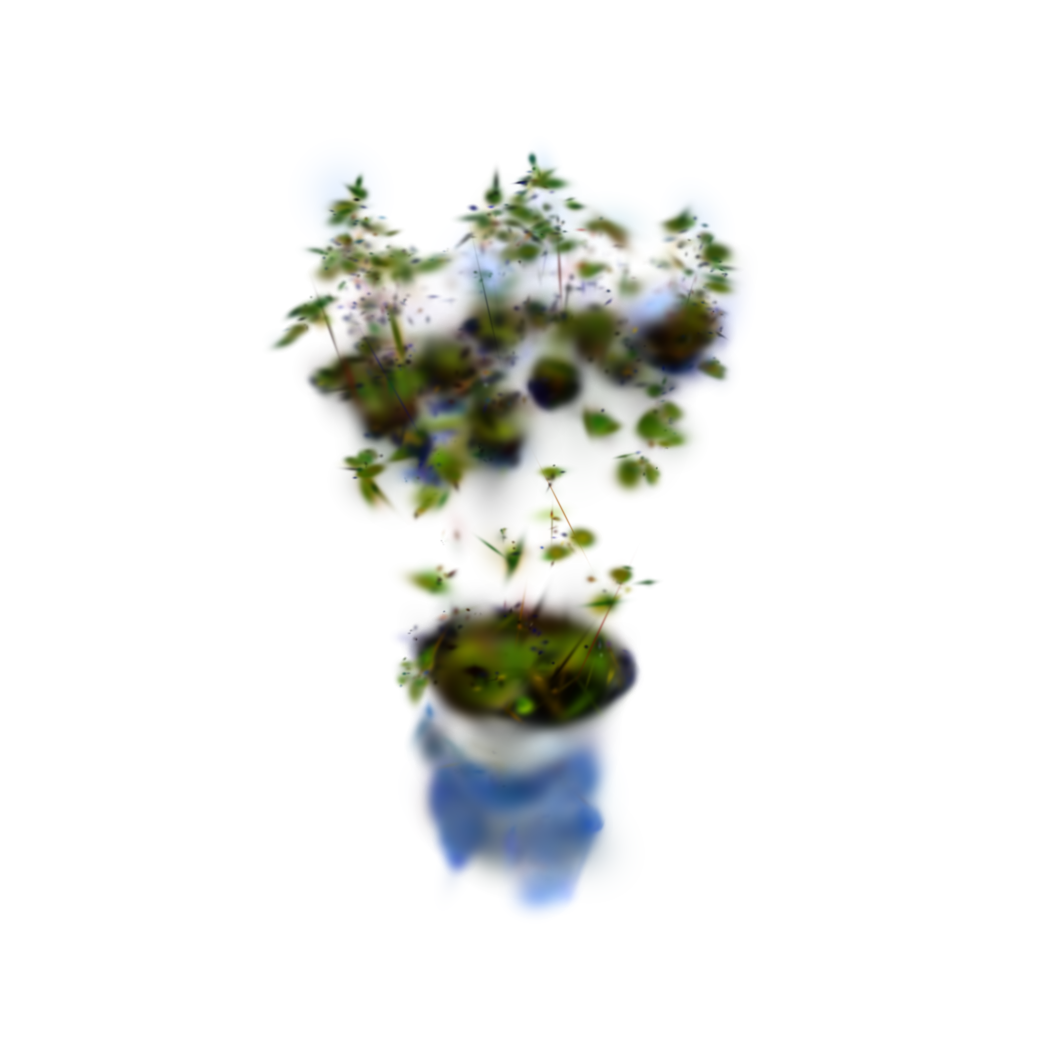} & \includegraphics[width=0.14\textwidth]{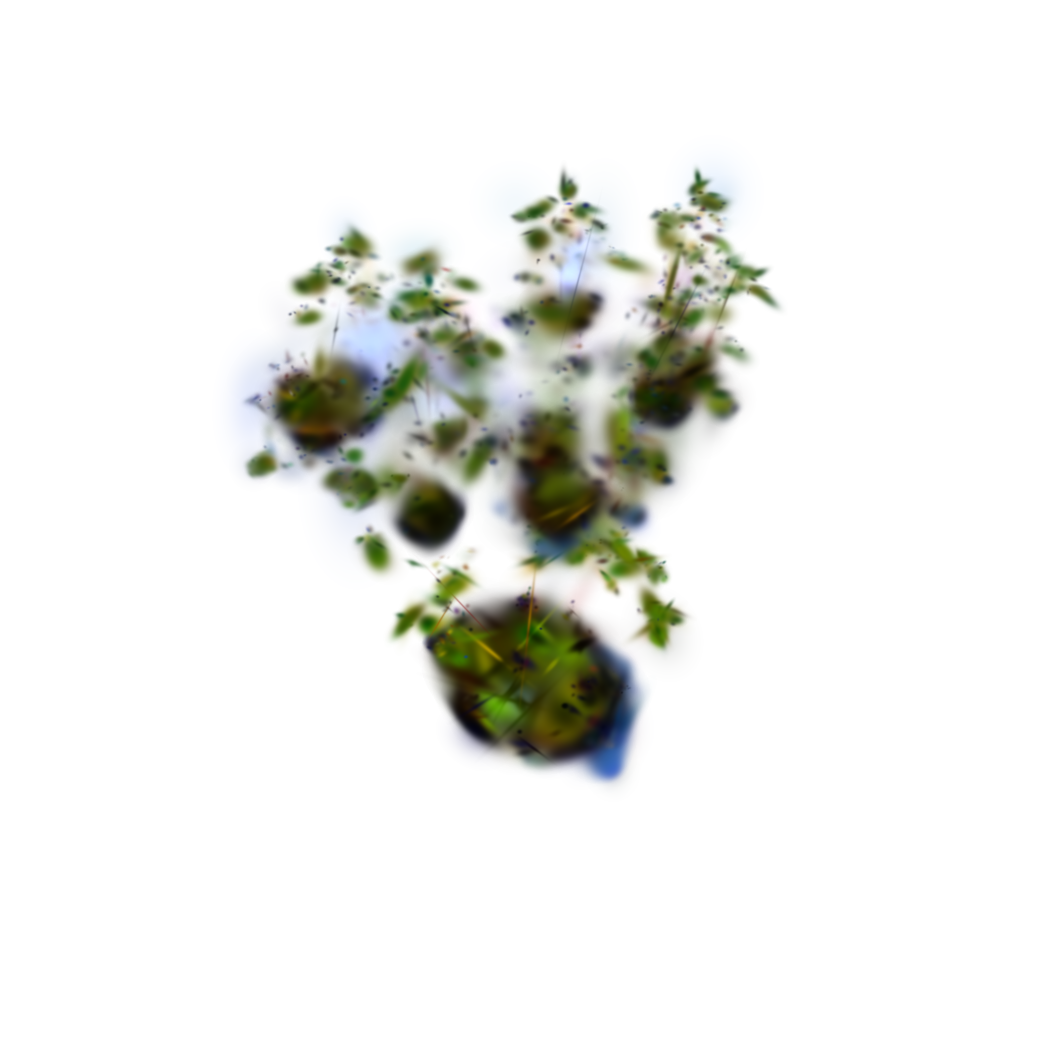} \\ 

\end{tabular}

\caption{Comparison of 3D plant models for kale, mint and bean after being initialised by 3DGS, MVS and SfM reconstructed point clouds}
\label{tab:your_table_label} 
\end{figure*}

\begin{figure*}[b] 
\centering
\setlength{\tabcolsep}{1.5pt}
\renewcommand{\arraystretch}{0.5}
\begin{tabular}{ccccccc} 

& \multicolumn{2}{c}{\textit{Bean Plant}} & \multicolumn{2}{c}{\textit{Kale Plant}}  & \multicolumn{2}{c}{\textit{Mint Plant}} \vspace{0.3cm}\\ 

\raisebox{3.5em}{\parbox{2.3cm}{\centering Black }} & 
\includegraphics[width=0.14\textwidth]{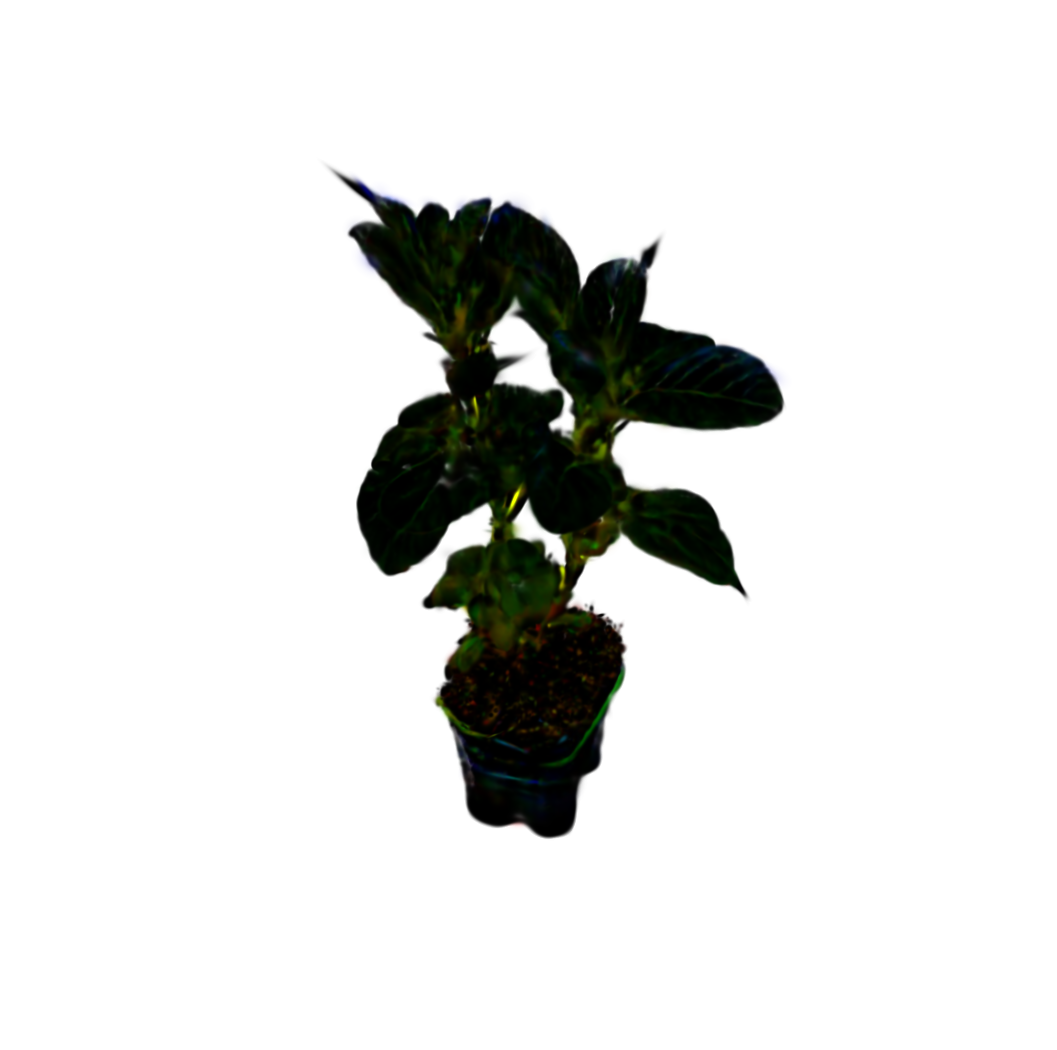} & 
\includegraphics[width=0.14\textwidth]{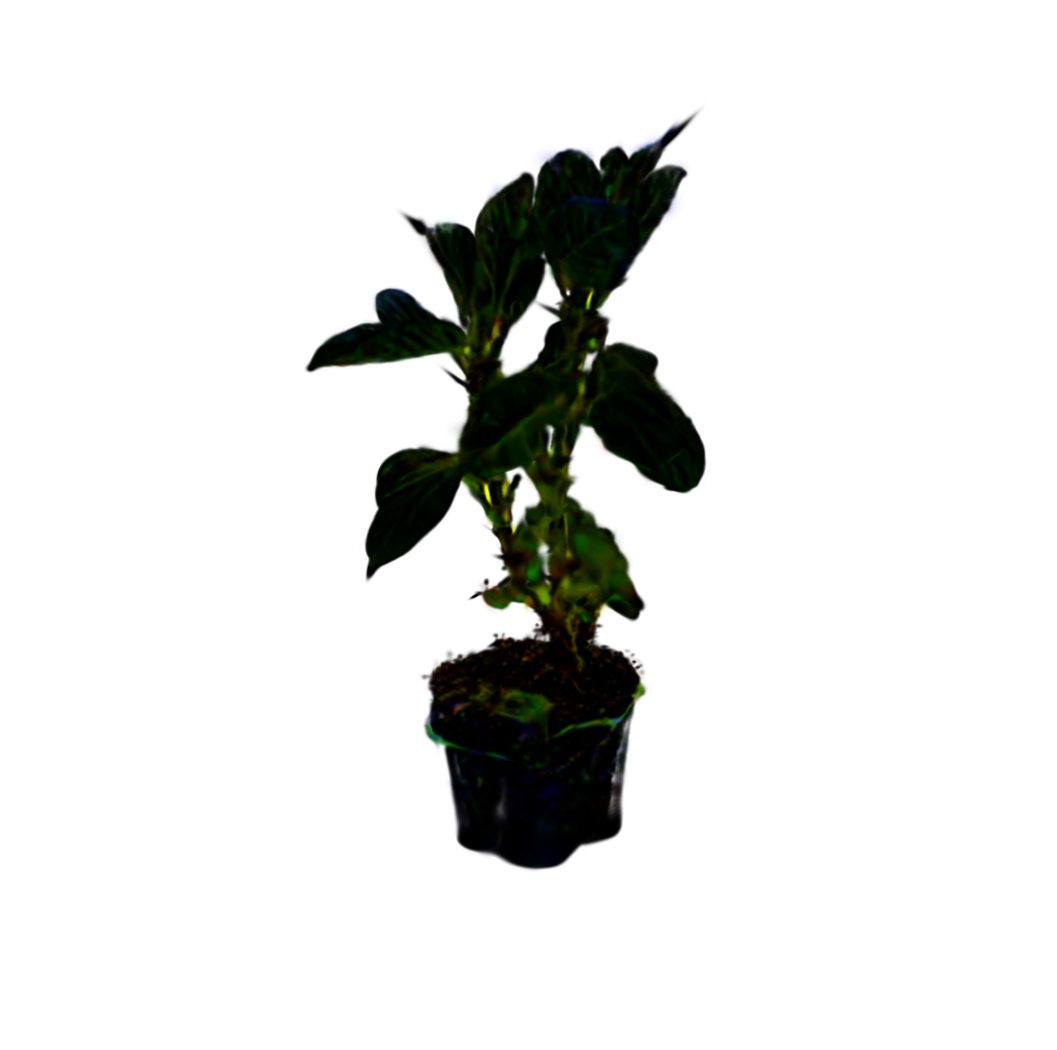} & \includegraphics[width=0.14\textwidth]{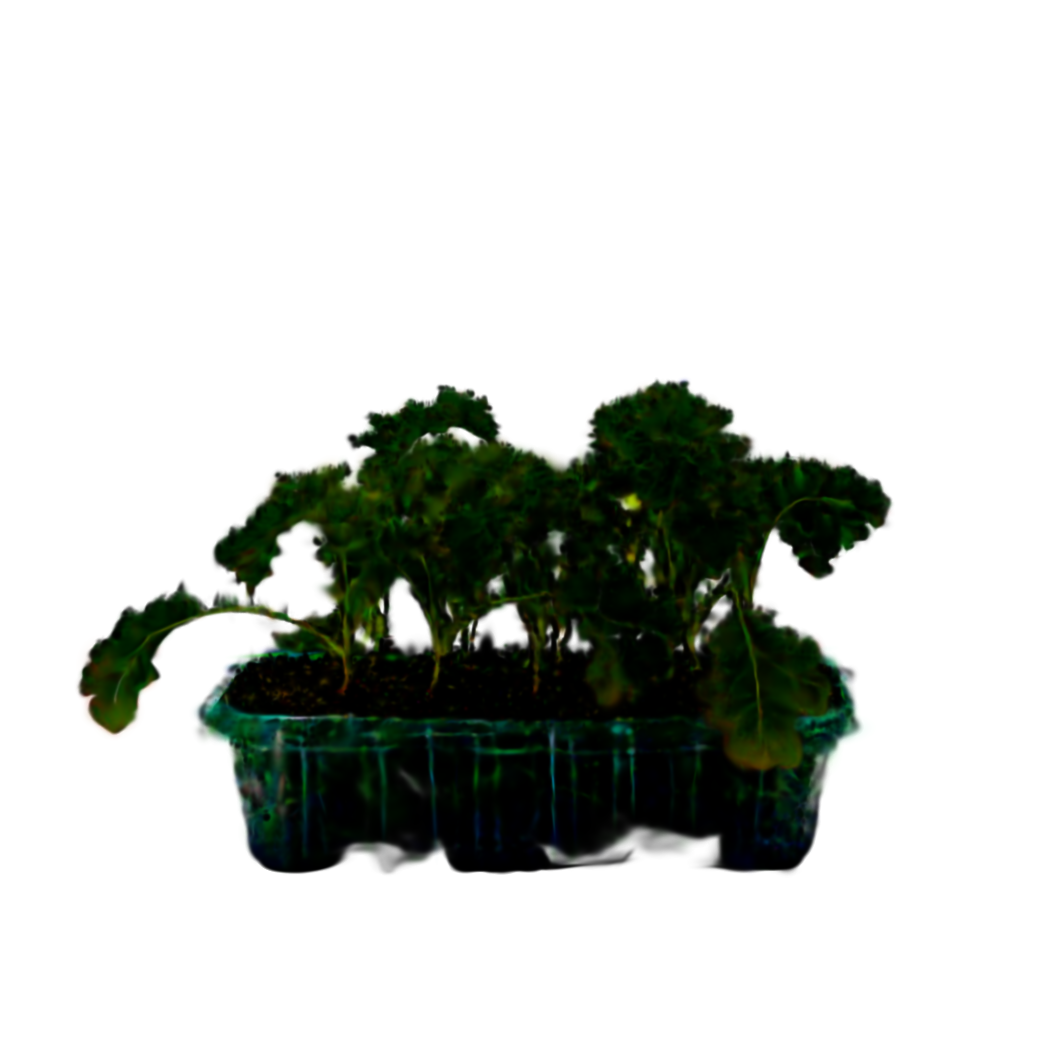} & \includegraphics[width=0.14\textwidth]{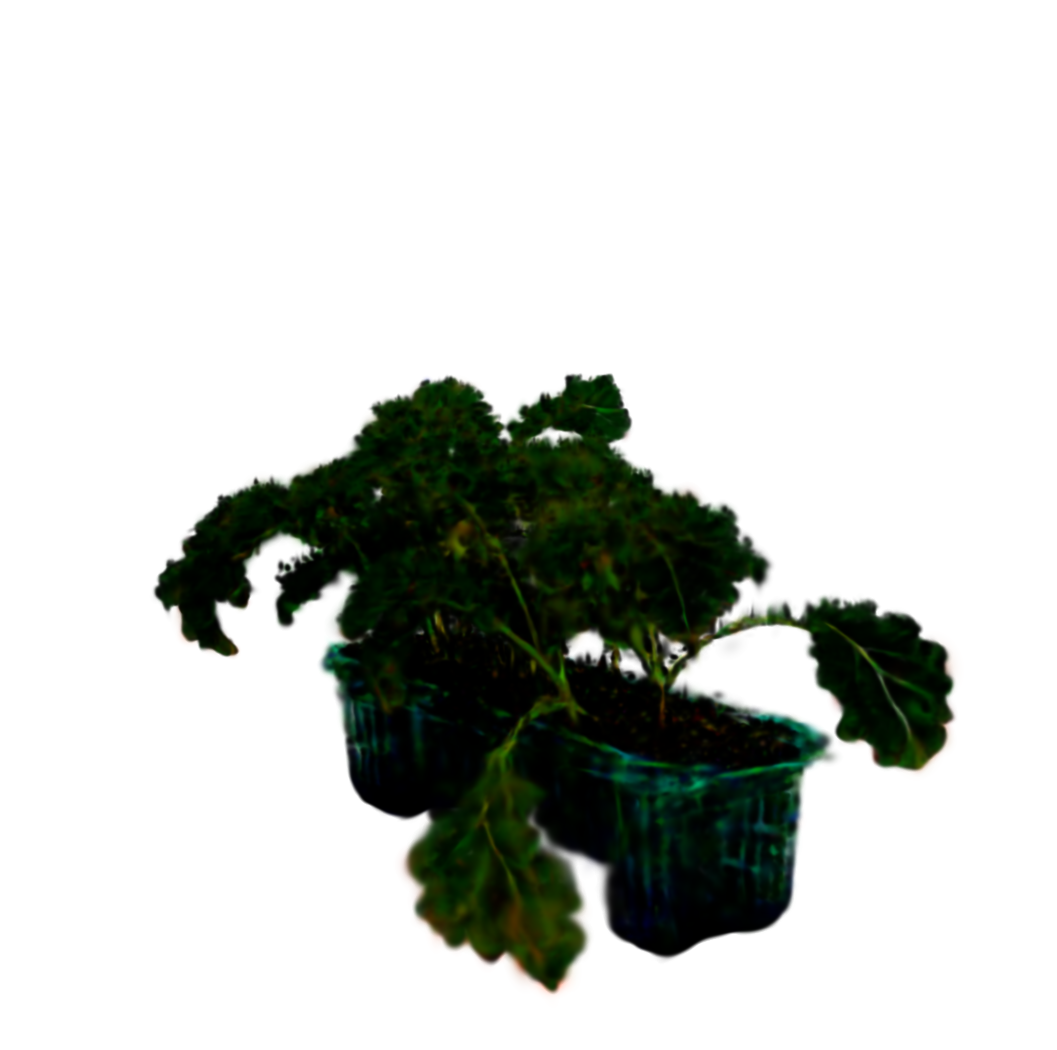} & \includegraphics[width=0.14\textwidth]{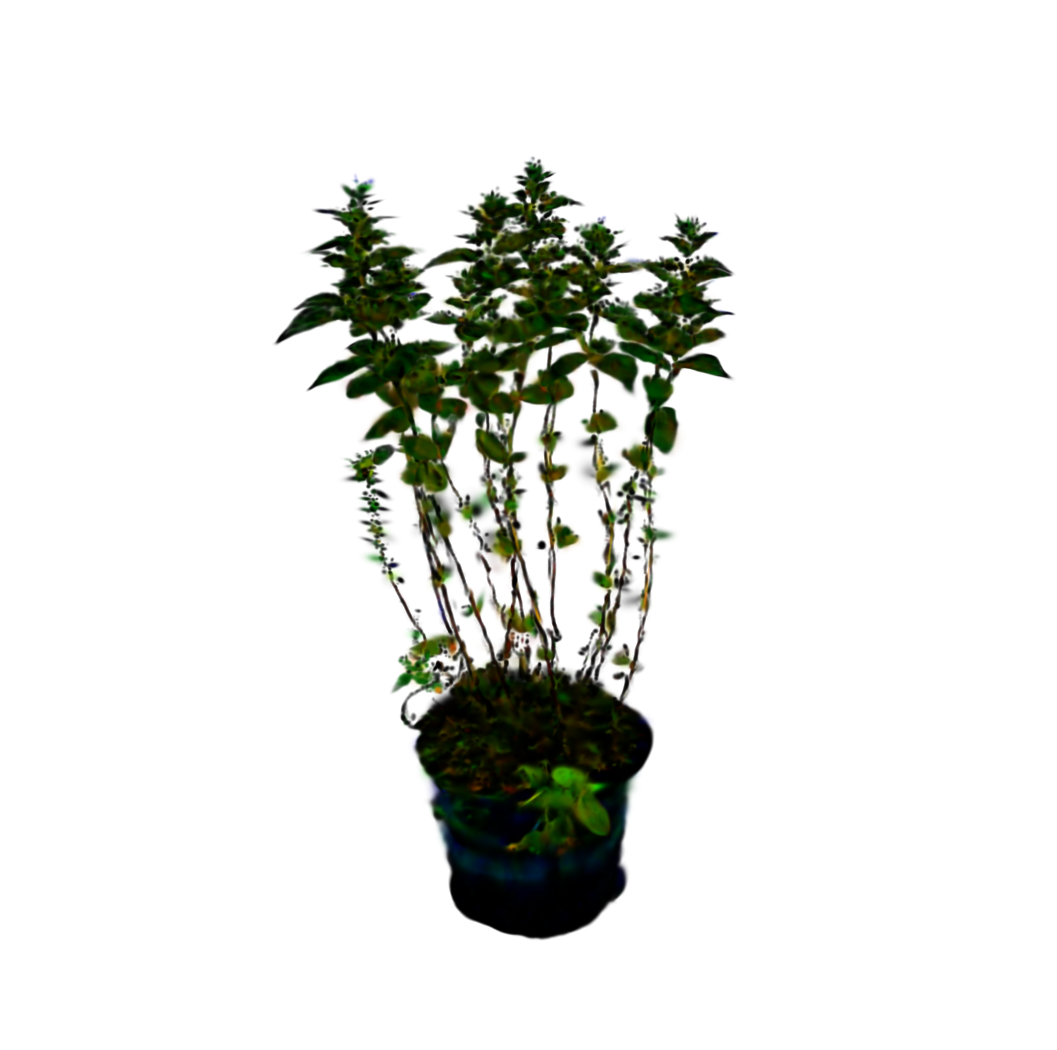} & \includegraphics[width=0.14\textwidth]{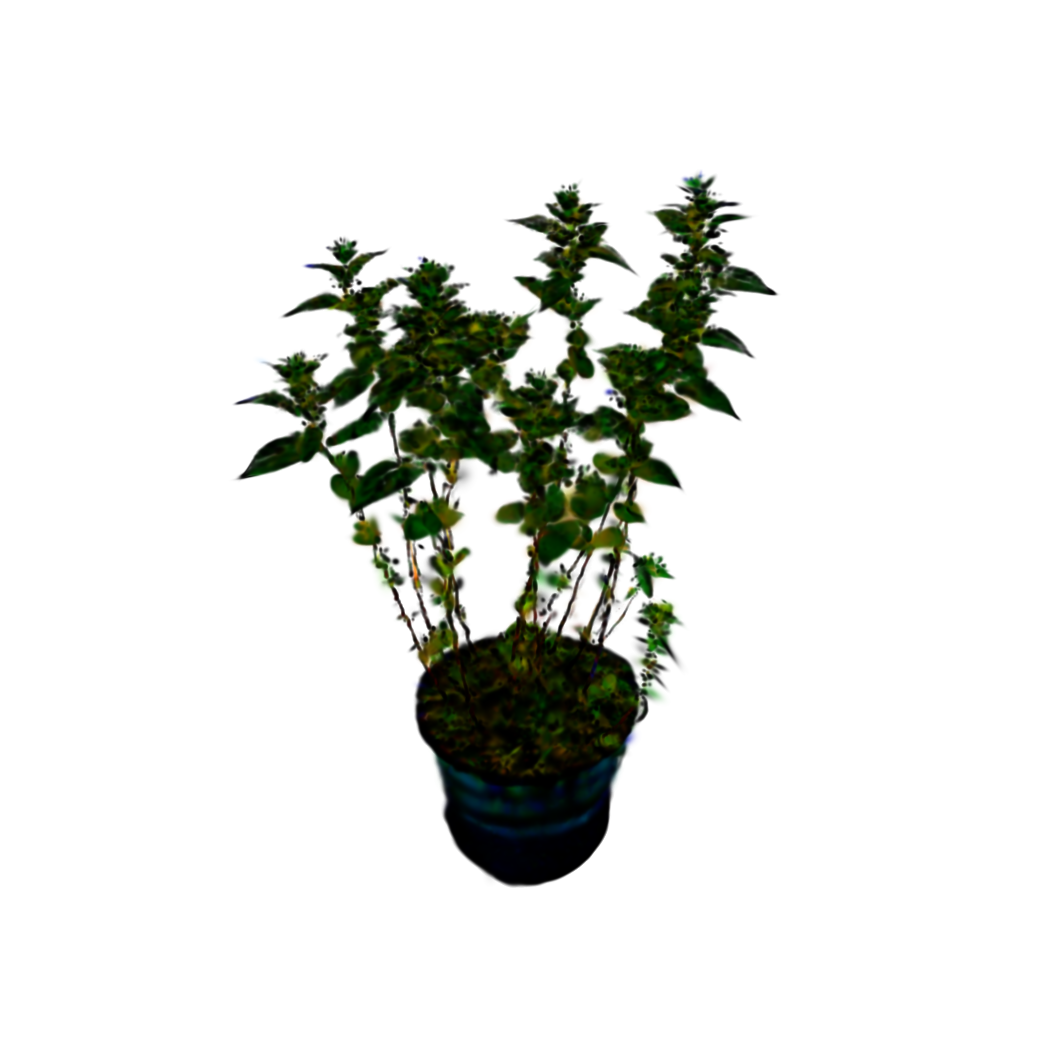} \\ 

\raisebox{3.5em}{\parbox{2.3cm}{\centering White }} & 
\includegraphics[width=0.14\textwidth]{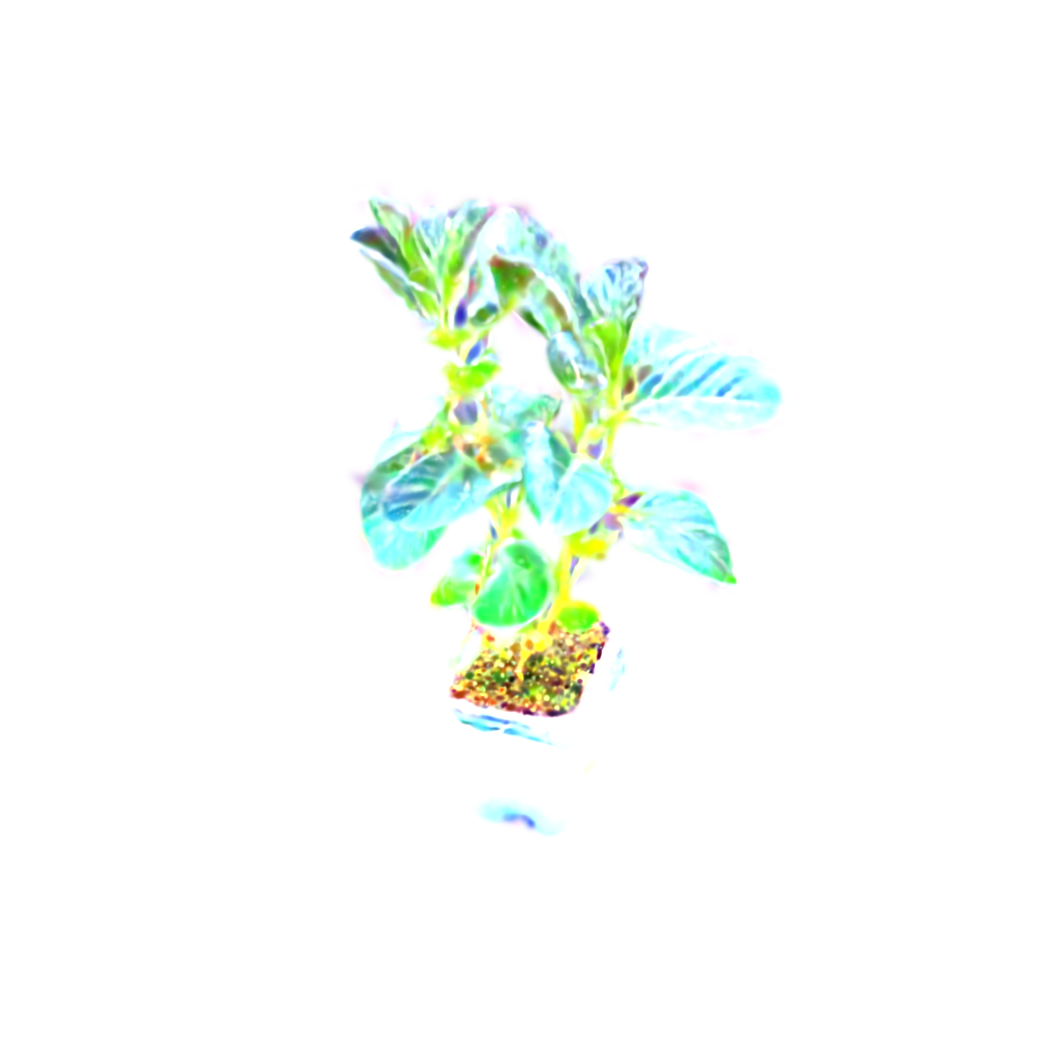} & 
\includegraphics[width=0.14\textwidth]{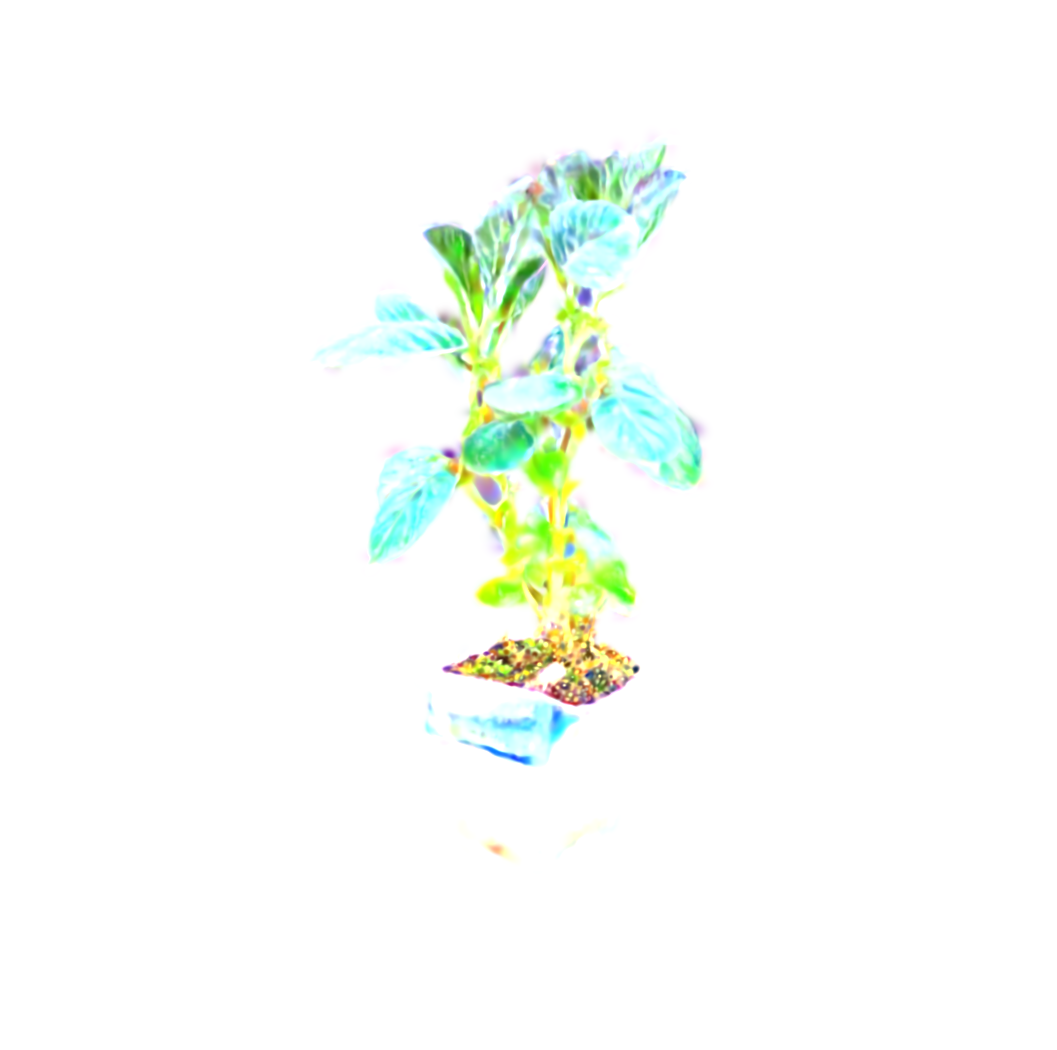} & \includegraphics[width=0.14\textwidth]{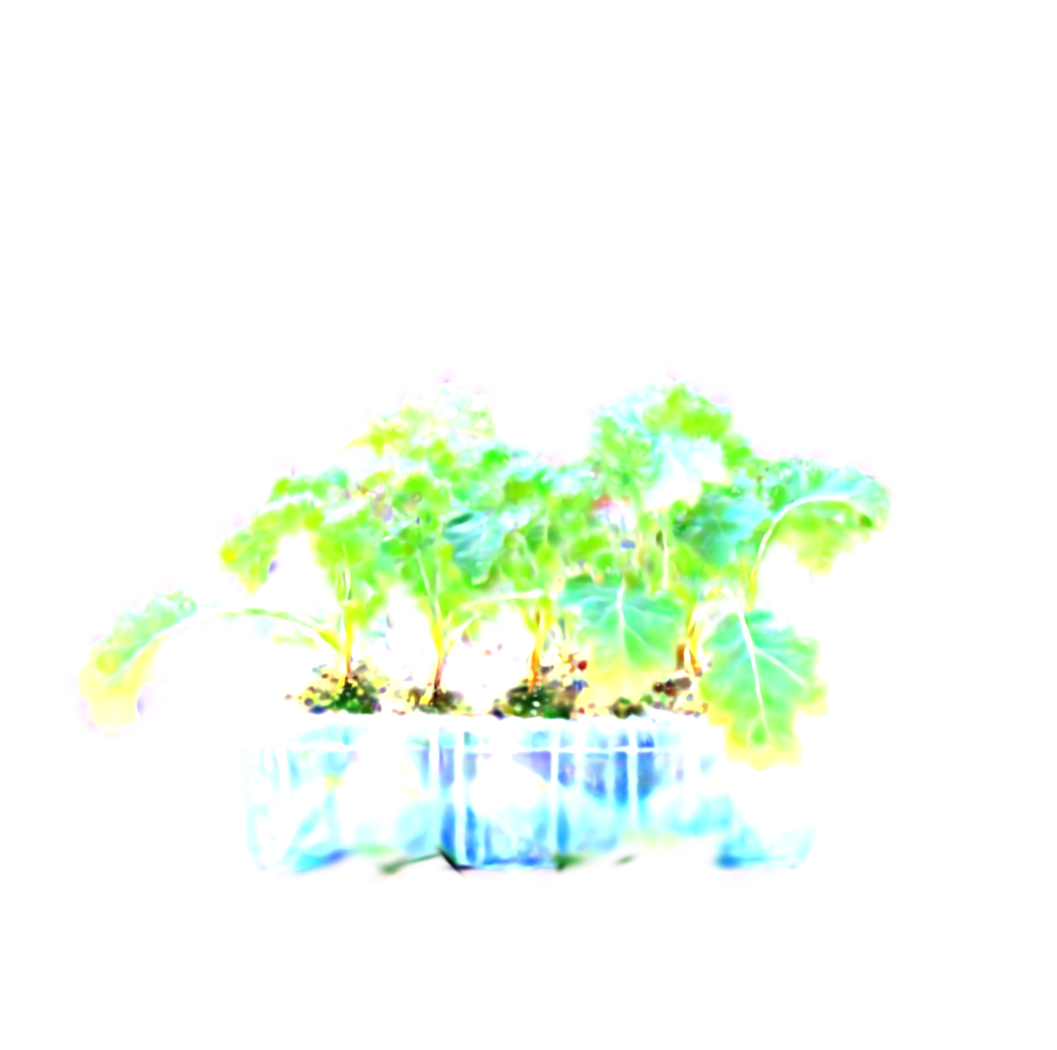} & \includegraphics[width=0.14\textwidth]{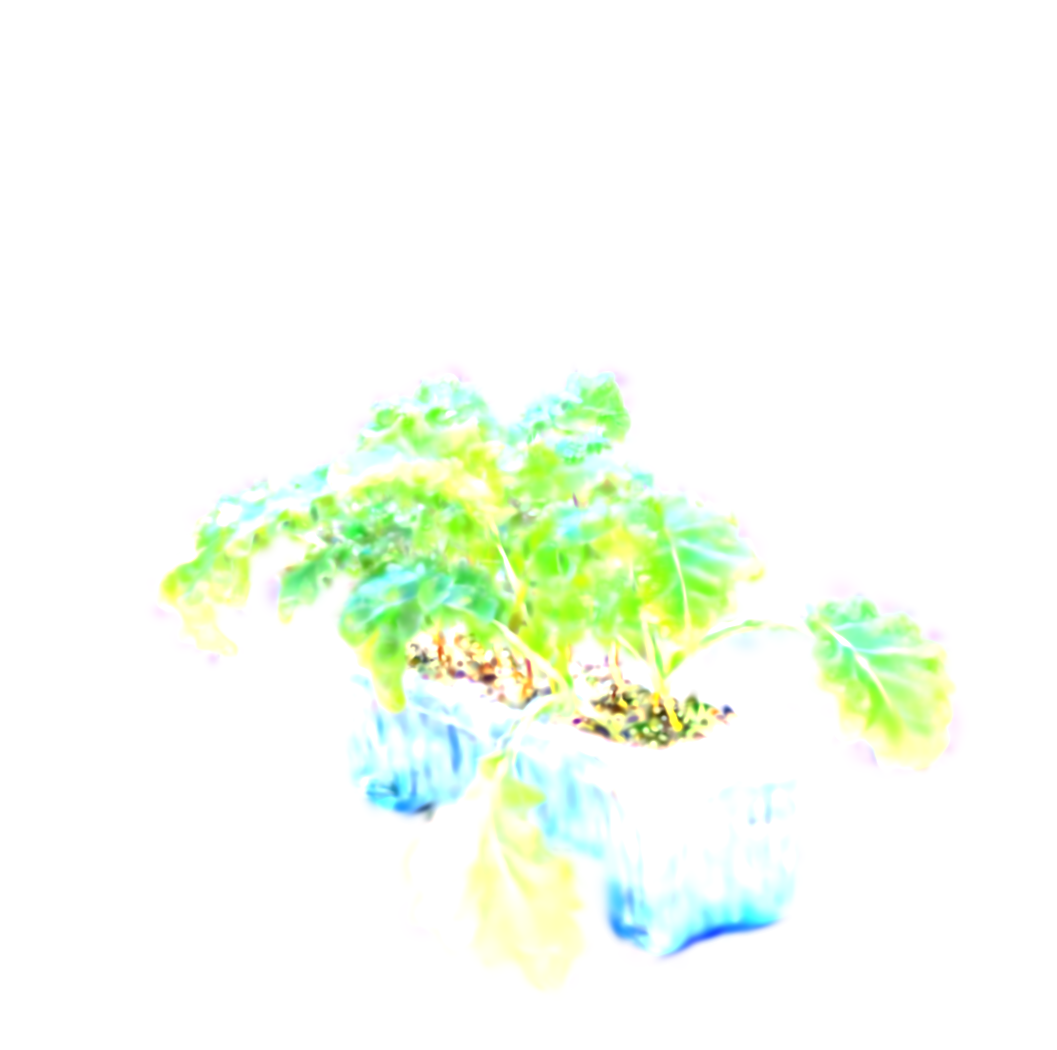} & \includegraphics[width=0.14\textwidth]{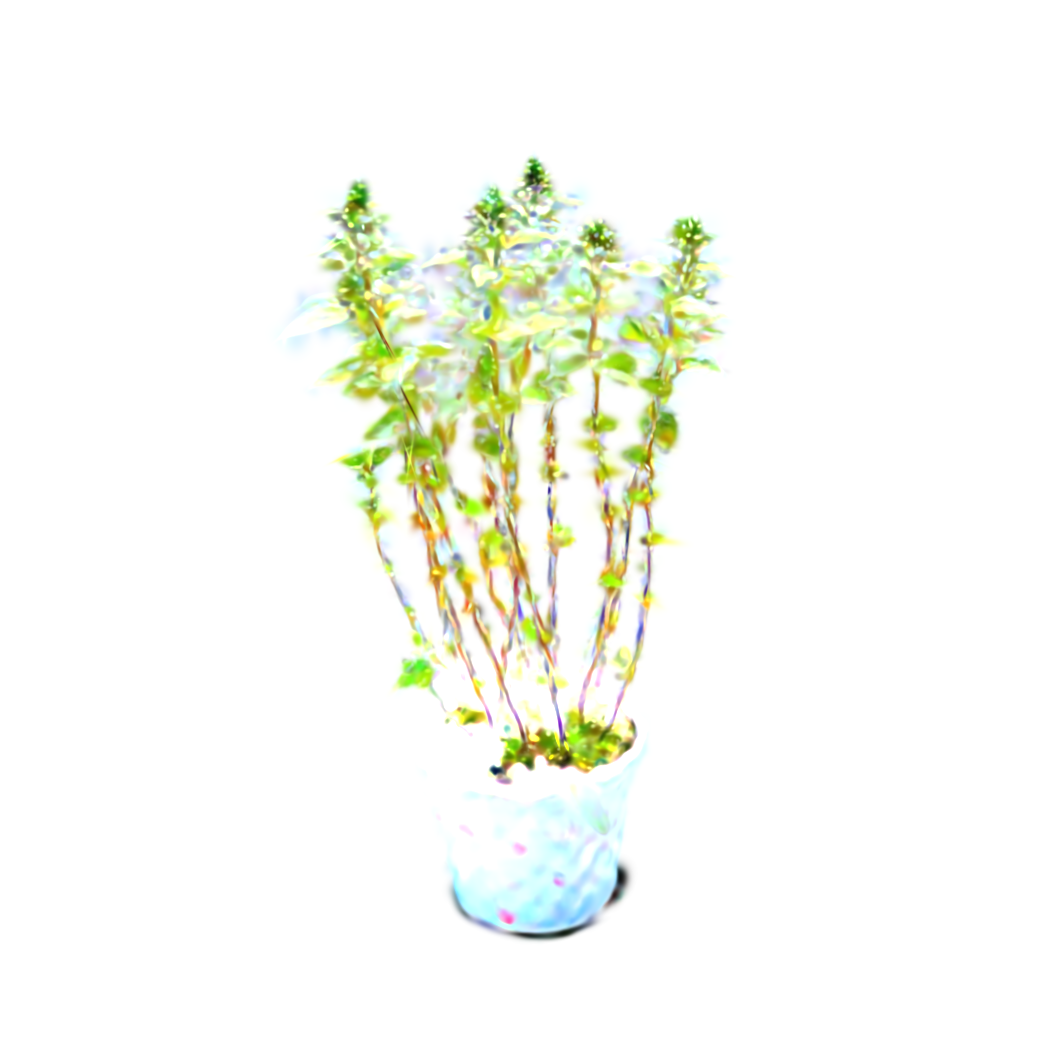} & \includegraphics[width=0.14\textwidth]{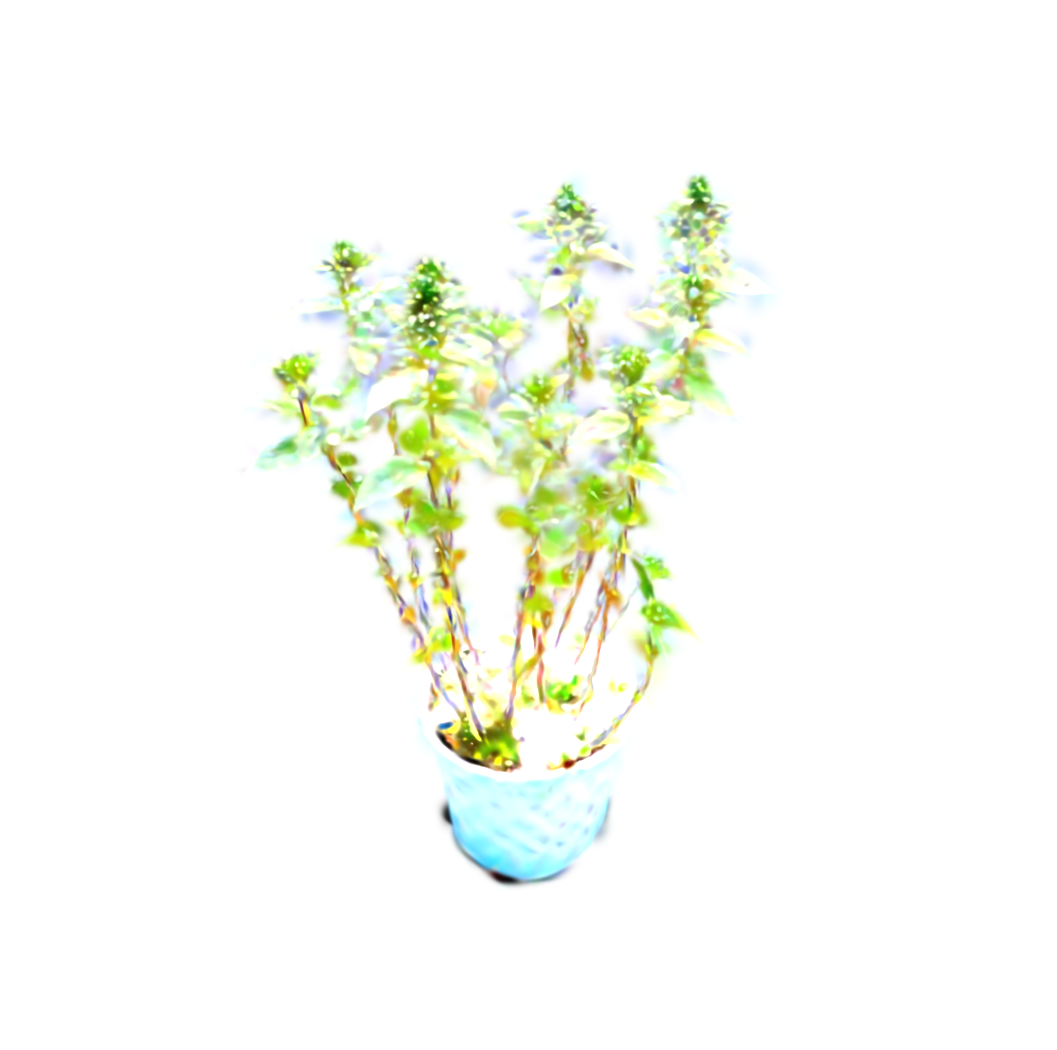} \\ 

\raisebox{3.5em}{\parbox{2.5cm}{\centering Noise }} & 
\includegraphics[width=0.14\textwidth]{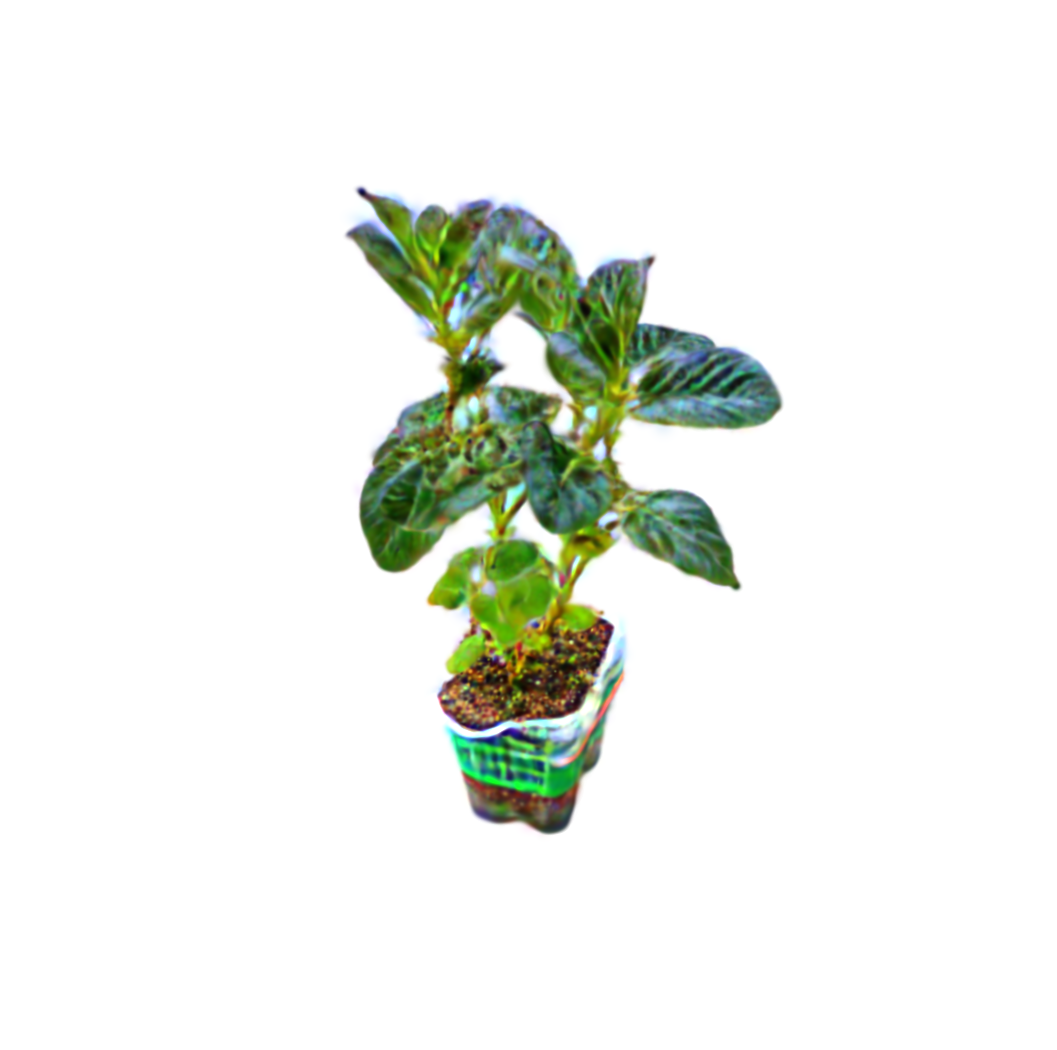} & 
\includegraphics[width=0.14\textwidth]{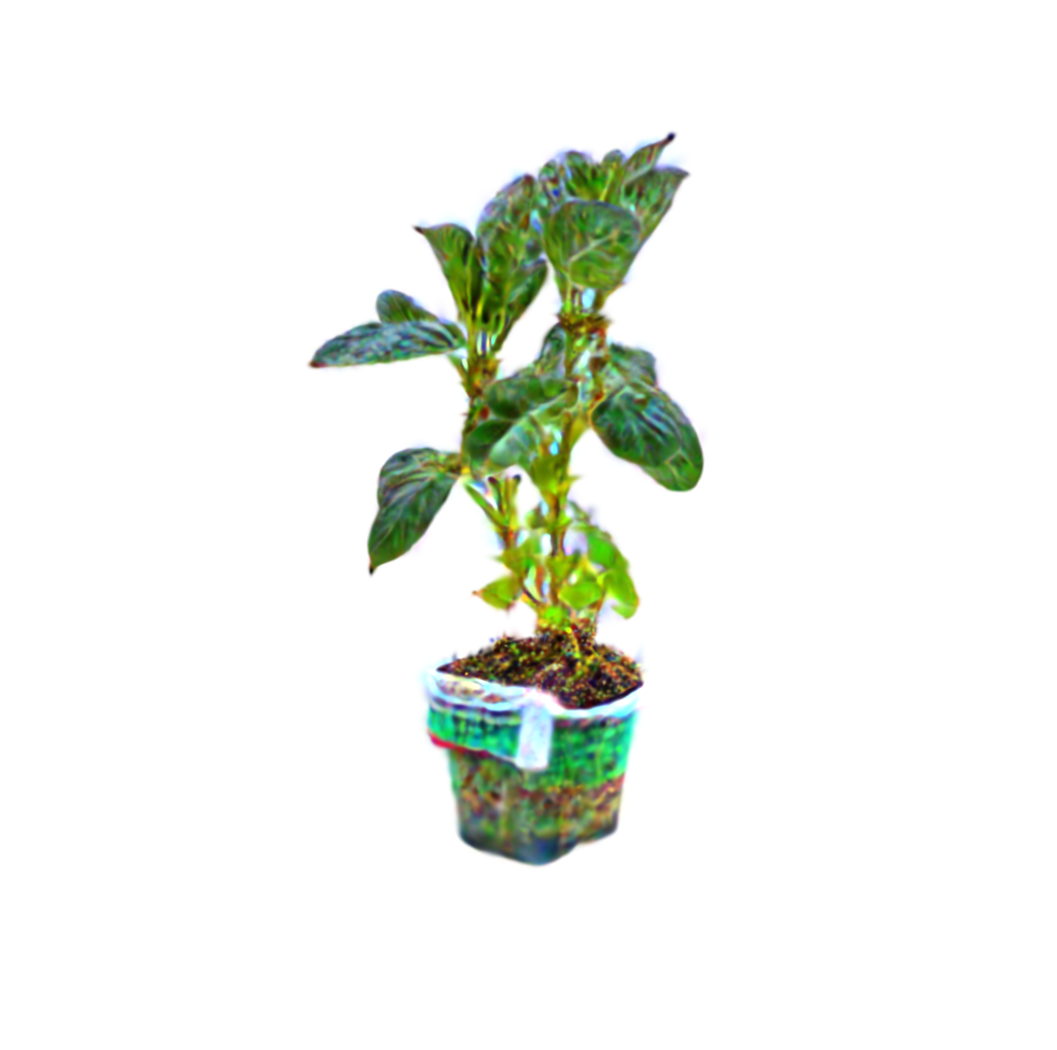} & \includegraphics[width=0.14\textwidth]{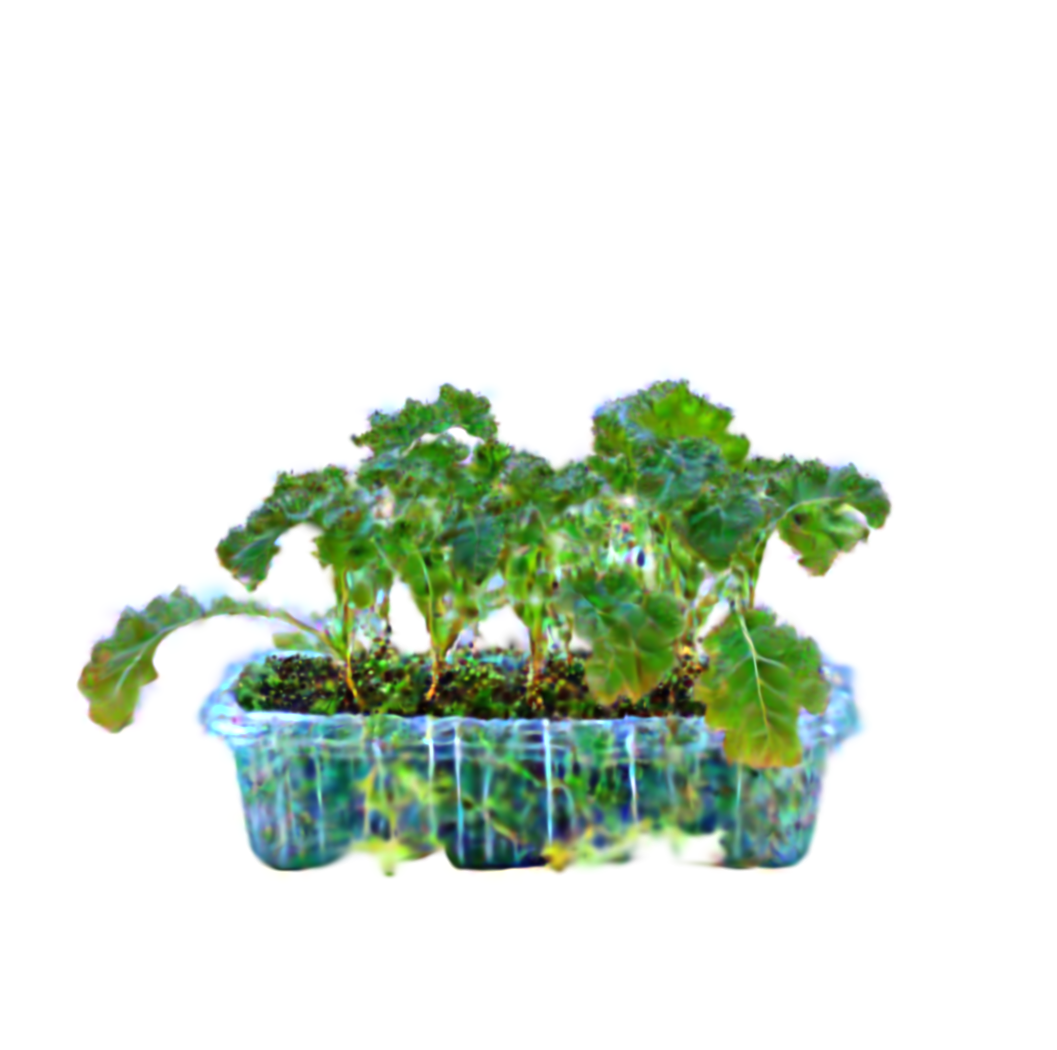} & \includegraphics[width=0.14\textwidth]{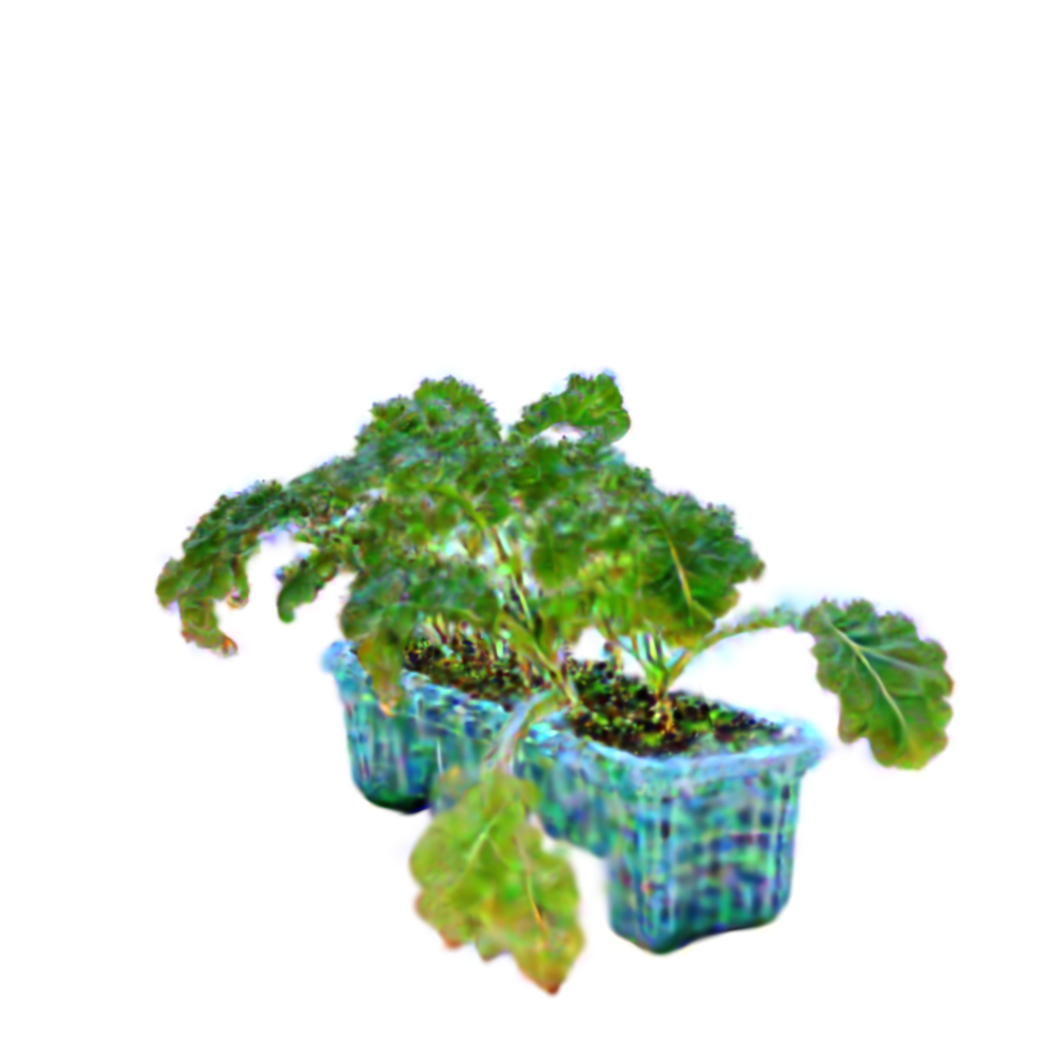} & \includegraphics[width=0.14\textwidth]{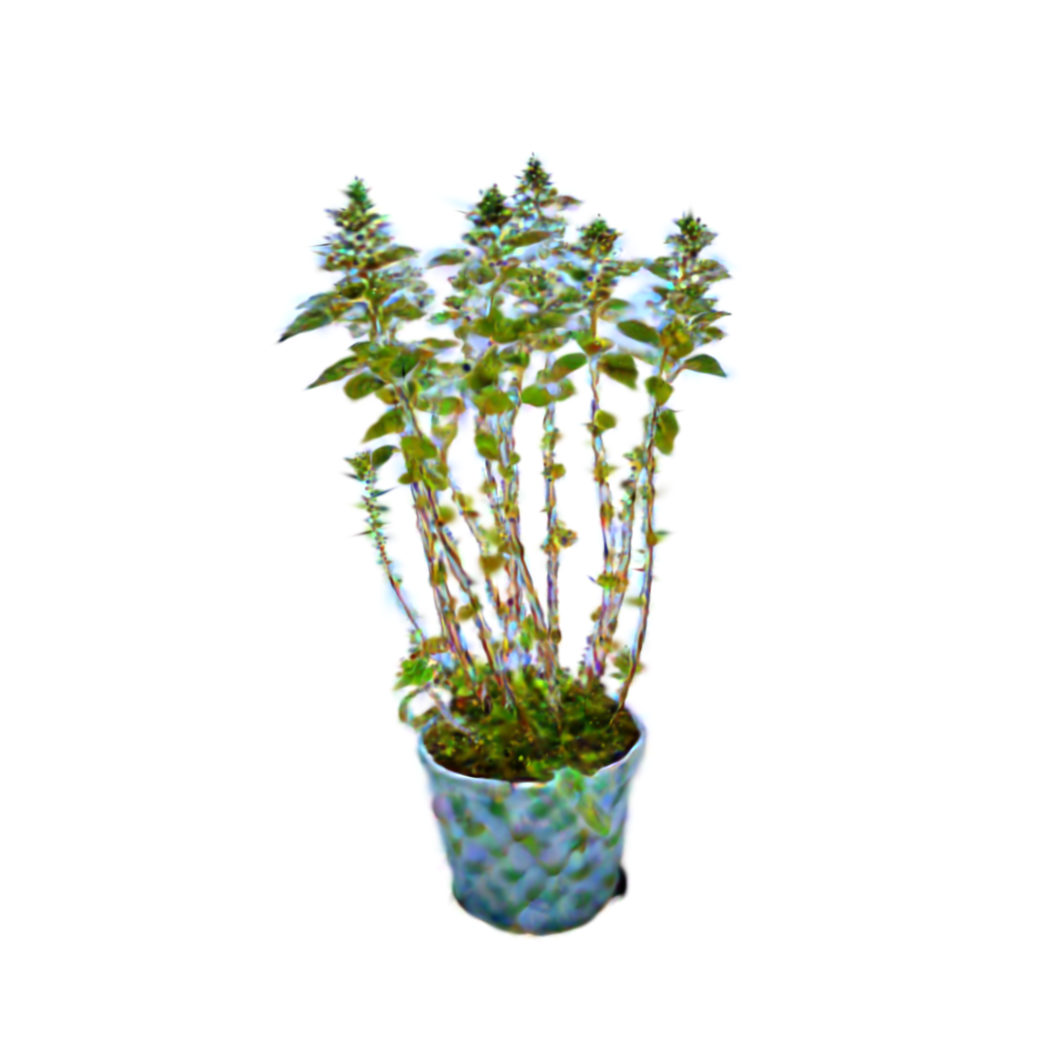} & \includegraphics[width=0.14\textwidth]{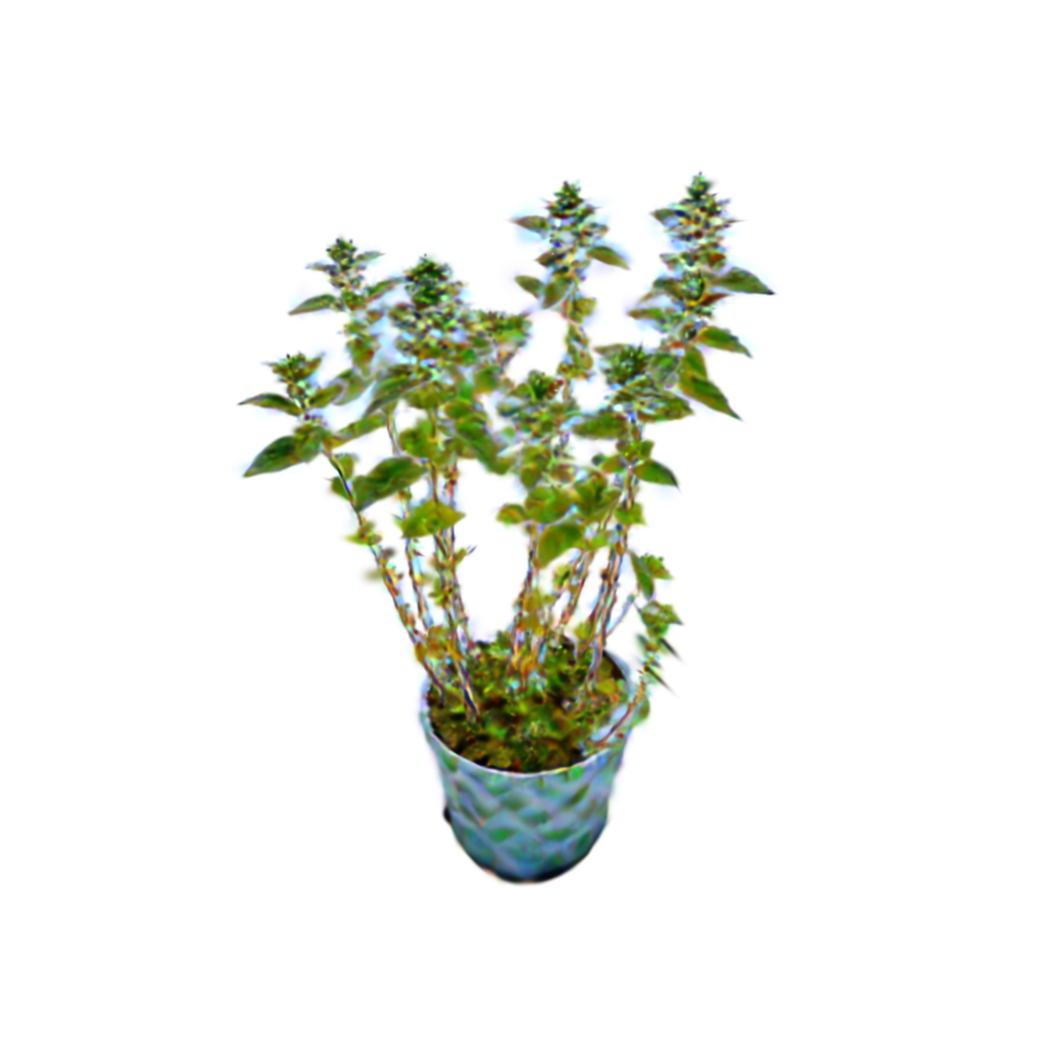} \\ 

\end{tabular}
\raggedleft
\caption{Comparison of 3D plant models for kale, mint and bean after being initialised by black, white and noised point clouds}
\label{tab:your_table_label} 
\end{figure*}

\clearpage

\section{More Real Initialised Model Renders}

\begin{figure}[h]
    \centering
\setlength{\tabcolsep}{1.0pt}
\renewcommand{\arraystretch}{0.5}

\begin{tabular}{ccccccc} 

& \multicolumn{6}{c}{\textbf{Real}} \vspace{0.2cm}\\

& \multicolumn{2}{c}{\textit{Bean Plant}} & \multicolumn{2}{c}{\textit{Kale Plant}}  & \multicolumn{2}{c}{\textit{Mint Plant}} \vspace{0.2cm}\\ 

\raisebox{3.5em}{\parbox{2.3cm}{\centering Ground Truth }} & \includegraphics[width=0.14\textwidth]{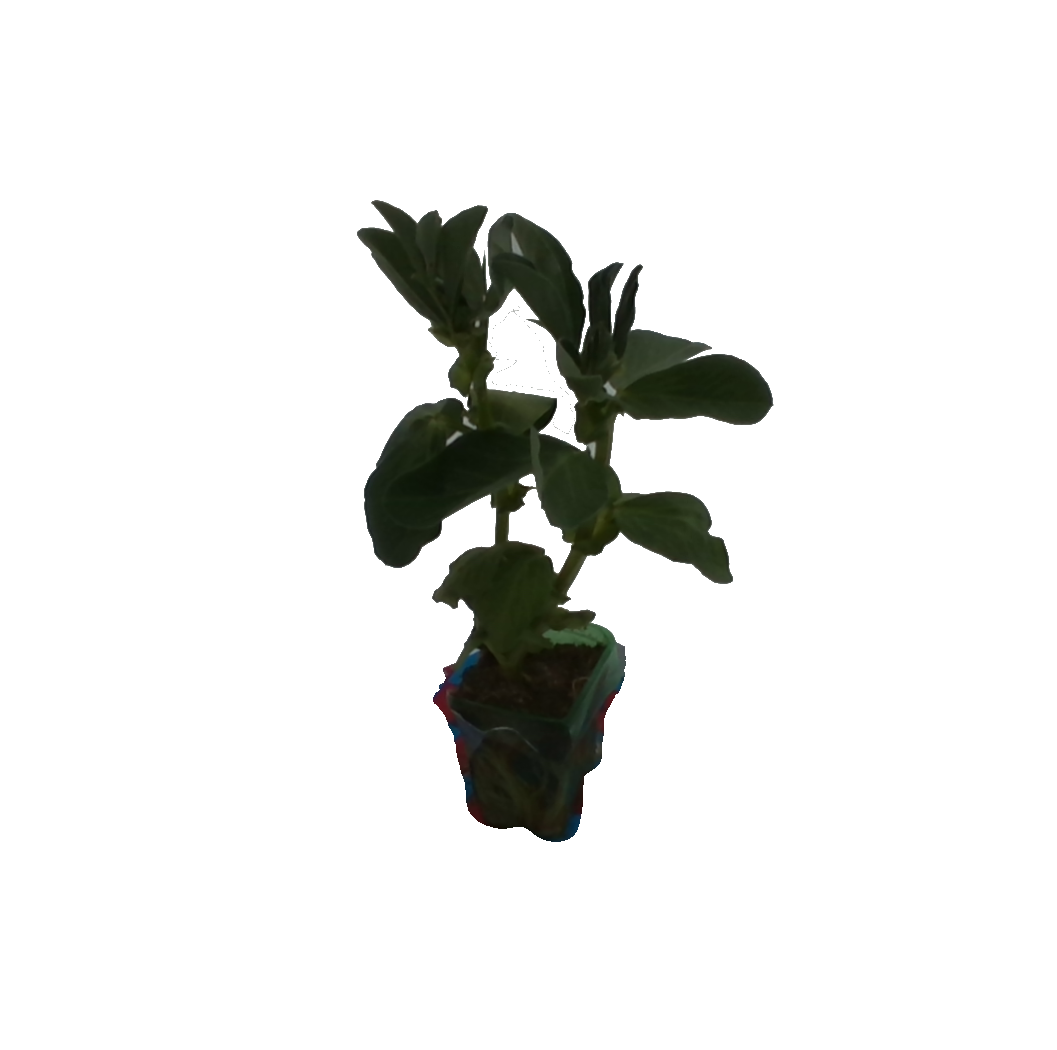} & \includegraphics[width=0.14\textwidth]{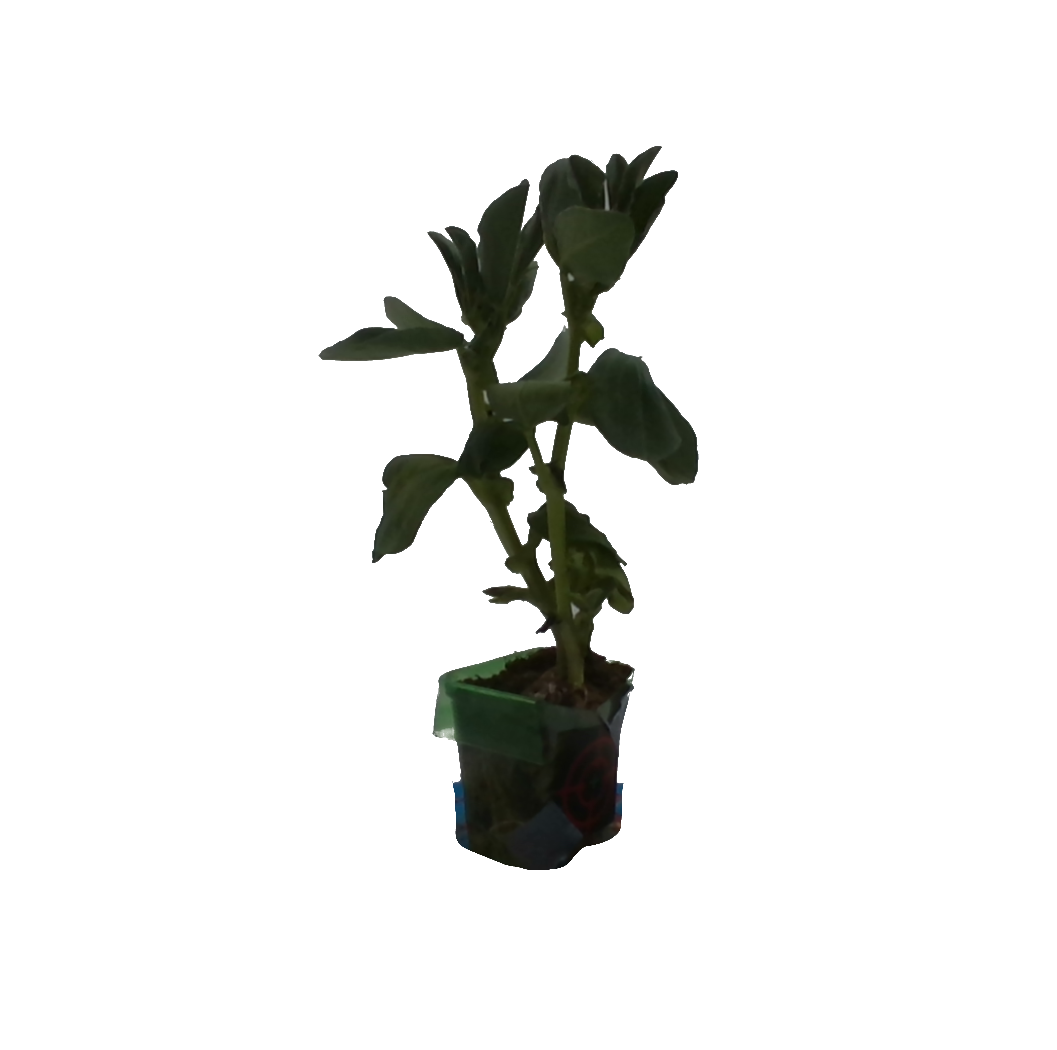} & \includegraphics[width=0.14\textwidth]{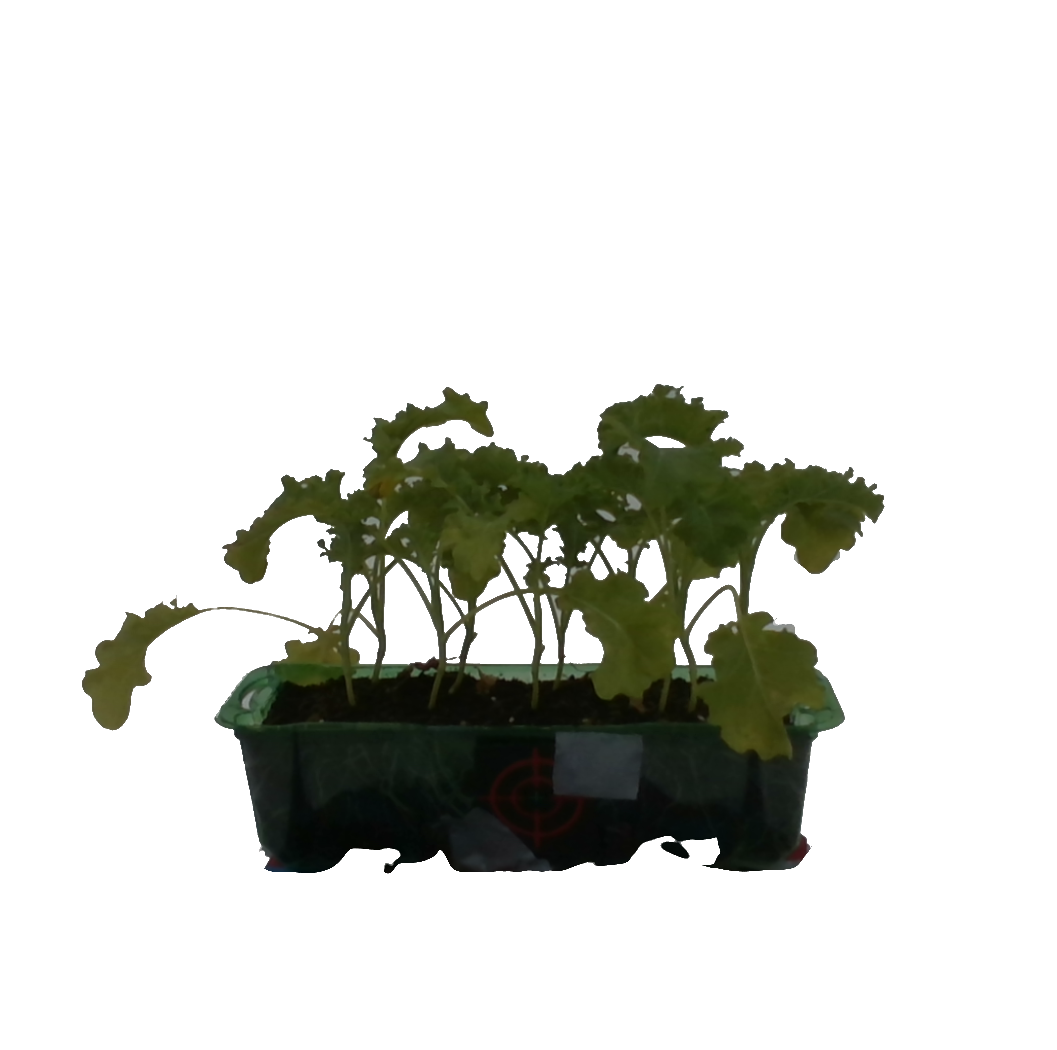} & \includegraphics[width=0.14\textwidth]{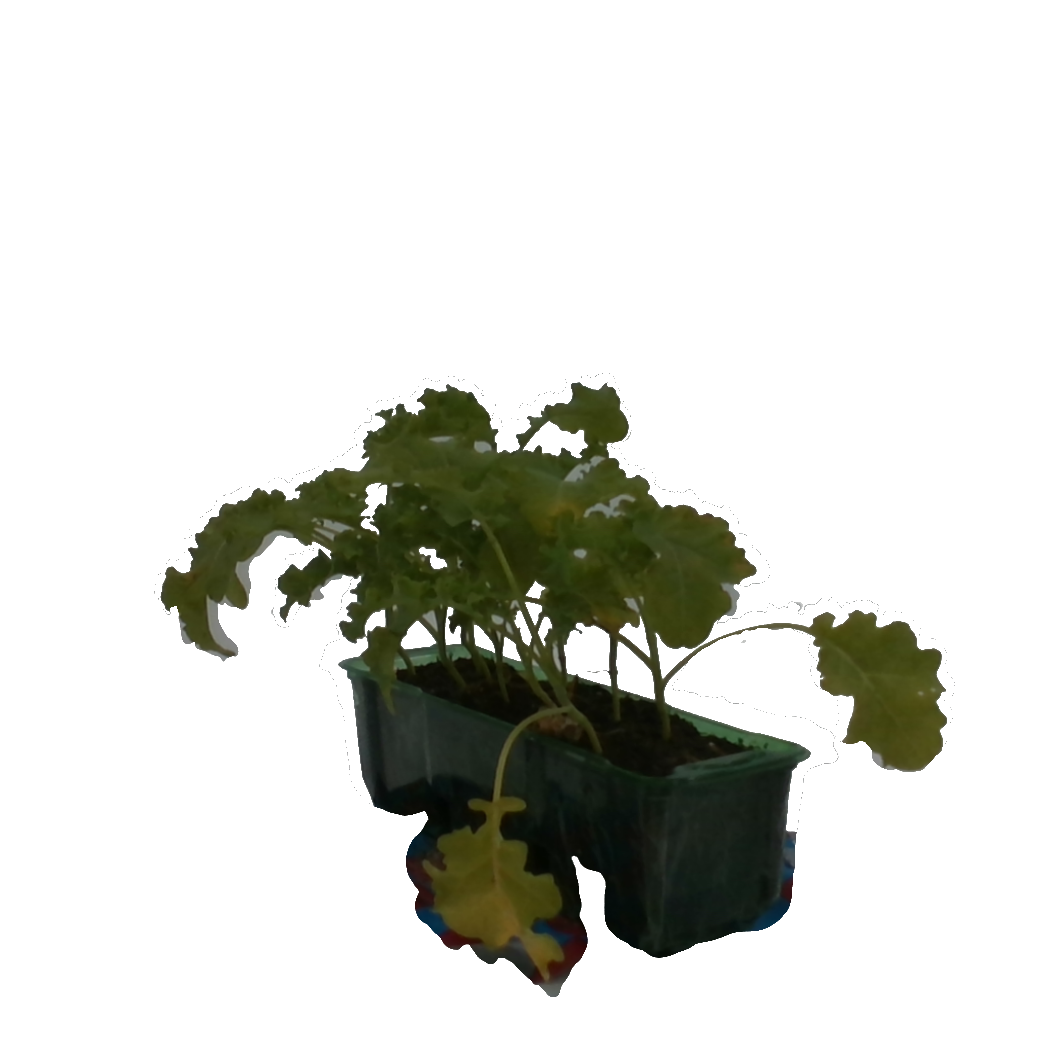} &
\includegraphics[width=0.14\textwidth]{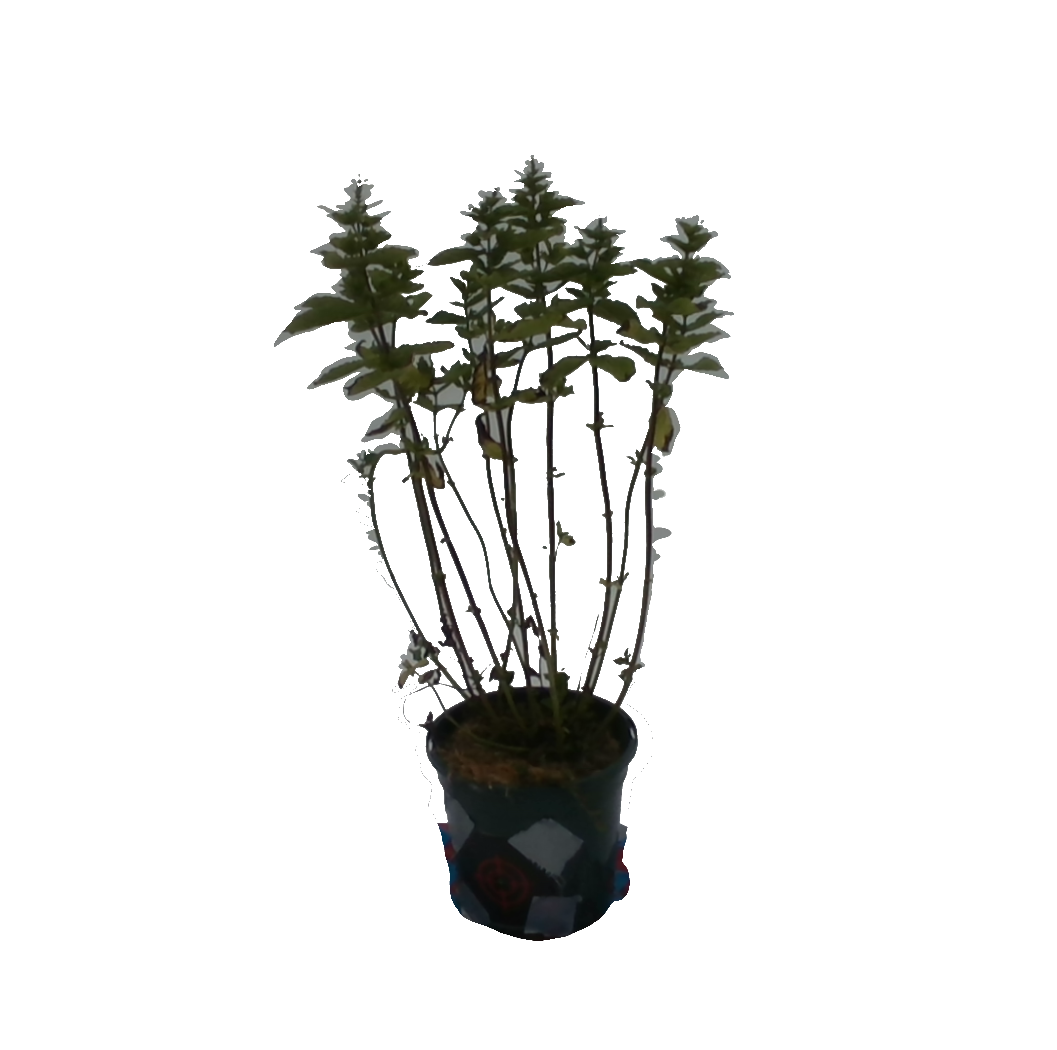} & 
\includegraphics[width=0.14\textwidth]{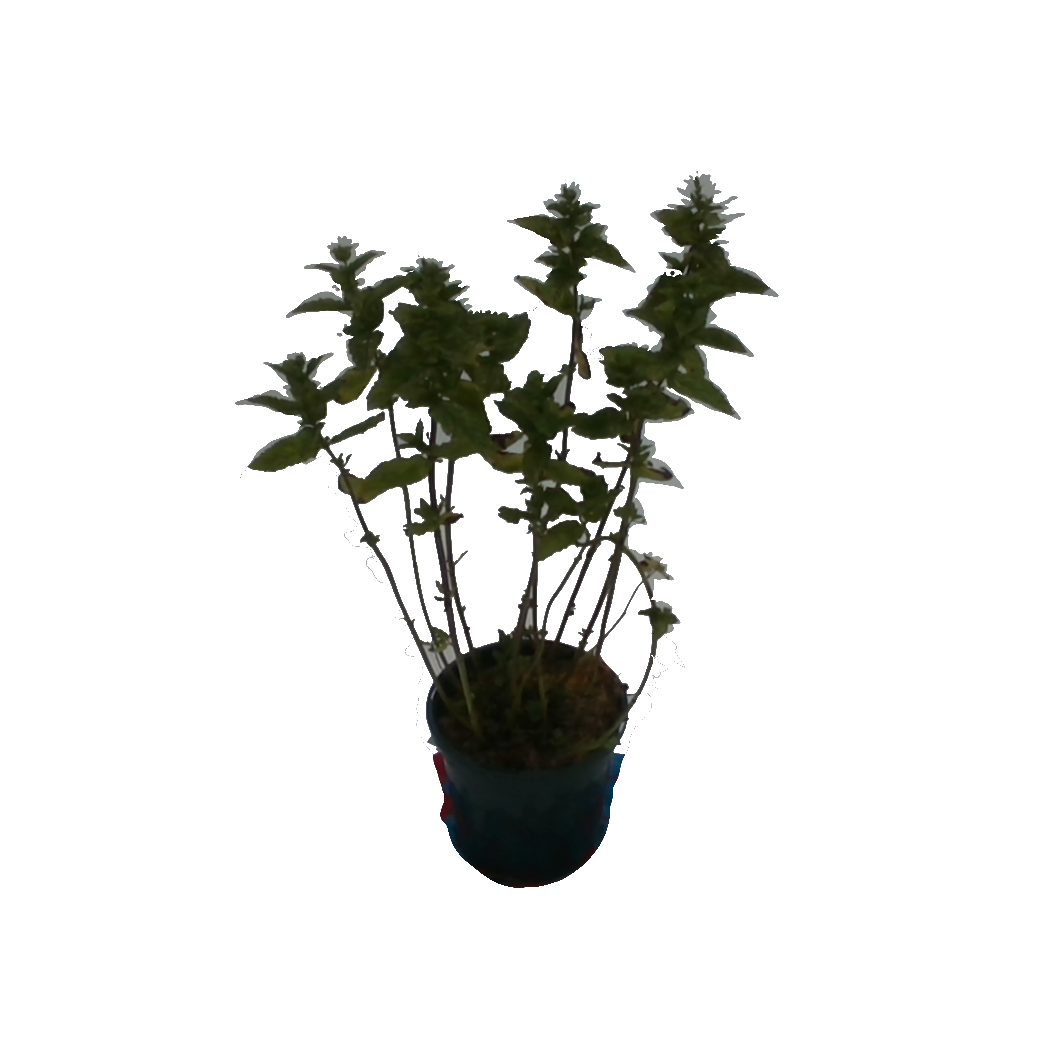}\\ 

\raisebox{3.5em}{\parbox{2.5cm}{\centering GaussianDreamer }} & \includegraphics[width=0.14\textwidth]{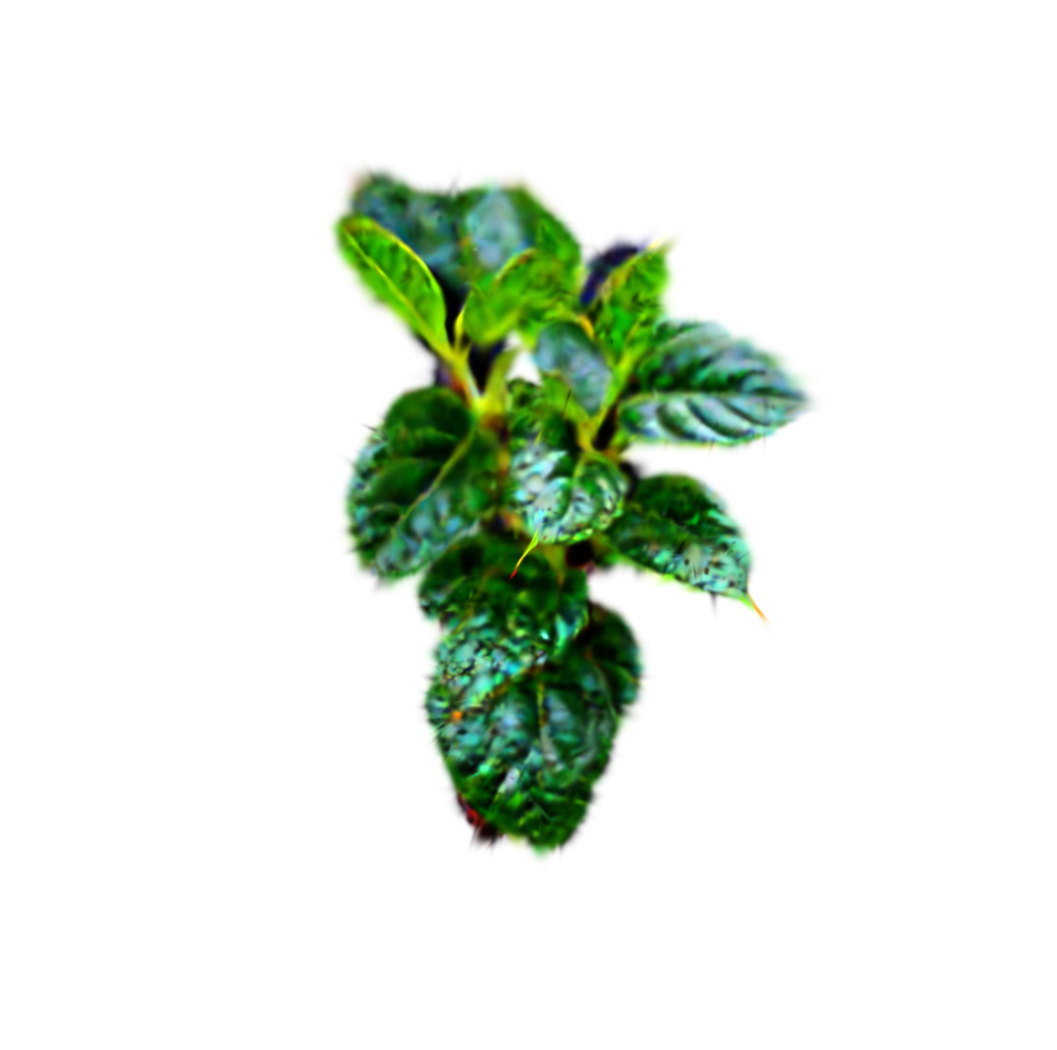} & \includegraphics[width=0.14\textwidth]{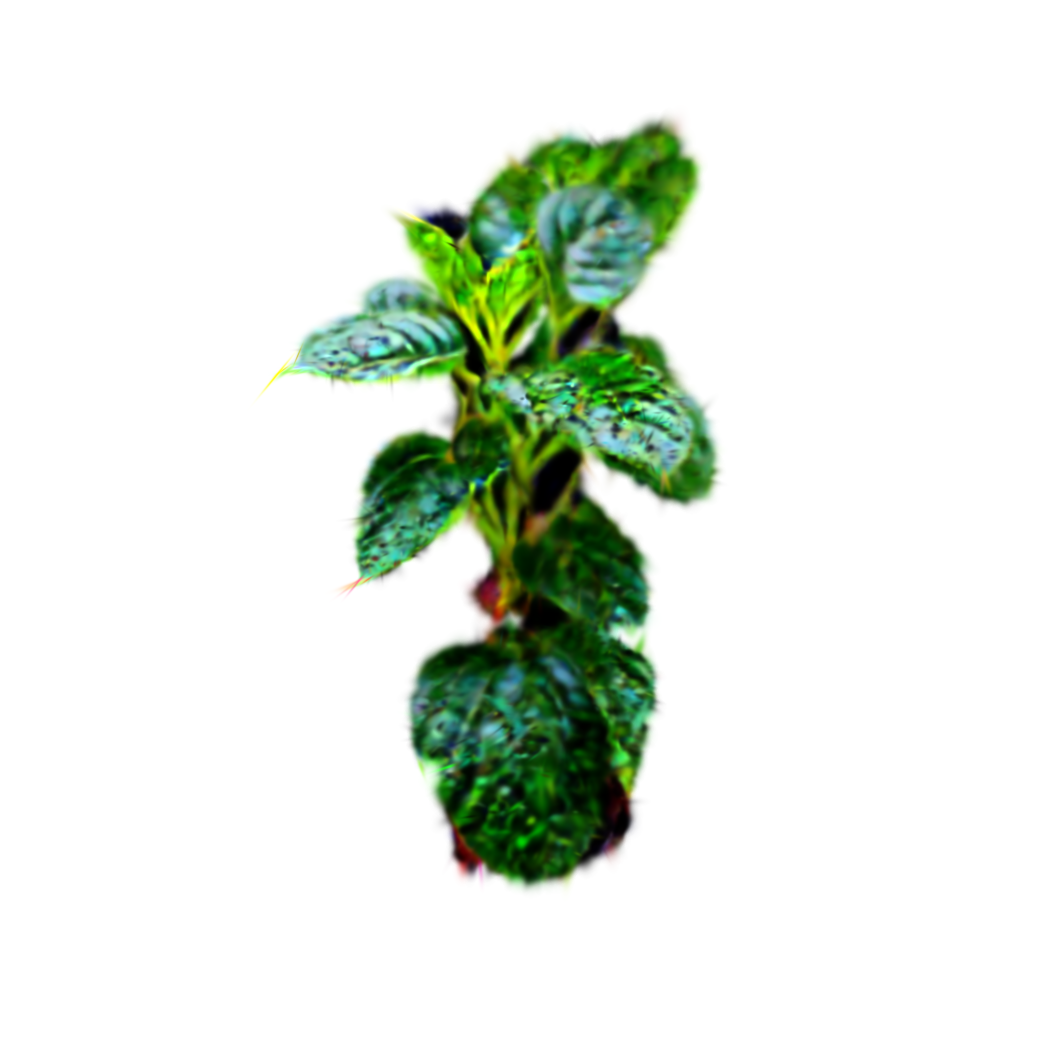} & \includegraphics[width=0.14\textwidth]{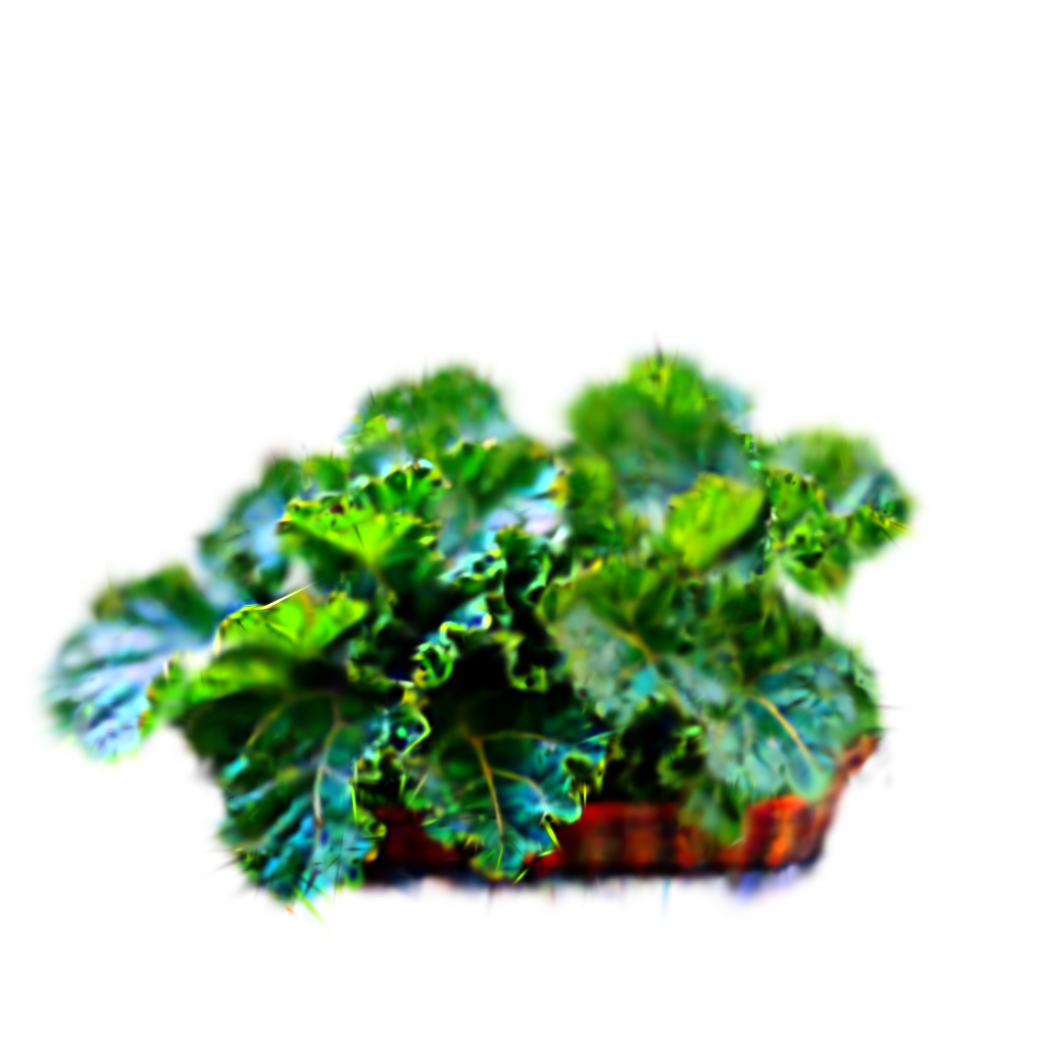} & \includegraphics[width=0.14\textwidth]{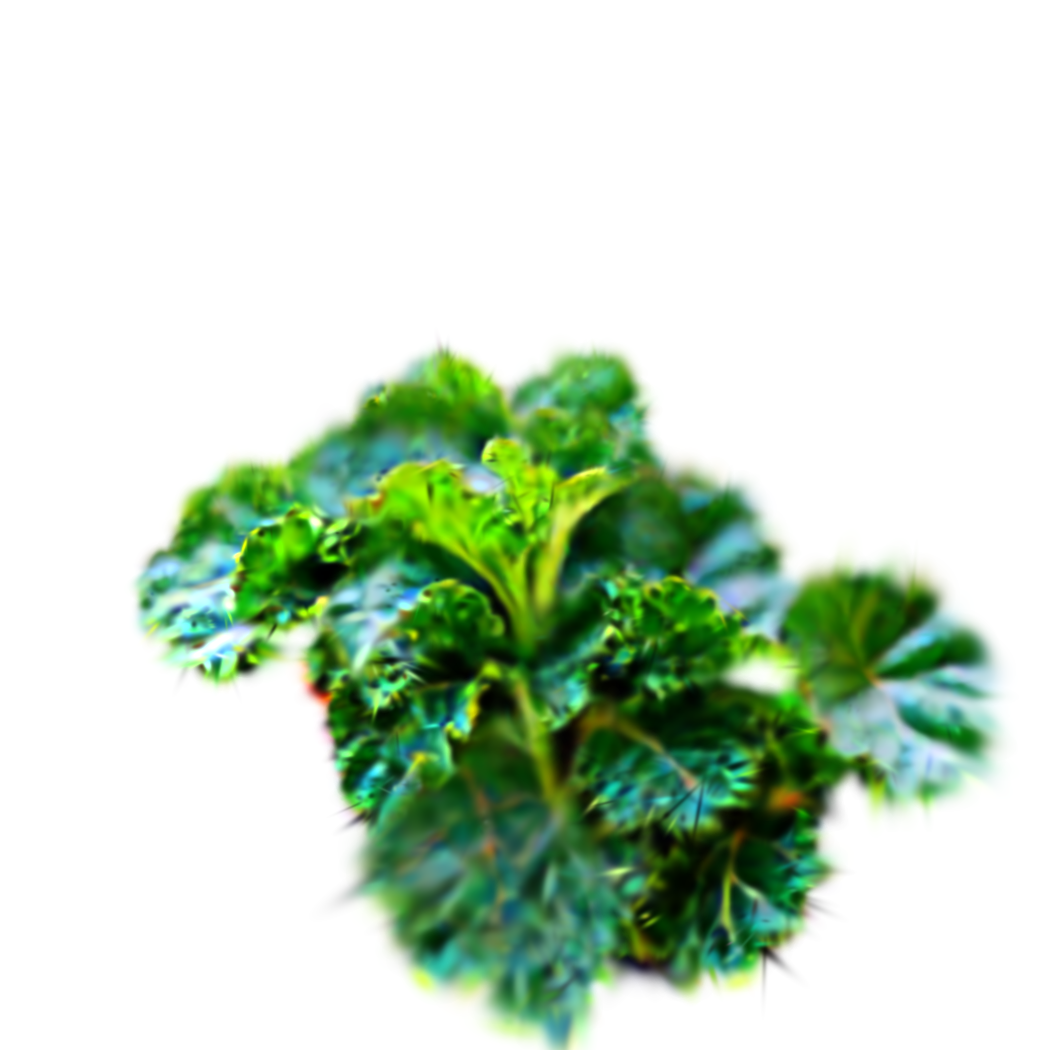} &
\includegraphics[width=0.14\textwidth]{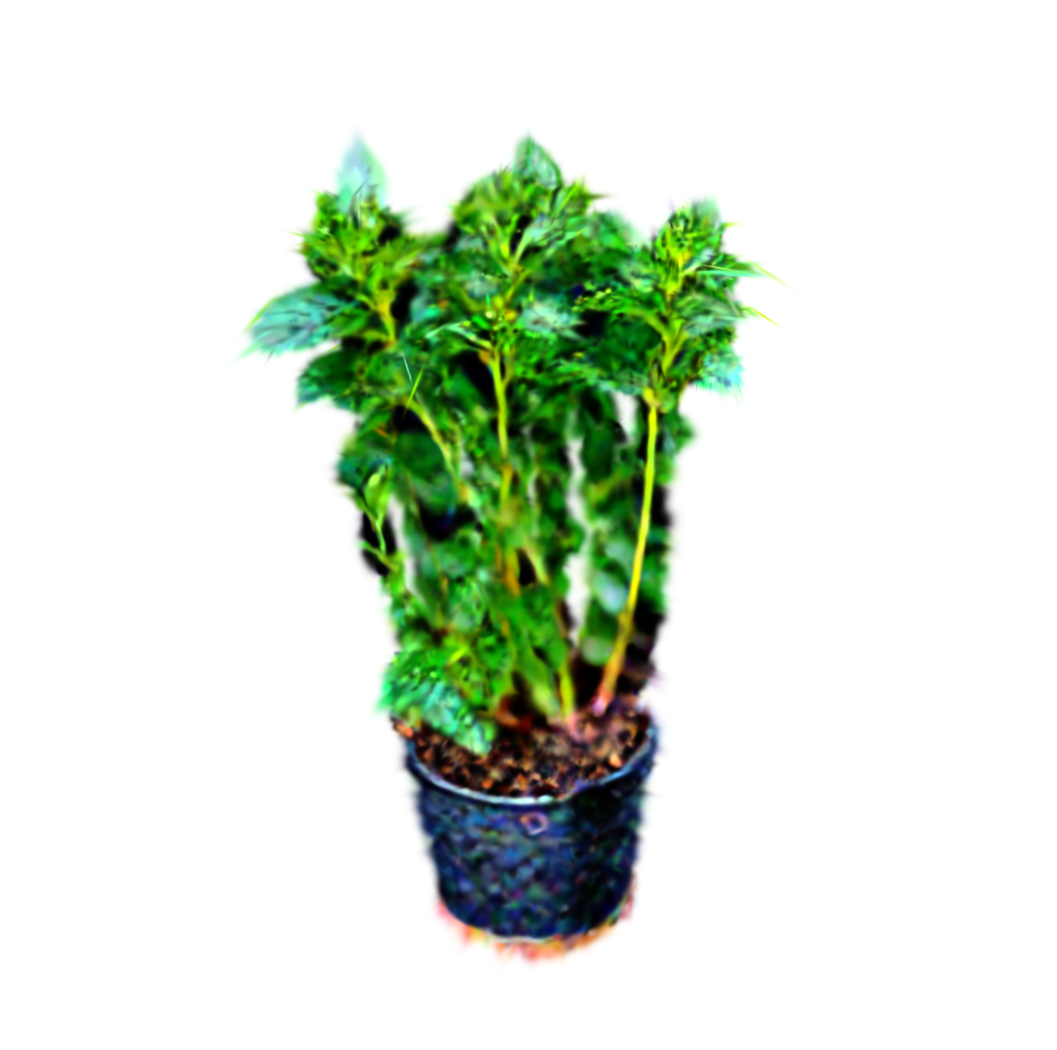} & 
\includegraphics[width=0.14\textwidth]{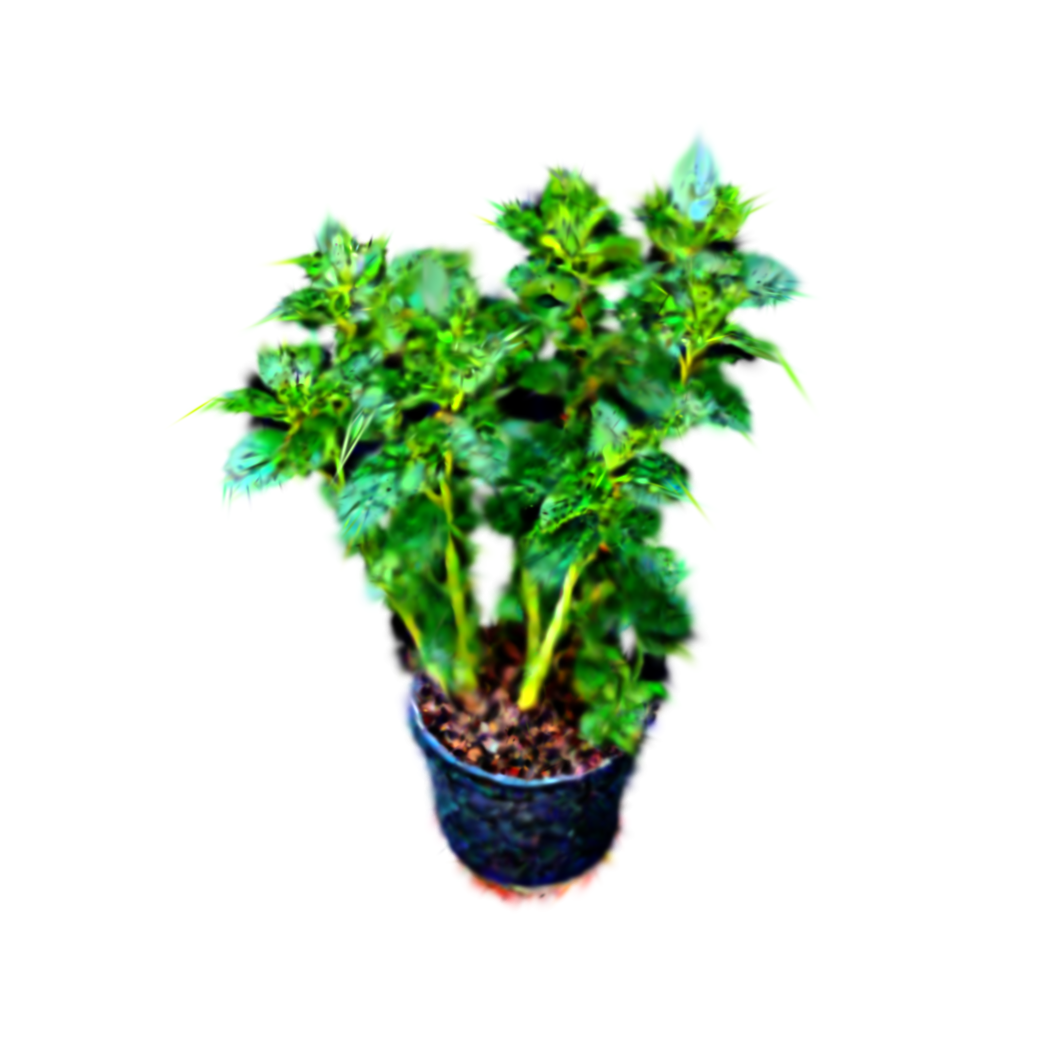}\\ 

\raisebox{3.5em}{\parbox{2.3cm}{\centering PlantDreamer }} & \includegraphics[width=0.14\textwidth]{images/rendered_imgs/Real/3DGS/00009_bean_2_real.png} & \includegraphics[width=0.14\textwidth]{images/rendered_imgs/Real/3DGS/00023_bean_2_real.png} & \includegraphics[width=0.14\textwidth]{images/rendered_imgs/Real/3DGS/00001_kale_2_real.png} & \includegraphics[width=0.14\textwidth]{images/rendered_imgs/Real/3DGS/00018_kale_2_real.png} &
\includegraphics[width=0.14\textwidth]{images/rendered_imgs/Real/3DGS/00002_mint_4_real.png} & 
\includegraphics[width=0.14\textwidth]{images/rendered_imgs/Real/3DGS/00017_mint_4_real.png}\\ 

\end{tabular}
\captionsetup{justification=justified, width=2\textwidth}
\raggedleft
\caption{\hspace{8cm} Figure 8. Comparison of 3D bean, kale and mint from GaussianDreamer and PlantDreamer models compared to a set of ground truth images.}
\label{tab:real_comparison} 
\end{figure}

\end{document}